\RequirePackage{fix-cm}
\documentclass[twocolumn]{svjour3}     % twocolumn
\smartqed                  % flush right qed marks

\usepackage[super]{nth}
\usepackage{graphicx}
\usepackage{textcomp}
\usepackage[table, x11names]{xcolor}
\usepackage{multirow}
\usepackage{placeins}
\usepackage{enumitem}
\usepackage{bm}
\usepackage{booktabs}
\usepackage{hyperref}
\usepackage{footmisc}
\usepackage{authblk}
\usepackage[tracking=true]{microtype}
\usepackage[toc,page]{appendix}
\usepackage{caption}
\usepackage{amsmath,amssymb,amsfonts}
\usepackage{mathtools}

\usepackage{algorithmic}
\usepackage{xfrac}
\usepackage{mathptmx}

\usepackage{svg}
\usepackage{graphicx}
\usepackage{subcaption}
\usepackage{adjustbox}
\usepackage{float}
\usepackage{tikz}
\usepackage{wrapfig}
\usetikzlibrary{positioning,chains}

\usepackage{xr}
\usepackage[round]{natbib}   % omit 'round' option if you prefer square brackets
\journalname{Computational Brain and Behaviour}
\begin{document}

\title{Approximating the Manifold Structure of Attributed Incentive Salience from Large Scale Behavioural Data}
\subtitle{A Representation Learning Approach Based on Artificial Neural Networks}
%\titlerunning{Short form of title}    % if too long for running head

\author[1, 2]{Valerio Bonometti}
\author[2]{Mathieu J. Ruiz}
\author[1]{Anders Drachen}
\author[3,4]{Alex Wade}
\affil[1]{Department of Computer Science, University of York}
\affil[2]{Analytics and Insights Team, Square Enix Ltd.}
\affil[3]{Department of Psychology, University of York}
\affil[4]{York Biomedical Research Institute, University of York}
\authorrunning{Bonometti et al.} % if too long for running head

\institute{
V.Bonometti \at
Department of Computer Science \\
University of York \\
Deramore Lane, York, YO10 5GH \\
\email{vb690@york.ac.uk}    
}

\date{Received: date / Accepted: date}

\maketitle

%%%%%%%%%%%%%%%%%%%%%%%%%%%%%%%%%%%%%%%%%%%%%%%%%%%%%%%%%%%%%%%%%%%%%%%%%%%%%%%%%%%%%%%%

 \begin{abstract}
Incentive salience attribution can be understood as a psychobiological mechanism ascribing relevance to potentially rewarding objects and actions. Despite being an important component of the motivational process guiding our everyday behaviour its study in naturalistic contexts is not straightforward. Here we propose a methodology based on artificial neural networks (ANNs) for approximating latent states produced by this process in situations where large volumes of behavioural data are available but no experimental control is possible. Leveraging knowledge derived from theoretical and computational accounts of incentive salience attribution we designed an ANN for estimating duration and intensity of future interactions between individuals and a series of video games in a large-scale ($N> 3 \times 10^6$) longitudinal dataset. We found video games to be the ideal context for developing such methodology due to their reliance on reward mechanics and their ability to provide ecologically robust behavioural measures at scale. When compared to competing approaches our methodology produces representations that are better suited for predicting the intensity future behaviour and approximating some functional properties of attributed incentive salience. We discuss our findings with reference to the adopted theoretical and computational frameworks and suggest how our methodology could be an initial step for estimating attributed incentive salience in large scale behavioural studies.

\subclass{MSC 97R30 \and MSC 97R40}
\keywords{Incentive Salience \and Behaviour \and Artificial Neural Networks \and Manifold Learning \and Representation Learning \and Video Games}

\end{abstract}

%%%%%%%%%%%%%%%%%%%%%%%%%%%%%%%%%%%%%%%%%%%%%%%%%%%%%%%%%%%%%%%%%%%%%%%%%%%%%%%%%%%%%%%%%

\section{Introduction}
\label{intro} Individuals' body and mind are subject to continuous changes driven by physiological, cognitive and affective processes. These changes are the constituent parts of so called "internal states"  which can be thought as dynamical latent constructs able to modulate observable behaviour \citep{eyjolfsdottir2016learning,song2017reward,merel2019deep,calhoun2019unsupervised}. Among these latent states, those related to reward and motivation processes are of pivotal importance \citep{berridge2008affective}. However, despite their relevance, the study of these entities in naturalistic contexts is not straightforward: their inference is often the solutions to an "inverse problem" \citep{bishop2006pattern} where observable and easy to acquire measures (e.g. patterns of behaviour) are used to estimate the internal factors that generated them (e.g. latent states related to motivation and reward processing) \citep{song2017reward,wang2018prefrontal}. This idea is not new \citep{spearman1961general}, but it has regained traction in recent years because of the increased availability of large volumes of data collected both inside and outside controlled experimental settings. Because they enable the study of phenomena in naturalistic settings, data collected with ecologically valid approaches are particularly interesting but come with their own set of challenges \citep{hashem2015rise}: the lack of strict control on the data gathering process (which can worsen the noise to signal ratio), the complexity of the data or construct under scrutiny and the need for algorithms able to scale to very large datasets. Recent approaches based on latent variable models (e.g. Hidden Markov Models (HMM)) have shown promise in taming some of these problems \citep{calhoun2019unsupervised} but might struggle to overcome the issues related to complexity  \citep{eyjolfsdottir2016learning,schuster2007introduction} and scalability (e.g. challenges in fitting large state spaces) \citep{touloupou2020scalable}.  \\
\\
These models rely on the assumption that the latent states, despite being embedded in a high dimensional space (e.g. patterns of behaviour or brain activity), can be effectively described using much less degrees of freedom \citep{seung2000manifold, pang2016dimensionality, luxem2020identifying}. Motivation for instance could be reduced, at any given time, to a 2D plane representing the intensity and the target of the the motivated behaviour \citep{simpson2016behavioral}. In this regard, a promising line of research is the approximation of latent states through the representation generated by Artificial Neural Networks (ANNs) \citep{eyjolfsdottir2016learning,song2017reward,merel2019deep,luxem2020identifying, pereira2020quantifying, mccullough2021unsupervised, shi2021learning}. ANNs are designed for applications with large amounts of data \citep{oh2004gpu}, provide noise resiliency and are able to capture complex interactions in the data \citep{bengio2017deep}. These desirable properties however come at the cost of interpretability and ANNs are often declared inaccessible black boxes only capable of  efficient input-output mapping. In line with a growing tendency in the literature \citep{barak2017recurrent,kietzmann2018deep, luxem2020identifying, pereira2020quantifying, mccullough2021unsupervised, shi2021learning}, we argue that this is only partially true and that given full access to the computations performed by an ANN, a certain degree of interpretability can be achieved. Through the use of prior theoretical knowledge, it is possible to constrain the input, the objective and the architecture of an ANN in order to generate so called "latent representations" (which can be thought as un-observed variables able to explain observable phenomena). Through reverse engineering, it is possible to extract the manifold structure embedded in these high dimensional representations and test it against theory driven hypotheses \citep{barak2017recurrent,kietzmann2018deep}. In line with this principle, we propose theoretical and methodological foundations for approximating the manifold structure of motivation and reward related latent states using behavioural data and ANN. Although our work aims to generalize to various areas of application, our experimental efforts exclusively make use of telemetry data coming from videogames. Since video games rely heavily on reward and motivational processes for producing playing behaviour  \citep{chumbley2006affect,wang2011game,phillips2013videogame,avserivskis2017computational, agarwal2017quitting, steyvers2019joint} and allow to record large volumes of behavioural data in a naturalistic fashion \citep{drachen2015behavioral}, we considered them to be a suitable initial test bed for our work. We trained an ANN model (designed according to a-priori theoretical knowledge) to estimate the duration and intensity of future interactions between individuals and a diverse range of video games. This was then used to generate latent representations that we compared, at the functional level, with the construct of attributed incentive salience, a particular type of latent state guiding motivated behaviour. We found attributed incentive salience to be a suitable framework for designing our model and interpreting the generated representations due to its strong connections with established psychobiological theories of reward, learning and motivation. Through a series of three experiments we show that, on the predictive task, our approach outperforms competing ones while also producing representations that mimic the functionalities of attributed incentive salience. \\
\\
The paper starts by illustrating theoretical and computational accounts of incentive salience attribution and shows how they inspired the implementation of our approach. We then describe how our model can be applied to behavioural data coming from video games, highlighting the value of this type of large-scale longitudinal data for the study of motivation-related latent states. A series of theory-driven hypotheses are then presented with the aim of evaluating the presence of specific desirable properties in the latent representation generated by our approach. These are investigated through a set of three different analyses, the results of which are then discussed in light of the adopted theoretical framework and potential application of our approach. Finally, we outline limitations of our work and the steps that could be adopted to mitigate them. 

%%%%%%%%%%%%%%%%%%%%%%%%%%%%%%%%%%%%%%%%%%%%%%%%%%%%%%%%%%%%%%%%%%%%%%

\section{Theoretical Framework}
\label{theoretical_framework}

In this section we will discuss the concept of motivation and how it can be used to describe the interactions that individuals have with particular objects or activities. We will first provide a short historical review of theories of reward-driven motivation focusing in the last section on how these led to the incentive salience hypothesis formulated by Berridge and Robinson \citep{berridge1998role}. While acknowledging the contribution of other theories in the definition of reward-driven motivation the present work will mostly focus on the area of behavioural neuroscience and make use of the framework provided by Berridge and Robinson \citep{berridge1998role}. The reason behind this choice lies in the fact that we found incentive salience to be more specific to motivation and most importantly to have a more robust and clear connection to behaviour. The sections closes with an overview on the idea that latent states, like those generated by incentive salience, might be represented as a manifold embedded in patterns of brain activity and behaviour.

\subsection{Motivation}
\label{motivation}
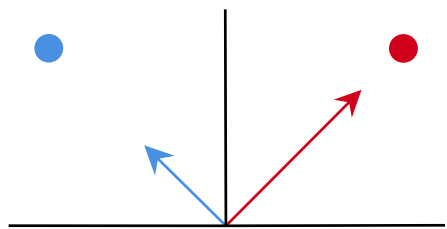
\begin{figure}[h]
  \begin{center}
    \begin{adjustbox}{width=0.7\columnwidth}
        \tikzset{every picture/.style={line width=0.75pt}} %set default line width to 0.75pt        
            \begin{tikzpicture}[x=0.75pt,y=0.75pt,yscale=-1,xscale=1]
            %uncomment if require: \path (0,300); %set diagram left start at 0, and has height of 300
            
            %Straight Lines [id:da17586827840068286] 
            \draw [color={rgb, 255:red, 208; green, 2; blue, 27 }  ,draw opacity=1 ]   (190.24,200.32) -- (237.89,152.16) ;
            \draw [shift={(240,150.03)}, rotate = 494.69] [fill={rgb, 255:red, 208; green, 2; blue, 27 }  ,fill opacity=1 ][line width=0.08]  [draw opacity=0] (9.82,-4.72) -- (0,0) -- (9.82,4.72) -- (6.52,0) -- cycle    ;
            %Straight Lines [id:da23704064540856618] 
            \draw [color={rgb, 255:red, 74; green, 144; blue, 226 }  ,draw opacity=1 ]   (190.24,200.32) -- (162.38,172.64) ;
            \draw [shift={(160.25,170.53)}, rotate = 404.81] [fill={rgb, 255:red, 74; green, 144; blue, 226 }  ,fill opacity=1 ][line width=0.08]  [draw opacity=0] (10.72,-5.15) -- (0,0) -- (10.72,5.15) -- (7.12,0) -- cycle    ;
            %Shape: Circle [id:dp3839733612485111] 
            \draw  [color={rgb, 255:red, 208; green, 2; blue, 27 }  ,draw opacity=1 ][fill={rgb, 255:red, 208; green, 2; blue, 27 }  ,fill opacity=1 ] (250.58,134.73) .. controls (250.58,132.08) and (252.73,129.93) .. (255.38,129.93) .. controls (258.03,129.93) and (260.18,132.08) .. (260.18,134.73) .. controls (260.18,137.38) and (258.03,139.53) .. (255.38,139.53) .. controls (252.73,139.53) and (250.58,137.38) .. (250.58,134.73) -- cycle ;
            %Shape: Circle [id:dp009447430116501954] 
            \draw  [color={rgb, 255:red, 74; green, 144; blue, 226 }  ,draw opacity=1 ][fill={rgb, 255:red, 74; green, 144; blue, 226 }  ,fill opacity=1 ] (120.47,134.66) .. controls (120.47,132.02) and (122.61,129.88) .. (125.25,129.88) .. controls (127.89,129.88) and (130.03,132.02) .. (130.03,134.66) .. controls (130.03,137.29) and (127.89,139.43) .. (125.25,139.43) .. controls (122.61,139.43) and (120.47,137.29) .. (120.47,134.66) -- cycle ;
            %Straight Lines [id:da5714219654763008] 
            \draw    (190,120.28) -- (190.24,200.32) ;
            %Straight Lines [id:da22247567054966455] 
            \draw    (110.5,200.28) -- (190.24,200.32) ;
            %Straight Lines [id:da4124676757039665] 
            \draw    (270.64,200.12) -- (190.24,200.32) ;

            \end{tikzpicture}
    \end{adjustbox}
  \end{center}
\caption{\textbf{Motivation as a vector}. Blue and red dots represent two objects with different characteristics while the two arrows illustrate the hypothetical motivational propensity of an individual (or two individuals) towards them. The black segments delineate the space created by the combination of the objects' characteristics and the motivational propensity of the individuals. Here the red object has the potential to generate more behaviour than the blue object possibly as a result of its characteristics and those of the individual interacting with it.}
\label{fig: vect_mot}
\end{figure}

Motivation is fundamental to everyday life: it is a process directing and helping individuals to achieve goals efficiently in  environments that present multiple courses of action at any moment \citep{ikemoto1996dissociations,mcclure2003computational}. A formal definition of motivation should encompass both the behavioural and the underlying psychobiological level. We will focus first on behavioural aspects of the process and outline in the following section a theoretical account of motivation which also covers the psychobiological level. Motivation describes why individuals react in particular ways when encountering stimuli regarded of high relevance and why they approach those stimuli at particular times \citep{berridge2004motivation}. These types of stimuli are said to possess "rewarding properties", which can be defined as the positive value ascribed to an object, a behavioral act or an internal physical state as the result of an active process of the mind and the brain \citep{schultz1997neural,berridge2008affective}. Importantly, motivation is not purely driven by the fulfilment of fundamental needs like nutrition or reproduction (so-called "primary reward objects" \citep{schultz2000reward}) or the avoidance of negative consequences like physical pain. It also extends to those volitional objects and activities which do not appear to be necessary for the survival of the individual (i.e. "secondary reward objects" \citep{berridge2008affective,sescousse2013processing}). For those activities the expectation of the amount of reward received is learned over time and may vary significantly between individuals \citep{berridge2008affective,simpson2016behavioral}.In spite of this, it would be inefficient to have dedicated and specialized motivational systems for every combination of individuals and objects (e.g. an individual's motivation for playing sport or eating food). Instead, we can think of motivation as a single overarching entity that controls the interaction between individuals and objects in an agnostic manner \citep{simpson2016behavioral}. An analogy may be drawn with the geometric concept of a vector. Looking at Figure \ref{fig: vect_mot} we can imagine the focus on a specific object being represented by the angle of the vector while its length is the intensity of the motivational process (or the amount of motivated behaviour) \citep{simpson2016behavioral}. This, can be thought as a dynamic quantity defined by the state of the individual and the rewarding properties of the object \citep{toates1994comparing,berridge2004motivation,zhang2009neural}.
If motivation acts as a single overarching process, we expect it to predict and explain goal directed behaviours seamlessly across a heterogeneous range of situations and individuals. Motivational theories based on the concepts of reward and incentive are promising candidates for this because, relying on consistent and plausible psychobiological bases, they tend to operate abstracting from the nature of the individuals and the objects. \citep{ikemoto1999role,berridge1998role,salamone2002motivational,berridge2004motivation,armony2013cambridge,corbit2015learning}.
\newline
\newline
One of the early formulations of reward-based motivation proposed by Bolles \citep{bolles1972reinforcement} suggested that individuals were motivated by the "expectation of incentive outcomes". In short,  individuals learn (and eventually anticipate) associations between their actions and the potential pleasurable outcomes associated to them \citep{bolles1972reinforcement,berridge2004motivation}. Expanding on this idea, Bindra suggested that the learning process does not just generate pleasure expectations in response to specific behaviours but it also allows individuals to perceive the behaviours themselves as a source of hedonic reward \citep{bindra1978adaptive,berridge2004motivation}. Further work by Toates \citep{toates1994comparing} asserted that the magnitude of the perceived incentives introduced by both Bolles and Bindra is modulated by the internal states of the individual \citep{toates1994comparing,berridge2004motivation}. In other words, the incentive expectations (and consequently the associated motivated behaviours) learned by an individual can change over time depending on the individual's internal state.

\subsection{Incentive Salience Hypothesis of Motivation}
\label{incentive_salience}
The approaches proposed by Bolles, Bindra and Toates,  provide an account of reward-based motivation but they assume that there is no distinction between the affective dimension of an incentive (i.e. how pleasurable it is) and the purely motivational aspect of it (i.e. how much goal directed behaviour it can produce) \citep{bindra1978adaptive,toates1994comparing}. Expanding on this, Berridge and Robinson proposed that the motivational process controlling the interaction between individuals and objects might not be a unitary mechanism but rather a composite process having specific and dissociable components which rely on specialized neurobiological mechanisms, namely: \emph{liking}, \emph{wanting} and \emph{learning} \citep{berridge1998role,berridge2009dissecting,smith2011disentangling}.

\paragraph{\textbf{Liking}}
\label{liking}
The \emph{liking} component describes the pleasure expected by an individual when interacting with an object \citep{berridge2009dissecting}. It is responsible for the hedonic quality of an experience and acts as a signal indicating that interacting again with that object might be beneficial. Despite the fact that \emph{liking} plays an important role in the incentive salience hypothesis of motivation it is difficult to measure it outside controlled laboratory environments \citep{berridge1998role} and it will not form a central theme of this research. Instead, we will focus on the "wanting" and "learning" components.

\paragraph{\textbf{Wanting}}
\label{wanting}
The \emph{wanting} component, or "incentive salience", has the function of generating and holding latent representation of objects and behavioural acts and of attributing value to them through learning mechanisms. These "valued representations" can then be used by action selection systems in order to make certain behaviours more likely \citep{ikemoto1996dissociations,berridge1998role,mcclure2003computational,berridge2004motivation}. As a consequence of this, when an object is attributed with incentive salience it will more likely draw the subject's attention and become the focus of goal directed behaviours \citep{berridge2004motivation}. Interestingly, \emph{wanting} seems to be more than a simple form of value-caching but rather a dynamic process in constant change \citep{robinson1993neural,zhang2009neural,tindell2009dynamic,berridge2012prediction}. This is because the saliency of an object depends both on its attributed value but also on the state of the individual interacting with it. A change in the individual's internal state can dampen, magnify or even revert the amount of attributed salience. \citep{robinson1993neural,zhang2009neural,tindell2009dynamic,berridge2012prediction}. It is important to note that \emph{wanting} is not the hedonic expectation associated to an object, (which is designated by \emph{liking}), but rather the process promoting the approach towards an object and the interaction with it \citep{berridge2009dissecting,robinson2015roles}. Despite the fact that \emph{liking} and \emph{wanting} are often correlated (i.e. I want what I like and vice versa) they can occasionally be triggered separately: addictive behaviours for instance are a notable example of \emph{wanting} without \emph{liking} \citep{robinson1993neural}. The functional dissociation between these two components is linked to differences in the underlying neurobiological substrate \citep{berridge2009dissecting,smith2011disentangling}. Neurotransmitters and brain areas responsible for \emph{wanting} appear to be more numerous, diverse and easily activated than those for \emph{liking} \citep{berridge2009dissecting,robinson2015roles}. As a consequence, increased incentive salience can be obtained by raising dopamine levels in many portion of the striatum without the need for the synchronized activity in other areas \citep{berridge2009dissecting,smith2011disentangling,meyer2015motivational}. This implies that the \emph{wanting} component tends to produce more robust behavioural indicators in the form of increased amount and frequency of interactions between an individual and an object \citep{berridge1998role}, which makes it a promising candidate for behavioural studies in conditions where strict experimental control is not possible.

\paragraph{\textbf{Learning}}
\label{learning}
The last component in the formulation proposed by Berridge and Robinson \citep{berridge1998role,berridge2004motivation} consists of mechanisms that provide an individual with the capability to predict, based on past experiences, the occurrence of future pleasurable outcomes (i.e. \emph{liking} reactions) when interacting with specific objects. These are similar to the learning process postulated by Bindra \citep{bindra1978adaptive} and have a twofold function. These mechanisms allow the attribution and change of incentive salience properties to previously \emph{liked} objects (e.g. primary reward objects) but they also enable subjects to learn the hedonic value of initially neutral stimuli (e.g. secondary reward objects). The \emph{learning} mechanism is based on classical conditioning: through repeated interactions with an object an individual will learn its hedonic properties and consequently attribute incentive salience to it \citep{berridge2004motivation,berridge2009dissecting}. This process is driven by mechanisms similar to those of reward-prediction error: learning is driven by spikes in dopaminergic activity generated by a mismatch between expected and experienced rewards. \citep{schultz1997neural,schultz2000multiple,flagel2011selective}.

\subsection{Manifold Representation of Incentive Salience}
\label{manifold_state}
As anticipated in the introduction, we can think of the level of attributed incentive salience (i.e. the amount of \textit{Wanting} attached to a particular object) as a latent state influencing motivated behaviour. The representation of these latent states is usually carried out by the activity of multiple brain regions responsible to generate patterns of observable behaviour. As we mentioned before, these multidimensional patterns are believed to reside on a manifold \citep{seung2000manifold, pang2016dimensionality}: a connected low dimensional region embedded within a high dimensional space \citep{bengio2017deep}. An intuitive example of this is how the brain generates and stores mental maps of the environment which are then used for navigation tasks \citep{derdikman2011manifold, nieh2021geometry}. The dimensionality of the encoding signal is much larger than the intrinsic dimensionality of the spatial information encoded within it, indeed the activity of large neuronal populations is involved in generating a mapping that needs to be only 3 dimensional. When applied to incentive salience an intuitive representation sees the manifold as a 2 dimensional space, similar to what presented in Figure \ref{fig: vect_mot}, generated by the activity of all those brain areas involved in the attribution of value and subsequent modulation of future motivated behaviour. This two dimensional space would represent, at any given time, the motivational saliency than an organism attributes to a potentially rewarding object and therefore also the intensity of the related  behaviour \citep{berridge1998role, simpson2016behavioral}. This idea of a neural manifold has found experimental support in different areas (e.g. motor control \citep{gallego2017neural}, mnemonic processes \citep{derdikman2011manifold, nieh2021geometry}, reward processing \citep{bromberg2010coding} visual \citep{seung2000manifold, ganmor2015thesaurus} and olfactory \citep{stopfer2003intensity} perception) and rely on the fact that neural activity is highly redundant and reducible to just few correlation patterns \citep{gallego2017neural}. In light of this, the use of dimensionality reduction techniques able to represent the manifold structure of highly dimensional data have proven to be valuable in making abstract entities like latent states more easily accessible and interpretable, while also facilitating their mapping onto brain \citep{gao2021nonlinear, rue2021decoding} and behavioural data \citep{luxem2020identifying, pereira2020quantifying, mccullough2021unsupervised, shi2021learning}. 
\\
\\
We have highlighted how motivation can be described as a mechanism that guides the interaction between individuals and objects. It controls and selects behaviours which are expected to lead to pleasurable outcomes for the individual (i.e. incentives or reward). These expectancies are the product of a learning process that can be modulated by the internal state of the individual. Therefore, from a behavioural point of view, an objects $O$ can acquire salience for an individual $I$ conditioned on its capacity to elicit rewarding experience $r$ \citep{berridge1998role,mcclure2003computational}. The amount of attributed salience is a valued representation of $O$ generated by $I$ and controls how likely and intense future interactions between the two will be. \citep{berridge1998role,mcclure2003computational}. Let $B$ represents the strength of an interaction between $I$ and $O$, $r$ a measure of how rewarding the interaction with $O$ is perceived to be and $V$ the generated attributed incentive salience.
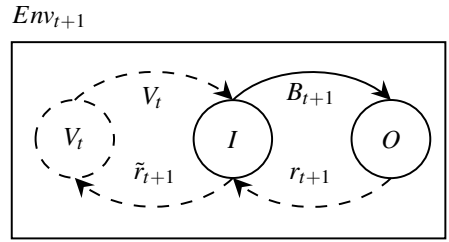
\begin{figure}[h]
  \begin{center}
    \begin{adjustbox}{width=0.7\columnwidth}

        \tikzset{every picture/.style={line width=0.75pt}}
        
        \begin{tikzpicture}[x=0.75pt,y=0.75pt,yscale=-1,xscale=1]
        
        %Shape: Circle [id:dp8739177923629333] 
        \draw   (123.48,130.05) .. controls (123.48,119.06) and (132.39,110.15) .. (143.38,110.15) .. controls (154.37,110.15) and (163.28,119.06) .. (163.28,130.05) .. controls (163.28,141.04) and (154.37,149.95) .. (143.38,149.95) .. controls (132.39,149.95) and (123.48,141.04) .. (123.48,130.05) -- cycle ;
        %Curve Lines [id:da2726669379173874] 
        \draw    (143.38,110.15) .. controls (161.99,91.83) and (200.82,90.88) .. (221.41,108.12) ;
        \draw [shift={(223.57,110.07)}, rotate = 224.02] [fill={rgb, 255:red, 0; green, 0; blue, 0 }  ][line width=0.08]  [draw opacity=0] (9.82,-4.72) -- (0,0) -- (9.82,4.72) -- (6.52,0) -- cycle    ;
        %Curve Lines [id:da4485667818478689] 
        \draw  [dash pattern={on 4.5pt off 4.5pt}]  (223.57,150.02) .. controls (204.36,169.14) and (165.2,169.5) .. (145.45,151.93) ;
        \draw [shift={(143.38,149.95)}, rotate = 405.89] [fill={rgb, 255:red, 0; green, 0; blue, 0 }  ][line width=0.08]  [draw opacity=0] (9.82,-4.72) -- (0,0) -- (9.82,4.72) -- (6.52,0) -- cycle    ;
        %Shape: Circle [id:dp8060688469385651] 
        \draw   (203.6,130.05) .. controls (203.6,119.01) and (212.54,110.07) .. (223.57,110.07) .. controls (234.6,110.07) and (243.55,119.01) .. (243.55,130.05) .. controls (243.55,141.08) and (234.6,150.02) .. (223.57,150.02) .. controls (212.54,150.02) and (203.6,141.08) .. (203.6,130.05) -- cycle ;
        %Shape: Rectangle [id:dp6054697635081123] 
        \draw   (31.2,80.5) -- (251.2,80.5) -- (251.2,180.83) -- (31.2,180.83) -- cycle ;
        %Shape: Circle [id:dp6091430023771998] 
        \draw  [dash pattern={on 4.5pt off 4.5pt}] (43.29,129.97) .. controls (43.29,118.98) and (52.2,110.07) .. (63.19,110.07) .. controls (74.18,110.07) and (83.09,118.98) .. (83.09,129.97) .. controls (83.09,140.96) and (74.18,149.87) .. (63.19,149.87) .. controls (52.2,149.87) and (43.29,140.96) .. (43.29,129.97) -- cycle ;
        %Curve Lines [id:da9876749075069771] 
        \draw  [dash pattern={on 4.5pt off 4.5pt}]  (143.38,149.95) .. controls (124.17,169.06) and (85.01,169.42) .. (65.26,151.85) ;
        \draw [shift={(63.19,149.87)}, rotate = 405.89] [fill={rgb, 255:red, 0; green, 0; blue, 0 }  ][line width=0.08]  [draw opacity=0] (9.82,-4.72) -- (0,0) -- (9.82,4.72) -- (6.52,0) -- cycle    ;
        %Curve Lines [id:da9412329086480583] 
        \draw  [dash pattern={on 4.5pt off 4.5pt}]  (63.19,110.07) .. controls (81.8,91.76) and (120.63,90.8) .. (141.21,108.05) ;
        \draw [shift={(143.38,110)}, rotate = 224.02] [fill={rgb, 255:red, 0; green, 0; blue, 0 }  ][line width=0.08]  [draw opacity=0] (9.82,-4.72) -- (0,0) -- (9.82,4.72) -- (6.52,0) -- cycle    ;
        
        % Text Node
        \draw (143.38,130.05) node  [font=\normalsize]  {$I$};
        % Text Node
        \draw (223.57,130.05) node  [font=\normalsize]  {$O$};
        % Text Node
        \draw (183.14,106.95) node  [font=\normalsize]  {$B_{t+1}$};
        % Text Node
        \draw (182.81,148.81) node  [font=\normalsize]  {$r_{t+1}$};
        % Text Node
        \draw (50.45,68.01) node  [font=\normalsize,rotate=-0.16]  {$Env_{t+1}$};
        % Text Node
        \draw (63.19,129.97) node  [font=\normalsize]  {$V_{t}$};
        % Text Node
        \draw (103.81,147.81) node  [font=\normalsize]  {$\tilde{r}_{t+1}$};
        % Text Node
        \draw (102.81,108.81) node  [font=\normalsize]  {$V_{t}$};
        \end{tikzpicture}
    \end{adjustbox}
  \end{center}
\caption{\textbf{The process of incentive salience attribution}. Solid and dashed elements represent respectively observable (i.e. behavioural) and latent aspects of the process. The individual and the object that they interact with are indicated by $I$ and $O$. The strength of the interaction is represented by $B$. The salience that $I$ attributes to $O$ is expressed by $V$ while $r$ and $\tilde{r}$ are the experienced reward and its weighted version produced by the state of $I$. All the contextual factors influencing both the amount of $r$ perceived and the magnitude of $B$ produced are represented by $Env$. The dynamical natures of the process is expressed though the arbitrary temporal unit $t$.}
\label{fig: incs}
\end{figure}
Following Figure \ref{fig: incs}, at time $t+1$ an individual will produce an interaction with an object of strength $B$ according to the previous $V_{t}$. If we recall from section \ref{learning}, this process relies heavily on learning mechanisms making $V$ by nature dynamic and mutable. It should be noted that $B$ can be represented as a multidimensional variable defined by the instrumental behaviours conventionally used for assessing the \emph{wanting} component in animal studies (e.g. frequency, amount and duration of feeding behaviours like bites, nibbles and sniffs) \citep{berridge1998role}. During and after the interaction $I$ will experience a variable degree of reward $r_{t+1}$ that, weighted by their internal state, will then be used for updating $V_{t}$.  It is worth noting that the individual's internal state is not the only factor involved in the modulation of $r$, also the context in which $I$ and $O$ interact ($Env_{t+1}$ in Figure \ref{fig: incs}) seems to contribute to this \citep{palminteri2015contextual}. Following the idea presented in section \ref{manifold_state}, the latent state defined by $V$ could be represented as a manifold defined by the activity of those regions responsible for the attribution of incentive salience. Moreover, given the strong coupling between attributed incentive salience and behaviour \citep{berridge1998role} we would also expect the structure of this manifold to be a suitable descriptor of the behavioural aspects of attributed incentive salience.

\section{Computational Framework}
\label{comp_framework}
In this section, we will propose a way to approximate the manifold structure of attributed incentive salience in scenarios where only large volumes of behavioural data are available and no strict experimental control is possible. First, we will briefly illustrate how previous work has framed the modelling of this construct as a reinforcement learning problem and solved it using Temporal Difference Learning (TD Learning) \citep{sutton1988learning}. This, will provide us with a psycho-biologically plausible computational model of attribute incentive salience and constitute the starting point for our approach. Then, we will highlight how video games are promising candidates for studying the behavioural aspects of incentive salience attribution in naturalistic settings. Finally, combining these two ideas, we will show how estimating the manifold structure of attributed incentive salience can be cast as the solution to a supervised learning problem and why Artificial Neural Networks (ANNs), thanks to their representation learning capabilities, are well suited for the task.

\subsection{Temporal Difference Learning}
\label{td_learning}

\begin{figure}[h]
  \begin{center}
    \begin{adjustbox}{width=0.7\columnwidth}

        \tikzset{every picture/.style={line width=0.75pt}}       
        
        \begin{tikzpicture}[x=0.75pt,y=0.75pt,yscale=-1,xscale=1]
        
        %Straight Lines [id:da2592063600931961] 
        \draw [color={rgb, 255:red, 208; green, 2; blue, 27 }  ,draw opacity=1 ]   (282.5,135.16) -- (300.67,135.02) ;
        \draw [shift={(279.5,135.18)}, rotate = 359.55] [fill={rgb, 255:red, 208; green, 2; blue, 27 }  ,fill opacity=1 ][line width=0.08]  [draw opacity=0] (9.82,-4.72) -- (0,0) -- (9.82,4.72) -- (6.52,0) -- cycle    ;
        %Straight Lines [id:da41585806917853485] 
        \draw [color={rgb, 255:red, 208; green, 2; blue, 27 }  ,draw opacity=1 ]   (223.33,134.87) -- (240.67,135.02) ;
        \draw [shift={(220.33,134.85)}, rotate = 0.47] [fill={rgb, 255:red, 208; green, 2; blue, 27 }  ,fill opacity=1 ][line width=0.08]  [draw opacity=0] (9.82,-4.72) -- (0,0) -- (9.82,4.72) -- (6.52,0) -- cycle    ;
        %Shape: Ellipse [id:dp31493856088290983] 
        \draw   (181,140.07) .. controls (181,128.75) and (189.89,119.58) .. (200.86,119.58) .. controls (211.83,119.58) and (220.72,128.75) .. (220.72,140.07) .. controls (220.72,151.4) and (211.83,160.57) .. (200.86,160.57) .. controls (189.89,160.57) and (181,151.4) .. (181,140.07) -- cycle ;
        %Shape: Ellipse [id:dp5103941395609547] 
        \draw  [dash pattern={on 4.5pt off 4.5pt}] (360.4,139.66) .. controls (360.4,128.34) and (369.29,119.16) .. (380.26,119.16) .. controls (391.23,119.16) and (400.12,128.34) .. (400.12,139.66) .. controls (400.12,150.98) and (391.23,160.16) .. (380.26,160.16) .. controls (369.29,160.16) and (360.4,150.98) .. (360.4,139.66) -- cycle ;
        %Straight Lines [id:da30559831832260953] 
        \draw [color={rgb, 255:red, 0; green, 0; blue, 0 }  ,draw opacity=1 ]   (220.07,144.9) -- (237.5,144.86) ;
        \draw [shift={(240.5,144.85)}, rotate = 539.86] [fill={rgb, 255:red, 0; green, 0; blue, 0 }  ,fill opacity=1 ][line width=0.08]  [draw opacity=0] (9.82,-4.72) -- (0,0) -- (9.82,4.72) -- (6.52,0) -- cycle    ;
        %Shape: Ellipse [id:dp12377292733336898] 
        \draw  [color={rgb, 255:red, 0; green, 0; blue, 0 }  ,draw opacity=1 ] (240,140.07) .. controls (240,128.75) and (248.89,119.58) .. (259.86,119.58) .. controls (270.83,119.58) and (279.72,128.75) .. (279.72,140.07) .. controls (279.72,151.4) and (270.83,160.57) .. (259.86,160.57) .. controls (248.89,160.57) and (240,151.4) .. (240,140.07) -- cycle ;
        %Shape: Ellipse [id:dp9248757796917767] 
        \draw   (300.4,139.66) .. controls (300.4,128.34) and (309.29,119.16) .. (320.26,119.16) .. controls (331.23,119.16) and (340.12,128.34) .. (340.12,139.66) .. controls (340.12,150.98) and (331.23,160.16) .. (320.26,160.16) .. controls (309.29,160.16) and (300.4,150.98) .. (300.4,139.66) -- cycle ;
        %Straight Lines [id:da050274800713848156] 
        \draw [color={rgb, 255:red, 208; green, 2; blue, 27 }  ,draw opacity=1 ]   (202.96,162.71) -- (240,200.37) ;
        \draw [shift={(200.86,160.57)}, rotate = 45.48] [fill={rgb, 255:red, 208; green, 2; blue, 27 }  ,fill opacity=1 ][line width=0.08]  [draw opacity=0] (9.82,-4.72) -- (0,0) -- (9.82,4.72) -- (6.52,0) -- cycle    ;
        %Straight Lines [id:da7059353635730191] 
        \draw [color={rgb, 255:red, 208; green, 2; blue, 27 }  ,draw opacity=1 ]   (262.01,162.66) -- (300.4,200) ;
        \draw [shift={(259.86,160.57)}, rotate = 44.2] [fill={rgb, 255:red, 208; green, 2; blue, 27 }  ,fill opacity=1 ][line width=0.08]  [draw opacity=0] (9.82,-4.72) -- (0,0) -- (9.82,4.72) -- (6.52,0) -- cycle    ;
        %Straight Lines [id:da17885982011746426] 
        \draw [color={rgb, 255:red, 208; green, 2; blue, 27 }  ,draw opacity=1 ] [dash pattern={on 4.5pt off 4.5pt}]  (322.38,162.28) -- (360.4,200.16) ;
        \draw [shift={(320.26,160.16)}, rotate = 44.9] [fill={rgb, 255:red, 208; green, 2; blue, 27 }  ,fill opacity=1 ][line width=0.08]  [draw opacity=0] (9.82,-4.72) -- (0,0) -- (9.82,4.72) -- (6.52,0) -- cycle    ;
        %Shape: Ellipse [id:dp3523005799699832] 
        \draw  [dash pattern={on 4.5pt off 4.5pt}] (360.4,200.16) .. controls (360.4,189.16) and (369.29,180.23) .. (380.26,180.23) .. controls (391.23,180.23) and (400.12,189.16) .. (400.12,200.16) .. controls (400.12,211.17) and (391.23,220.09) .. (380.26,220.09) .. controls (369.29,220.09) and (360.4,211.17) .. (360.4,200.16) -- cycle ;
        %Shape: Ellipse [id:dp3509500153089894] 
        \draw   (300.4,200) .. controls (300.4,189) and (309.29,180.09) .. (320.26,180.09) .. controls (331.23,180.09) and (340.12,189) .. (340.12,200) .. controls (340.12,210.99) and (331.23,219.9) .. (320.26,219.9) .. controls (309.29,219.9) and (300.4,210.99) .. (300.4,200) -- cycle ;
        %Shape: Ellipse [id:dp17336392317383575] 
        \draw  [color={rgb, 255:red, 0; green, 0; blue, 0 }  ,draw opacity=1 ] (240,200.37) .. controls (240,189.4) and (248.89,180.5) .. (259.86,180.5) .. controls (270.83,180.5) and (279.72,189.4) .. (279.72,200.37) .. controls (279.72,211.34) and (270.83,220.23) .. (259.86,220.23) .. controls (248.89,220.23) and (240,211.34) .. (240,200.37) -- cycle ;
        %Straight Lines [id:da21459790152477998] 
        \draw [color={rgb, 255:red, 0; green, 0; blue, 0 }  ,draw opacity=1 ]   (279.5,144.52) -- (297.5,144.8) ;
        \draw [shift={(300.5,144.85)}, rotate = 180.91] [fill={rgb, 255:red, 0; green, 0; blue, 0 }  ,fill opacity=1 ][line width=0.08]  [draw opacity=0] (9.82,-4.72) -- (0,0) -- (9.82,4.72) -- (6.52,0) -- cycle    ;
        %Straight Lines [id:da8588950475203374] 
        \draw [color={rgb, 255:red, 208; green, 2; blue, 27 }  ,draw opacity=1 ] [dash pattern={on 4.5pt off 4.5pt}]  (342.5,135.16) -- (360.67,135.02) ;
        \draw [shift={(339.5,135.18)}, rotate = 359.55] [fill={rgb, 255:red, 208; green, 2; blue, 27 }  ,fill opacity=1 ][line width=0.08]  [draw opacity=0] (9.82,-4.72) -- (0,0) -- (9.82,4.72) -- (6.52,0) -- cycle    ;
        %Straight Lines [id:da9169375689094925] 
        \draw [color={rgb, 255:red, 0; green, 0; blue, 0 }  ,draw opacity=1 ] [dash pattern={on 4.5pt off 4.5pt}]  (339.5,144.52) -- (357.5,144.8) ;
        \draw [shift={(360.5,144.85)}, rotate = 180.91] [fill={rgb, 255:red, 0; green, 0; blue, 0 }  ,fill opacity=1 ][line width=0.08]  [draw opacity=0] (9.82,-4.72) -- (0,0) -- (9.82,4.72) -- (6.52,0) -- cycle    ;
        
        % Text Node
        \draw (320.26,201.59) node    {$r_{t+1}$};
        % Text Node
        \draw (200.86,140.07) node    {$\widehat{V}(s_{t-1})$};
        % Text Node
        \draw (320.26,139.66) node    {$\widehat{V}(s_{t+1})$};
        % Text Node
        \draw (380.26,139.66) node  [font=\small]  {$\widehat{V}(s_{T})$};
        % Text Node
        \draw (259.86,202) node  [color={rgb, 255:red, 208; green, 9; blue, 2 }  ,opacity=1 ]  {$\textcolor[rgb]{0,0,0}{r}\textcolor[rgb]{0,0,0}{_{t}}$};
        % Text Node
        \draw (259.86,140.07) node  [color={rgb, 255:red, 208; green, 9; blue, 2 }  ,opacity=1 ]  {$\textcolor[rgb]{0,0,0}{\widehat{V}}\textcolor[rgb]{0,0,0}{(s_{t})}$};
        % Text Node
        \draw (380.26,200.16) node    {$r_{T}$};

        \end{tikzpicture}
    \end{adjustbox}
  \end{center}
\caption{\textbf{Graphical representation of TD Learning}. Red arrows indicate the flow of the computations for deriving $\delta$ and updating $\widehat{V}$ expressed by equations \ref{td_error} and \ref{td_update}. Black arrows instead indicate the changes of $\widehat{V}$ moving from $s$ to $s_{t+1}$. Solid circles indicate states which have already been observed while dashed ones represent future not-yet observed states.}
\label{fig: td_learning}
\end{figure}
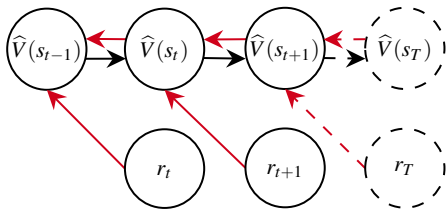

The first attempt to model incentive salience attribution was carried out by McClure \textit{et. al.} using TD Learning \citep{mcclure2003computational}. The use of TD Learning in simulation studies involving reward learning, is often motivated by its good approximation of the reward-prediction error signal generated by dopaminergic neurons \citep{schultz1997neural,flagel2011selective}. Algorithms in the family of TD Learning attempt to learn a value function $V$ by iteratively refining an estimate $\widehat{V}$ over time \citep{sutton2018reinforcement}. In the most basic form, called TD(0), this is done by simply observing the reward $r$ associated with a particular state $s$ at time $t_{+1}$ and using it to adjust the estimate of $V$ produced at time $t$ \citep{sutton2018reinforcement}. Here $t$ is an arbitrary unit of time, it can be specific (i.e. seconds) or generic (i.e. a point in an ordered series of events) depending on the type of application. If we let $S=\{s_{t}: t \in T\}$ be a sequence of states, then the value $V$  at $s_{t}$ is given by the sum of all future discounted rewards expected when transitioning from $s_t$ to $s_{t+1}$
\begin{align}
\label{td_v}
    V(s_t) 
        &= E[
            r_{t+1} + 
            \gamma r_{t+2} + 
            \gamma^{2} r_{t+3} +
            ... +
            \gamma^{T} r_{T}
        ]\\
        &= E[
            r_{t+1} + 
            \gamma V(s_{t+1}) 
        ] \nonumber
\end{align}
with $\gamma \in [0, 1]$ being a discounting factor for $r$. The iterative refining of $\widehat{V}$ carried out by TD learning is achieved by first computing an error signal $\delta$ at time $t$
\begin{align}
    \label{td_error}
    \delta(t) = r_{t+1} + \gamma \widehat{V}(s_{t+1}) - \widehat{V}(s_{t})
\end{align}
which quantifies the difference between the current $\widehat{V}$ and what is expected when transitioning to $s_{t+1}$. Once the error signal is computed, $\widehat{V}(s_{t})$ is updated as: 
\begin{align}
    \label{td_update}
    \widehat{V}(s_t)  \leftarrow \widehat{V}(s_t) + \alpha(\delta)
\end{align}
where $\alpha \in [0, 1]$ is a constant controlling the amount of updating or the "learning rate". This process called TD update is illustrated by the diagram presented in Figure \ref{fig: td_learning}. Conventionally the transition from $s$ to $s_{t+1}$ is the result of an action selection process guided by $\widehat{V}$, because in optimal control settings the role of reinforcement learning is to select the course of action that maximizes future rewards \citep{schultz1997neural,mcclure2003computational,sutton2018reinforcement}.
McClure \textit{et. al.} proposed that incentive salience is represented by $V$ as defined in equation \ref{td_v} while the error signal expressed by equation \ref{td_error} represents the activity of dopaminergic neurons with the dual function of driving the attribution of incentive salience (through reward prediction error coding as specified in section \ref{incentive_salience}) and guiding the previously mentioned action selection process \citep{schultz1997neural,mcclure2003computational,o2003temporal}. However, in later work, Zhang \textit{et. al.} highlighted the fact that the model proposed by McClure \textit{et. al.} fails to take into account an important part of the original incentive salience hypothesis: the dynamic modulation produced by the individual's internal state (see section \ref{wanting}) \citep{toates1994comparing,mcclure2003computational,berridge2004motivation,zhang2009neural,tindell2009dynamic,berridge2012prediction}. Zhang \textit{et. al.} therefore proposed a modification of the original TD Learning model to include a modulatory factor $k \in [0, +\infty]$ which can enhance ($k > 1$), dampen or even revert ($k < 1$) previously learned incentive salience values
\begin{align}
    \label{zhang_td_v}
    V(s_t) = E[\tilde{r}(r_{t+1},k_{t}) + \gamma V(s_{t+1})]
\end{align}
here $\tilde{r}(.,.)$ is a function of two variables and can assume either an additive or multiplicative form \footnote{See \citep{zhang2009neural} for detailed description of the two forms and their functional differences.}. The main difference between the approaches of McClure \textit{et. al.} and Zhang \textit{et. al.} lies in the interpretation of $V$. Both authors see it as a combination of cached value (i.e. what has been learned from past experiences) and expectation over future $r$ but for McClure \textit{et. al.} all the future $r$ have the same weight while for Zhang \textit{et. al.} the state of the individual dynamically modulates the weighting of $r$. Using the notation from section \ref{incentive_salience}, we can say that the interaction $s$ between $I$ and $O$ at time $t_{+1}$ arises from the $V$ (i.e. incentive salience) generated after $s_{t}$. The mismatch between the predicted amount of reward and the actual reward received at time $t_{+1}$ generates an error signal that allows $I$ to learn about the "correct" magnitude of $V(s_{t})$ \citep{schultz2017reward} . As an example, an individual may anticipate that eating their favourite meal would be a rewarding experience but instead (for some reason) it was underwhelming. They therefore reduce the salience previously attributed to it. Importantly, $V(s_{t})$ does not just encompass the previous history of interactions between $I$ and $O$ but also the current state of $I$: the individual has learned from long experience that eating is a pleasurable activity but currently, since they are sated they do not expect much reward from doing it again in the near future.  

\paragraph{\textbf{From TD to Supervised Learning}}
\label{td_to_supervised}
The approaches discussed above frame the estimation of attributed incentive salience as a reinforcement learning task. This requires the simulation of a sequence of interactions between $I$ and $O$ and the concomitant delivery of $r$  \citep{schultz1997neural,mcclure2003computational,zhang2009neural}. However, it is not always straightforward to replicate these interactions in real world scenarios, especially when dealing with human participants. The control on the internal state of $I$ and amount of $r$ delivered that McClure and Zhang assume is usually based on strict assumptions and can be achieved only in controlled experimental settings \citep{mcclure2003computational,zhang2009neural}. As an alternative solution for inferring $V$ outside the laboratory we propose to learn its manifold structure through supervised learning. Differently to what reported in the literature \citep{calhoun2019unsupervised, mccullough2021unsupervised, luxem2020identifying, pereira2020quantifying, shi2021learning} we argue that in this case the use of supervised in place of un-supervised techniques is to be preferred. Indeed, since we are dealing exclusively with behavioural data and trying to solve an inverse problem  we would like to learn a manifold structure which is not just a generic indicator of behavioural phenotype \citep{luxem2020identifying} but also obeys to specific functional constrains.\\
\\
In this approach, an experimenter gathers data on a set of interactions between $I$ and $O$ and let a learning algorithm to estimate two functions:
\begin{gather}
\label{supervised_v}
    V(s_{t}) = f^{1}(O, \tilde{r_{t}}, V(s_{t-1}); \theta^{1}) \\
    r_{t+1} = f^{2}(V(s_{t}); \theta^{2}) \nonumber
\end{gather}
here $f^{1}$ and $f^{2}$ are arbitrarily complex functions while $\theta^{1}$ and $\theta^{2}$ are parameters that the learning algorithm has to infer. The future reward that an individual expects after an interaction with an object is produced by the current level of attributed salience, which  itself is a function of the current internal state of the individual (expressed through the amount of reward just experienced) and the incentive salience previously attributed to the object. It is important to note that the two functions above need to be recursive over all $s \in S$ (see equations \ref{td_v} and \ref{zhang_td_v}) in order to provide $V(s_{t})$ with the dual purpose of caching all the past $V$ and being a suitable predictor for all the $r$. This formulation however still requires a measure of the $r$ experienced by $I$ (or more precisely its weighted version $\tilde{r}$) after interacting with $O$, which is not easily accessible. However, Thorndike's law of effect \citep{thorndike1927law} and Skinner's operant conditioning principles \citep{skinner1965science} suggest that $r$, which  like $V$ is a non observable latent variable, manifests itself through the intensity of interactions between $I$ and $O$ (i.e. $B$ in Figure \ref{fig: incs} and section \ref{motivation}): the frequency and amount of object-directed behaviours increase or decrease as a function of the rewards an individual expects to receive \citep{berridge2004motivation,schultz2017reward}. Since $V(s_{t})$ predicts how much $r$ an $I$ expects to receive from interacting with $O$, we should also expect the strength of their future interactions to be a function of $V(s_{t})$. This can be represented re-arranging the equations in \ref{supervised_v} in a more compact form as a chain of functions
\begin{align}
\label{supervised_b}
    B_{t+1} = f^{2}(f^{1}(O, B_{t}, V(s_{t-1}); \theta^{1});  \theta^{2})
\end{align}
To approximate the above expression, a learning algorithm would require records of behaviours generated by individuals while interacting with a diverse set of potentially rewarding objects. Here, we argue that video games are one way to obtain this type of data at scale while also achieving some level of ecological validity.

\subsection{Video Games and Telemetry}
\label{videogame_telemetries}
Interacting with video games is a volitional activity driven largely by the capacity of the games to provide pleasurable experiences \citep{boyle2012engagement}. Behaviour within the game is best understood as the result of a value attribution process similar to that of secondary reward objects (see section \ref{incentive_salience}). Indeed, it appears that the play behaviour is often produced and maintained by the structural characteristics of the game (e.g. game mechanics) \citep{king2010video} which, working like conventional reinforcement mechanisms \citep{chumbley2006affect,wang2011game,phillips2013videogame,avserivskis2017computational}, produce effects similar to operant conditioning \citep{skinner1965science}. Although caution should be applied when complex activities are investigated using neuroimaging techniques, evidence suggest that the maintenance of video games playing behaviour engages the same cortico-striatal structures \citep{hoeft2008gender,mathiak2011reward,cole2012interactivity,klasen2012neural,lorenz2015video,gleich2017functional} and neurotransmitters \citep{koepp1998evidence} involved in reward processing. This, seems also supported at the behavioural level where the ammount of experienced in-game reward appears to play a role in controlling how likely is an individual to keep engaging in playing behaviour \citep{agarwal2017quitting, steyvers2019joint}. This, in conjunction with a growing literature highlighting similarities between certain video game mechanics and activities driven by secondary rewards (e.g. gambling) \citep{king2010role,drummond2018video,zendle2018video}, corroborates the idea that video games are able to elicit behavioural responses through incentive mechanics. In this view, video games with different structural characteristics could be seen as objects possessing rewarding properties that heavily depend on the individuals interacting with them (e.g. an individual's preference for a specific game mechanic). Hence, similarly to the process specified in section \ref{theoretical_framework}, we can expect that through repeated interactions, an individual will experience varying degrees of reward determined by their internal state and the characteristics of the game. These interactions will produce continuous adjustments in the level of saliency attributed to playing that specific game which in turn will influence the frequency and amount of future interactions with that same game. Other than offering a context for observing the process of incentive salience attribution, video games allow us to obtain large volumes of behavioural data (similar to those mentioned in section \ref{td_to_supervised}) in a naturalistic fashion. This is made possible by the widespread practice of obtaining high frequency records (i.e. telemetry\footnote{See \citep{el2016game} for a more technical description of telemetry in video games.}) of players' behaviour during the game \citep{drachen2015behavioral}. This approach, despite offering less control and rigour than conventional experimental procedures, allows us to obtain a more faithful representation of natural behaviour (similarly to field studies) while avoiding some of the limitations connected with laboratory-based studies (e.g. sampling and observer biases).
\newline
\newline
In order to use this type of behavioural data to model attributed incentive salience, a learning algorithm should possess the following properties. First, it should be scalable and noise resilient, to leverage large volumes of naturalistic data in an efficient and effective manner. Second, it should be able to approximate arbitrarily complex functions, given that the shape of the functions specified in equation \ref{td_to_supervised} is not known a-priori. And finally, it should be able to produce an approximation of $V(s_{t})$ that can be inspected in order to evaluate if its functional properties can be compared with those of attributed incentive salience. Artificial Neural Networks (ANNs) appear to satisfy these requirements.

\subsection{Artificial Neural Networks}
\label{artificial_neural_networks}
In their conventional form, ANNs can be seen as chains of nested functions (the layers of the network). These layers are vector valued (there are multiple units or neurons in each layer) and organized as directed acyclic computational graphs (information only flows forward). When the number of layers is greater than two, the prefix "deep" is usually applied \citep{bengio2017deep}. The goal of this ensemble of functions is to create a mapping between an input $x$ and a target $y$. Following the example illustrated in Figure \ref{fig: ffnn}, given the set of parameters $\Theta = \{\theta^1, \theta^2 \}$ an ANN would first infer a function $h = f(x;\theta^{1})$, mapping the input to a new representation $h$. The same representation $h$ would then become the input of a second function $\widehat{y} = f^{1}(h;\theta^{2})$ which produces an estimate of the target \citep{bengio2017deep}. In this sense, we can think of each layer as a collection of many non-linear vector to scalar functions taking the previous layer as input and generating the units for the layer that follows \citep{bengio2017deep}. By increasing the number of layers and units, ANNs can approximate an extremely large class of functions \citep{rumelhart1986learning}.
\input{ffnn}
An ANN finds the optimal values for $\Theta$ by taking forward and a backward passes through the computational graph. In the forward pass, information flows from the input $x$ to the estimate $\widehat{y}$ according to the operations specified in Figure \ref{fig: ffnn}. During the backward pass, the error between $\widehat{y}$ and the target is first computed
\begin{gather}
\label{loss}
    E =  L(y, \widehat{y})
\end{gather}
Here $L$ is a generic convex and differentiable function measuring the distance between $y$ and $\widehat{y}$. Then, the gradient of the error with respect to all the parameters is found and an update is performed taking steps of size $\alpha \in [0, 1]$ in the direction opposite to the gradient
\begin{gather}
\label{delta_rule}
    \Delta w^{j}_{i} = -\alpha\frac{\partial E}{\partial w^{j}_{i}} \\
    w^{j}_{i} \leftarrow w^{j}_{i} + \Delta w^{j}_{i} \nonumber
\end{gather}
What we illustrated here is the application of the delta rule for updating the $i^{th}$ parameter of the $j^{th}$ layer through gradient descent \citep{widrow1960adaptive}. Deep feedforward ANNs rely on a generalization of this rule (i.e. backpropagation \citep{rumelhart1986learning}) for efficiently computing the gradient for all the parameters in the network.  
\\
\\
Returning to the supervised learning problem specified in section \ref{td_to_supervised}, a feedforward ANN approximates $V(s_{t})$ by mapping the inputs of equation \ref{supervised_b} to a candidate $\widehat{V}(s_{t})$ which is then used to generate an estimate $\widehat{B}_{t+1}$. Then, during the backward pass $\widehat{V}(s_{t})$ is adjusted based on the degree of mismatch between the estimation that it produced and the real value of $B_{t+1}$. It is of interest to note that there is a certain degree of overlap between how ANNs adjust their weights and the TD update illustrated in section \ref{td_learning}. Indeed, in single-step scenarios (i.e. predicting $s_{t+1}$ based on $s_{t}$ for each $s \in S$) the parameter changes produced by the two methods are the same \citep{sutton1988learning}. The major difference lies in the delivery of the update: TD learning performs it at every step while backpropagation-based algorithms must  wait until the end of the sequence in order to collate all the observed errors in a single term \citep{sutton1988learning}.

\paragraph{\textbf{Recurrent Neural Networks}}
\label{rnn_theory}
Despite their universal function approximation properties \citep{hornik1989multilayer}, feedforward ANNs are not suitable for the type of recursive operations expressed in paragraph \ref{td_to_supervised} \citep{bengio2017deep}. As we can see from Figure \ref{fig: ffnn_rnn}A, given a sequence of inputs and targets, a conventional feedforward ANN would be limited to learning a temporally local function of the form
\begin{gather}
\label{td_ffnn}
    B_{t+1} = f^2(f^1(O, B_{t}; \theta^{1}); \theta^{2})
\end{gather}
Even when $\Theta$ are shared across time, the estimated $\widehat{V}(s_t)$ cannot incorporate information from past $\widehat{V}(s)$ nor guarantee predictive power for the future $B$. A solution to this problem is offered by ANNs with feedback connections like Recurrent Neural Networks (RNNs). These  are a class of ANNs that are able to efficiently process long sequences of data while also relaxing the requirements of conventional feedforward ANNs for fixed length inputs \citep{bengio2017deep}. Looking at Figure \ref{fig: ffnn_rnn}B, we see that for each $t \in T$ a RNN would compute $\widehat{V}(s_t)$ using both the input $OB_{t}$ and the previously estimated representation $\widehat{V}(s_{t-1})$. This, in combination with the recursive application of $\Theta$, allows the network to learn a function over the entire temporal sequence and to provide $\widehat{V}(s_t)$ with the desirable properties mentioned in section \ref{td_to_supervised}. The structure of $\Theta$ is more complex in RNNs than in feedforward ANNs \footnote{See \citep{bengio2017deep} for a description of the parameters' structure in RNNs.} and a detailed derivation of the underlying optimization process is outside the scope of the present work. Nevertheless, it is worth singling out how the recurrent nature of the computations underlying the generation of $\widehat{V}(s_t)$  makes RNNs suitable for approximating the function specified in section \ref{td_to_supervised}. \\
\\
Following Figure \ref{fig: ffnn_rnn}B, let $\widehat{V}(s_t)$ be the representation inferred by the model at time $t$ and its associated parameters. Optimal parameter values are found through the same update rule used in feedforward ANNs
\begin{gather}
\label{bptt_1}
    \widehat{V}(s_t) \leftarrow \widehat{V}(s_t) + -\alpha \frac{\partial E}{\partial \widehat{V}(s_t)}
\end{gather}
however, since $E$ can now only be observed at the end of a temporal sequence, computing $\frac{\partial E}{\partial \widehat{V}(s_t)}$ requires us to take into account all the intermediate steps from $t$ to $T$. This is achieved applying the chain rule and propagating the error gradient backward in time \citep{bengio2017deep,lillicrap2019backpropagation}
\begin{gather}
\label{bptt_2}
    \frac{\partial E}{\partial \widehat{V}(s_t)} = 
    \frac{\partial E}{\partial \widehat{V}(s_{T})}
    \frac{\partial \widehat{V}(s_{T})}{\partial \widehat{V}(s_{t+1})}
    \frac{\partial \widehat{V}(s_{t+1})}{\partial \widehat{V}(s_{t})}
\end{gather}
This implies that, similarly to TD update, the error flow forces $\widehat{V}(s_t)$ to retain information from $OB_t$ and $\widehat{V}(s_{t-1})$ in order to perform estimation of $B_{t+1}$ while still being useful for generating $\widehat{V}(s_{t+1})$ as we can see from Figure \ref{fig: ffnn_rnn}B. This process is made more efficient by an RNN variant called Long Short-Term Memory (LSTM) \citep{hochreiter1997long}, which, as well as improving the propagation of the error gradient, has specialized mechanisms for inferring, at each point in time, which portion of information should be kept or discarded in order to minimize $E$ \citep{hochreiter1997long,bengio2017deep}.
\input{ffnn_rnn}

\subsection{Representation and Manifold Learning}
\label{manifold_learning}
\begin{figure}[h]
  \begin{center}
    \begin{adjustbox}{width=0.6\columnwidth}
        \tikzset{every picture/.style={line width=0.75pt}} %set default line width to 0.75pt        
        
        \begin{tikzpicture}[x=0.75pt,y=0.75pt,yscale=-1,xscale=1]
        %uncomment if require: \path (0,300); %set diagram left start at 0, and has height of 300
        
        %Straight Lines [id:da4267952151606098] 
        \draw    (319.82,122) -- (320.2,191.88) ;
        \draw [shift={(319.8,119)}, rotate = 89.69] [fill={rgb, 255:red, 0; green, 0; blue, 0 }  ][line width=0.08]  [draw opacity=0] (8.93,-4.29) -- (0,0) -- (8.93,4.29) -- (5.93,0) -- cycle    ;
        %Straight Lines [id:da6989378371909234] 
        \draw [color={rgb, 255:red, 208; green, 2; blue, 27 }  ,draw opacity=1 ]   (310.17,188.88) -- (309.53,119) ;
        \draw [shift={(310.2,191.88)}, rotate = 269.48] [fill={rgb, 255:red, 208; green, 2; blue, 27 }  ,fill opacity=1 ][line width=0.08]  [draw opacity=0] (8.93,-4.29) -- (0,0) -- (8.93,4.29) -- (5.93,0) -- cycle    ;
        %Shape: Ellipse [id:dp8707618719305075] 
        \draw   (353.08,94.91) .. controls (353.08,81.53) and (363.7,70.7) .. (376.81,70.7) .. controls (389.91,70.7) and (400.53,81.53) .. (400.53,94.91) .. controls (400.53,108.28) and (389.91,119.12) .. (376.81,119.12) .. controls (363.7,119.12) and (353.08,108.28) .. (353.08,94.91) -- cycle ;
        %Straight Lines [id:da13933094112344846] 
        \draw [color={rgb, 255:red, 208; green, 2; blue, 27 }  ,draw opacity=1 ]   (341.1,212.65) -- (376.81,119.12) ;
        \draw [shift={(340.03,215.46)}, rotate = 290.89] [fill={rgb, 255:red, 208; green, 2; blue, 27 }  ,fill opacity=1 ][line width=0.08]  [draw opacity=0] (8.93,-4.29) -- (0,0) -- (8.93,4.29) -- (5.93,0) -- cycle    ;
        %Shape: Ellipse [id:dp9097319337144372] 
        \draw   (291.56,95.2) .. controls (291.56,81.82) and (302.18,70.99) .. (315.29,70.99) .. controls (328.39,70.99) and (339.01,81.82) .. (339.01,95.2) .. controls (339.01,108.57) and (328.39,119.41) .. (315.29,119.41) .. controls (302.18,119.41) and (291.56,108.57) .. (291.56,95.2) -- cycle ;
        %Straight Lines [id:da3769355222150532] 
        \draw [color={rgb, 255:red, 208; green, 2; blue, 27 }  ,draw opacity=1 ]   (291.47,212.67) -- (254.35,119.55) ;
        \draw [shift={(292.58,215.46)}, rotate = 248.27] [fill={rgb, 255:red, 208; green, 2; blue, 27 }  ,fill opacity=1 ][line width=0.08]  [draw opacity=0] (8.93,-4.29) -- (0,0) -- (8.93,4.29) -- (5.93,0) -- cycle    ;
        %Straight Lines [id:da1253267073766362] 
        \draw [color={rgb, 255:red, 0; green, 0; blue, 0 }  ,draw opacity=1 ]   (263.71,120.96) -- (296.02,202.56) ;
        \draw [shift={(262.61,118.17)}, rotate = 68.4] [fill={rgb, 255:red, 0; green, 0; blue, 0 }  ,fill opacity=1 ][line width=0.08]  [draw opacity=0] (8.93,-4.29) -- (0,0) -- (8.93,4.29) -- (5.93,0) -- cycle    ;
        %Shape: Ellipse [id:dp7350394708339132] 
        \draw   (292.58,215.46) .. controls (292.58,202.09) and (303.2,191.25) .. (316.3,191.25) .. controls (329.41,191.25) and (340.03,202.09) .. (340.03,215.46) .. controls (340.03,228.83) and (329.41,239.67) .. (316.3,239.67) .. controls (303.2,239.67) and (292.58,228.83) .. (292.58,215.46) -- cycle ;
        %Shape: Ellipse [id:dp34753658855295044] 
        \draw   (230.62,95.34) .. controls (230.62,81.97) and (241.24,71.13) .. (254.35,71.13) .. controls (267.45,71.13) and (278.07,81.97) .. (278.07,95.34) .. controls (278.07,108.71) and (267.45,119.55) .. (254.35,119.55) .. controls (241.24,119.55) and (230.62,108.71) .. (230.62,95.34) -- cycle ;
        %Straight Lines [id:da6916637423351373] 
        \draw [color={rgb, 255:red, 0; green, 0; blue, 0 }  ,draw opacity=1 ]   (367.15,120.44) -- (336.7,202.13) ;
        \draw [shift={(368.2,117.63)}, rotate = 110.44] [fill={rgb, 255:red, 0; green, 0; blue, 0 }  ,fill opacity=1 ][line width=0.08]  [draw opacity=0] (8.93,-4.29) -- (0,0) -- (8.93,4.29) -- (5.93,0) -- cycle    ;
        
        % Text Node
        \draw (316.33,214.5) node   [align=left] {$\displaystyle \widehat{V}( s_{t})$};
        % Text Node
        \draw (254.43,94.4) node   [align=left] {$\displaystyle B^{1}_{t+1}$};
        % Text Node
        \draw (315.37,94.25) node   [align=left] {$\displaystyle B^{2}_{t+1}$};
        % Text Node
        \draw (377.91,95.99) node   [align=left] {$\displaystyle B^{n}_{t+1}$};
        % Text Node
        \draw (171.67,200.4) node [anchor=north west][inner sep=0.75pt]  [font=\LARGE]  {$h$};
        % Text Node
        \draw (172,80.4) node [anchor=north west][inner sep=0.75pt]  [font=\LARGE]  {$y$};
        
        \end{tikzpicture}
    \end{adjustbox}
\end{center}
\caption{\textbf{Multi-task learning in an ANN}. Adapted from \cite{bengio2017deep}. The figure represents how multi-task learning could be used in an ANN to force the the latent representation $h$ to be a sensible approximation of $V(s_t)$. Here $\widehat{V}(s_t)$ indicates the representation generated by a recurrent layer at time $t$ while $B_{t+1}=\{B^n_{t+1}: n \in N\}$ are $N$ targets quantifying the strength of the next interaction (in terms of frequency and amount of behaviour)  between $I$ and $O$. Black and red arrows are respectively the direction of the computations and the flow of the error gradient. Circles indicate computational blocks similar to those in figures \ref{fig: ffnn} and \ref{fig: ffnn_rnn}.}
\label{fig: multi_task}
\end{figure}
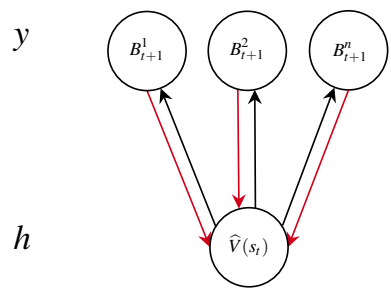

As mentioned in the previous sections, ANNs are tasked to create latent representations (e.g. $V(s_{t})$) which are not explicitly defined by their input or target but are nevertheless functional for connecting the two \citep{rumelhart1986learning,bengio2017deep,lillicrap2020backpropagation}. This is based on the hypothesis that the relationship between the input and the target can be expressed in terms of variations in coordinates on a manifold \citep{bengio2017deep}. In the lower dimensional space of this manifold, the input is re-organized to improve estimation and elements which are similar to each other tend to appear close together \citep{bengio2017deep}. In this view, during optimization, each layer of an ANN attempts to place its input on a manifold that is useful for the layer that follows. This process continues until the last layer. Here the inputs are organized in such way that it makes easier for the network to produce good predictions of the target \citep{bengio2017deep}. Moving along this final manifold allows one to reach inputs with different characteristics leading to variations in the predictions produced by the model. We hypothesize that the amount of attributed incentive salience (i.e. $V(s_{t})$) can be modeled as a manifold on which the history of individual-object interactions is placed in order to best predict the intensity of all future interactions. This relates to the concept of motivation as a vector presented in sections \ref{motivation} and \ref{manifold_state}: the representation $V(s_{t})$ estimated by an ANN can be thought of as a vector in an $h$ dimensional space, where $h$ is the number of units of the layer producing the representation, indicating the amount of attributed incentive salience after observing $t$ interactions. As we can see, differently from completely un-supervised approaches this approach forces the learned manifold to obey to specific representational and predictive functionalities that are shared with the construct of attributed incentive salience. Given the potentially large number of layers in an ANN, locating this representation and most importantly ensuring that it is a suitable approximation of $V(s_{t})$ are potential issues. A possible solution is to impose a form of architectural constraint on the optimization process through multi-task learning. Multi-task learning closely resemble multivariate analysis, it  works on the assumption that a common latent factor underlying a set of targets exists and it can be constrained in a single representation used by the ANN for producing multiple predictions \citep{bengio2017deep}. An example of this process is shown in figure \ref{fig: multi_task}. As mentioned in section \ref{incentive_salience}, the amount of attributed incentive salience $V(s_t)$ that an individual $I$ assigns to an object $O$ should be a latent factor that indicates how intense future interactions with that object will be. Therefore, if a layer in an ANN is forced to produce a single representation which is then used to estimate multiple behavioural indicators of the intensity of these interactions, this should provide a sensible approximation of the amount of attributed incentive salience. 

\section{Methodology}
\label{method}
We hypothesize that the amount of attributed incentive salience could be approximated by the representation produced by a model taking as input the intensity of all the interactions that an individual had with an object while estimating the intensity of the one that follows. We expect ANNs, and their recurrent variant in particular, to be well suited to this task and to provide better fits to the data. As mentioned in sections \ref{motivation} and \ref{incentive_salience}, in order for this representation to effectively describe the amount of attributed incentive salience it must distinguish between different objects while also maintaining the ability to differentiate individuals based on the expected intensity of their future interactions. It should also maintain this property over time and capture potential variations occurring in the history of interactions between the individual and the object.\\
\\
Below we detail the methodological approach used to test these hypotheses, highlighting the connection between the theoretical and computational frameworks presented (see sections \ref{theoretical_framework} and \ref{comp_framework}), the data, the models, and the experimental pipeline employed. We fit a series of models to behavioural data coming from videogames. These models aim to predict the ammount, length and frequency of future playing behaviour that users have with a specific game based on the intensity of their previous interactions. We then analyzed the representation generated by one of these models, a RNN of our design, in order to evaluate its functional similarity with the construct of attributed incentive salience.\\
\\
\subsection{Data}
\label{data}
To validate our approach and hypotheses we needed to acquire records of interactions between individuals and potentially rewarding objects in naturalistic contexts. As mentioned in section \ref{videogame_telemetries}, video games are particularly suited for this purpose given their learning-dependent reinforcing properties and the large amount of longitudinal data streams that they can generate. We used gameplay data from  six video games published by our partner company, \textit{Square Enix Ltd.}. The games were \emph{Hitman Go} (hmg), \emph{Hitman Sniper} (hms), \emph{Just Cause 3} (jc3), \emph{Just Cause 4} (jc4), \emph{Life is Strange} (lis), and \emph{Life is Strange: Before the Storm} (lisbf). Due to the diversity in their in-game mechanics, each of these games was considered as an "object" with different reinforcing properties (see section \ref{videogame_telemetries}). This allowed us to mimic a situation where a single model had access to data coming from a heterogeneous set of potentially rewarding entities (similarly to what we described in section \ref{motivation}). The resulting dataset contained entries from 3,209,336 individuals, evenly distributed across the six games, and randomly sampled from all users who played the games between their respective release date and January 2020. All data were obtained and processed in compliance with the European Union's General Data Protection Regulation \citep{EUdataregulations2018}. In order to represent state transition dynamics (i.e. sequences of interactions between $I$ and $O$) for each individual, we retrieved a set of six different types of behavioural telemetry over variable-length sequences of game sessions. A game session was defined from the moment an individual started the game software until it was closed. We retrieved all sessions produced by an individual from the moment the data they generated first appeared in the game's servers. Since our modelling approach requires to predict, in a supervised manner, the intensity of future playing behaviour given the history of previous interactions, we only considered users with two or more observed game sessions. The reason for this is two fold: sequences of length one do not entail any temporal structure and do not allow to generate a supervised target.
\begin{table}[H]
\caption{Description of Selected Telemetries}
\label{metricsdescription}
  \resizebox{\columnwidth}{!}{
  \begin{tabular}{@{}ll@{}}
    \toprule
    \textbf{Metric}      & \textbf{Description}          \\ \midrule
    {Absence}    & Temporal distance between sessions (hours)  \\
    {Session Time}     & Overall session duration (minutes)       \\ 
    {Active Time}      & Percentage of Session Time actively playing  \\ 
    {Session Activity}    & Count of user initiated gameplay-related actions. E.g.\\ 
                & "Attack an enemy" is considered a valid\\ 
                & action while "Being attacked by an enemy" is not.\\
    {N°Sessions}    & Number of played sessions.\\ 
    {Object}    &  Game object identifier.  \\
    \bottomrule
  \end{tabular}
  }
\end{table}
The telemetry (see Table \ref{metricsdescription}) were selected to be generalizable and comparable with metrics employed in other behavioural studies of incentive salience attribution \citep{berridge1998role,mcclure2003computational,zhang2009neural}. We note that the high dispersion values (Inter Quartile Range  or IQR), reported for some of the telemetry are due to the extreme skewness in the distribution of the data. This is caused both by the nature of the phenomenon they describe (e.g. Absence is a classic case of time-to-event measure) and by their typical behaviour in the context of videogames \citep{bauckhage2012players}. The final dataset was composed of 6 columns and 28,155,199 rows. A table of descriptive statistics can be found in \ref{game_description}.

\begin{table*}[h]
\centering
\caption{Descriptive Statistics of Considered Metrics and Games}
\label{game_description}
  \begin{tabular}{cccccccc}
  \toprule
  \multirow{2}{*}{\textbf{Game}} &
   \multirow{2}{*}{\textbf{\begin{tabular}[c]{@{}c@{}}Sample \\ Size\end{tabular}}} &
   \textbf{\begin{tabular}[c]{@{}c@{}}Number \\ of \\ Sessions\end{tabular}} &
   \textbf{\begin{tabular}[c]{@{}c@{}}Absence \\ (minutes)\end{tabular}} &
   \textbf{\begin{tabular}[c]{@{}c@{}}Session \\ Time\\ (minutes)\end{tabular}} &
   \textbf{\begin{tabular}[c]{@{}c@{}}Active \\ Time\\ (\% Session Time)\end{tabular}} &
   \textbf{\begin{tabular}[c]{@{}c@{}}Session \\ Activity\end{tabular}} &
   \multirow{2}{*}{\textbf{\begin{tabular}[c]{@{}c@{}}Game \\ Description\end{tabular}}} \\ \midrule
   &
    &
   \multicolumn{5}{c}{\textbf{\begin{tabular}[c]{@{}c@{}}(Median $\pm$ IQR)\end{tabular}}} &
    \\ \midrule
  \textbf{hmg} &
   501,649 &
   \begin{tabular}[c]{@{}c@{}}3 $\pm$ 3\end{tabular} &
   \begin{tabular}[c]{@{}c@{}}84$\pm$ 2,169\end{tabular} &
   \begin{tabular}[c]{@{}c@{}}22 $\pm$ 22 \end{tabular} &
   \begin{tabular}[c]{@{}c@{}}64 $\pm$  42 \end{tabular} &
   \begin{tabular}[c]{@{}c@{}}25 $\pm$  31\end{tabular} &
   \begin{tabular}[c]{@{}c@{}}Mobile\\ Strategy\end{tabular} \\
  \textbf{hms} &
   504,504 &
   \begin{tabular}[c]{@{}c@{}}8 $\pm$ 9\end{tabular} &
   \begin{tabular}[c]{@{}c@{}}24 $\pm$ 198\end{tabular} &
   \begin{tabular}[c]{@{}c@{}}28 $\pm$ 8\end{tabular} &
   \begin{tabular}[c]{@{}c@{}}42 $\pm$ 35\end{tabular} &
   \begin{tabular}[c]{@{}c@{}}6 $\pm$ 8 \end{tabular} &
   \begin{tabular}[c]{@{}c@{}}Mobile\\ Shooting Gallery\end{tabular} \\
  \textbf{jc3} &
   540,000 &
   \begin{tabular}[c]{@{}c@{}}7 $\pm$ 8\end{tabular} &
   \begin{tabular}[c]{@{}c@{}}64 $\pm$ 488\end{tabular} &
   \begin{tabular}[c]{@{}c@{}}162 $\pm$ 23\end{tabular} &
   \begin{tabular}[c]{@{}c@{}}60 $\pm$ 55\end{tabular} &
   \begin{tabular}[c]{@{}c@{}}19 $\pm$ 23\end{tabular} &
   \begin{tabular}[c]{@{}c@{}}Console\\ Action Open World\end{tabular} \\
  \textbf{jc4} &
   571,501 &
   \begin{tabular}[c]{@{}c@{}}5 $\pm$ 6 \end{tabular} &
   \begin{tabular}[c]{@{}c@{}}64 $\pm$ 406\end{tabular} &
   \begin{tabular}[c]{@{}c@{}}133 $\pm$ 64\end{tabular} &
   \begin{tabular}[c]{@{}c@{}}43 $\pm$ 46\end{tabular} &
   \begin{tabular}[c]{@{}c@{}}46 $\pm$ 64\end{tabular} &
   \begin{tabular}[c]{@{}c@{}}Console\\ Action Open World\end{tabular} \\
  \textbf{lis} &
   533,364 &
   \begin{tabular}[c]{@{}c@{}}4 $\pm$ 4\end{tabular} &
   \begin{tabular}[c]{@{}c@{}}143  $\pm$ 3,004\end{tabular} &
   \begin{tabular}[c]{@{}c@{}}96 $\pm$ 50\end{tabular} &
   \begin{tabular}[c]{@{}c@{}}48 $\pm$ 44\end{tabular} &
   \begin{tabular}[c]{@{}c@{}}40 $\pm$ 50\end{tabular} &
   \begin{tabular}[c]{@{}c@{}}Console\\ Graphic Adventure\end{tabular} \\
  \textbf{lisbf} &
   517,782 &
   \begin{tabular}[c]{@{}c@{}}4 $\pm$ 5\end{tabular} &
   \begin{tabular}[c]{@{}c@{}}71 $\pm$ 1,162\end{tabular} &
   \begin{tabular}[c]{@{}c@{}}102 $\pm$ 32\end{tabular} &
   \begin{tabular}[c]{@{}c@{}}79 $\pm$ 20\end{tabular} &
   \begin{tabular}[c]{@{}c@{}}23 $\pm$ 32\end{tabular} &
   \begin{tabular}[c]{@{}c@{}}Console\\ Graphic Adventure\end{tabular} \\ \bottomrule
  \end{tabular}
\end{table*}

\subsection{Models} 
\label{models}
When defining the models used for evaluating our hypotheses, we first established two reasonable single-parameter baselines. The first is a \textit{Lag 1} model producing predictions according to the following rule:
\begin{equation}
   \begin{gathered}  
     B_{t+1} = B_{t}
     \label{lag_1}
  \end{gathered}
\end{equation}
here $t$ represent a single game session in a sequence of $T$ observed interactions while $B$ are the behavioural metrics described in section \ref{data} except for N°Sessions, for which the model provides a constant prediction of 1. Indeed, the lag-1 version of N°Sessions is not a realistic prediction as it linearly increases with $t$. The second is a \textit{Median} model computing the expectancy of each of the 4 targets according to the formula:
\begin{equation}
  \begin{gathered}  
    \overline{B_{t+1}} = \frac
      {\sum_{i=1}^{t+1} wB_{i}}
      {\sum_{i=1}^{t+1} w }\\
    B_{t+1} = median(\overline{B_{t+1}}) 
    \label{median}
  \end{gathered}
\end{equation}
here $\overline{B_{t+1}}$ is an exponentially weighted average of all the $B_t$ up to $t+1$ observed when fitting the model. This is computed separately for every individual in the dataset and the median value of each of the 5 targets is used a a constant prediction. These apparently naive models provide a surprisingly robust prediction baseline for time series that are not white noise \citep{hyndman2018forecasting} other than having a nice interpretation in terms of behavioural momentum \citep{nevin2000behavioral}: in conditions of high experienced reward the behaviour of an individual tends to be consistent over time (i.e. resistant to change). An ElasticNet (\textit{E-Net}) \citep{zou2005regularization}, a form of additive model combining both $l1$ and $l2$ regularization, was used to evaluate the performance of simple linear functions. To test the effect of non linearity, a Multilayer Perceptron (\textit{MLP}), the most common type of deep feedforward ANN, was used. Finally to evaluate the effect of recurrency in combination with non-linearity, we designed a hybrid approach (\textit{RNN}) integrating recurrent and feedforward operations in a single ANN. Despite being markedly different, this last architecture shares similarities with the technique proposed by Calhoun et. al. \citep{calhoun2019unsupervised} where latent states were generated dynamically and employed by Generalized Linear Models for producing predictions of observed behaviours. A representation of the computational graphs constituting the parametric models can be seen in Figure \ref{fig: rnn}. It should be noted that E-Net is architecturally equivalent to MLP with the only difference being the replacement of stacked feedforward layers with a single layer with linear activation. As illustrated in Figure \ref{fig: ffnn_rnn} the inputs to each model were sequences of the same behavioural metrics reported in Table \ref{metricsdescription} plus the associated game object while the targets were simply the lead 1 version of the same sequences. We want to highlight how each model was fit to sequences coming from the same game object and that the predictions only pertained the behavioural metrics. Indeed, the aim of each model was to solve
\begin{equation}
    \begin{gathered}
        E(B_{t+1} | \{B_{t}, B_{t-1}, \dots \} , O)
        \label{model_obj}
    \end{gathered}
\end{equation}
jointly for each $O$ present in the data. All the models adopted a multi-task approach (see section \ref{manifold_learning}) and were designed to perform estimation in a sequence-to-sequence fashion: given an input sequence of length $T$ its lead-1 version was predicted. The MLP and RNN models both used a $ReLU$ activation function. The function has shape $ReLU(x) = max(0, x)$ with $x$ being the value computed by a single hidden unit. All the models, except for \textit{Median}, were implemented using Tensorflow's high level API "Keras" \citep{tensorflow2015-whitepaper,chollet2015keras}. The $Median$ model was implemented using the libraries for scientific computing Pandas and Numpy \citep{reback2020pandas,harris2020array}.

\input{architectures}

\subsection{Experimental Procedure}
\label{procedure}
Our experimental pipeline is shown in Figure \ref{fig: pipeline}
\input{pipeline}

\paragraph{\textbf{Data Preparation}} 
\label{data_preparation}
When querying the data from the game servers, we excluded from the random sampling procedure individuals having at least one of the considered behavioural metrics over the game population's \nth{99} percentile. This allowed us to eliminate potentially faulty data which are often present when dealing with telemetry. At this point data were re-arranged in a format suitable for time series modelling (i.e. arrays of shape $(batch\times T \times B)$ with $T$ being the number of available game sessions and $B$ the number of considered behavioural metrics) and randomly split into a tuning (i.e. 10 \%) and validation set (i.e. 90 \%). For the sake of clarity we report an example of how the data from a game session are generated and how they are parsed by the models.\\
\\
\textit{
"A user decides, 36 hours after the release of game X, to enter the game world for the first time. This is when a session starts and actual playing behaviour can be observed. During this session they engage in various activities leading to 20 non-unique and user-initiated actions (e.g. being attacked by a non-playable character is not counted as a valid action). After roughly 60 minutes spent playing, the user exits the game world and the session ends. Of this session, 80\% of time has been spent actively playing, the remaining 20\% has seen the game on pause or the user away from the console (i.e. idle time). After 48 hours the user logs into the game world again and a new session starts"}\\
\\
What we described here would correspond to a single time step $t_{1}$ in the sequence $T$ of total interactions (i.e. sessions) between the user and the specific game context X. The models will parse this session as a vector of length 4 with values 36, 20, 60, 80 and 20 along with another vector of length 1 containing the numerical index for the game. When all the sessions are observed the models will receive as inputs sequences of length $T$ of the same vectors. This implies that that each model is fitted on 4 continuous and 1 discrete (i.e. the game context) metric. The median and lag-1 models do not explicitly make use of the categorical variable as they are "fit" on separate partitions of the data (i.e. one for each game context). The target of each model is then constituted of 5 variables: the lead-one version of the aformetioned 4 continuous variables plus the total number of future sessions observed at time $t$.

\paragraph{\textbf{Performance Analysis}} 
\label{perf_analysis}
The first step in our performance analysis aimed to control for the contribution of hyper-parameters in the performance of the parametric models (especially for MLP and RNN). The choice of factors such as the number of layers and hidden units can influence the number of free parameters and expressive power of an ANN. Manually picking their optimal value is often a challenging combinatorial problem that can lead to unexpected outcomes if left to the subjective choice of the experimenter. Therefore, to find the best hyperparameters we adopted a more impartial and efficient approach relying on algorithmic search. This was done using the Keras Tuner implementation \citep{omalley2019kerastuner} of the Hyperband algorithm \citep{li2017hyperband}. Hyperband is an optimized version of random search that achieves faster convergence through adaptive resources allocation and early termination of training. It can lead to better or equivalent results to other optimization algorithms but in a fraction of the time \citep{li2017hyperband}. When initializing the tuning step we allowed each model to grow as much as the others (except for E-Net, which,  due to the fact that it is a linear model, is naturally constrained to a fixed number of parameters) so that any observed difference in number of parameters was related to characteristics of the model architecture. The tuning step was conducted running one full iteration of Hyperband with a budget of 40 epochs \footnote{See  \citep{li2017hyperband,hyperwebs} for a detailed description of the Hyperband technique.} on the tuning set. To trigger early stopping for a specific configuration of hyperparameters, we monitored the decrease in loss over a 20\% random sample of the tuning set (i.e. the validation tuning set) and we terminated training when the loss reduced by less than $\delta = 1\mathrm{e}{-4}$ for 10 consecutive epochs. Once the best set of hyperparameters was found we proceeded to fit all the models specified in section \ref{models} on the training set using a 10-fold Cross Validation Strategy. This  divided the validation set in 10 equally sized folds and iteratively used 9 of them for training and 1 for testing. The continuous inputs in the training data were min-max scaled according to the formula
\begin{equation}
  \begin{gathered} 
  \label{min_max}
        MinMax(x) =\frac{x - \min(x)} {\max(x) - \min(x)} 
  \end{gathered}
\end{equation}
where $x$ is the input vector to be scaled, while the categorical input (i.e. game object) was encoded ordinally. In order to take into account the contributions of time, game and target, the performance of each model was given by computing the Symmetric Mean Percentage Error (SMAPE) \citep{zhu2017deep} for each combination of the aforementioned dimensions (e.g. SMAPE of Session Time at $t1$ for the game object hmg). Each model was trained for a maximum of 200 epochs and interrupted using the same early stopping strategy mentioned above (i.e. absence of $\delta$ reduction in loss on a 20\% hold-out for 10 consecutive epochs). To maintain the ability to fit each model on temporal series of varying length, we adopted a data generator approach \footnote{See \citep{chollet2015keras,tensorflow2015-whitepaper} for implementation details.} feeding data to the models in random batches of 256 time series with constant length within a batch. The models were trained with stochastic gradient descent using the Adaptive Moment Estimation (Adam) \citep{kingma2014adam} algorithm to find the set of parameters minimizing the SMAPE between the targets and the predictions generated by the model. 
\begin{equation}
  \begin{gathered} 
  \label{smape}
     SMAPE(y, \widehat{y}) = 
    \frac{1}{N} 
    \sum_{i=1}^{N}
    \frac{| y_{i} - \widehat{y}_{i} |} {|y_{i}| + |\widehat{y_{i}}|}  
  \end{gathered}
\end{equation}
here $y$ and $\widehat{y}$ are respectively the ground truth and the estimate produced by the model while $N$ indicates the size of the batch. The SMAPE is bounded between 0 and 100 and can be interpreted as percentage deviation from the target with lower values indicating better model fit. The choice of SMAPE was dictated by the fact that the targets were expressed on largely different scales (i.e. coming from different games and expressed on different units of measure see Table \ref{metricsdescription}) and therefore required a loss function measuring relative distance from the target. To evaluate the overall performance, we first summed the SMAPE relative to each target in a single global performance indicator: this is the loss function that each model attempts to minimize during training. W then divided the total by 5 (i.e. the total number of targets) in order to express the metric in its original scale (i.e. 0 to 100). This was then regressed using a Linear Mixed-effects Model (LMM) with fold number, game object and time as random effects and model type as fixed effect (treatment coded with RNN as reference). Subsequently, for a more thorough investigation of model performance we conducted the same regression analysis separately on each target. Both regression analyses were followed by post-hoc comparisons (i.e. t-tests with Bonferroni correction) for testing the following pairwise hypotheses on the estimated coefficients: Lag 1 $<$ Median $<$ ENet $<$ MLP. All statistical analyses were conducted using the python library statsmodels \citep{seabold2010statsmodels}. 

\paragraph{\textbf{Representation Analysis}}
\label{representation_analysis}
After comparing the performance of the RNN model with that of alternative approaches, we proceeded to analyze the representation inferred by the two ANN, with particular attention to the one generated by the RNN. First, we re-fitted both models on a random sample (i.e. 90\%) of the validation-set following the same procedure specified in paragraph \ref{perf_analysis}. Then, we created an encoder composed of all the transformations and relative parameters leading to the shared-representation layer (red highlight in Figure \ref{fig: rnn}). As illustrated in paragraph \ref{manifold_learning}, this is the portion of the model that we expected to approximate the manifold structure of attributed incentive salience. Subsequently, we passed the remaining portion of the validation-set (i.e. 10\%) as an input to the encoders, producing arrays of shape $(batch\times T \times h)$ with $h$ being the number of units in the shared layer and $T$ the number of sessions observed for $batch$ number of individuals. In order to visualize and inspect this multidimensional representation, we used the Uniform Manifold Approximation and Projection (UMAP) algorithm \citep{2018arXivUMAP}, a dimensionality reduction  technique based on manifold learning. Given a high dimensional dataset, UMAP first infers its topological structure and then, using stochastic gradient descent, attempts to structurally reproduce it in a lower dimensional space (two or three for visualization purposes) \citep{2018arXivUMAP}. Compared to other similar dimensionality reduction approaches (for example, the t-distributed Stochastic Neighbor Embedding), UMAP tends to better preserve both global and local structure of the original data, meaning that distances in the underlying dataset should be more faithfully reproduced. Moreover, when given a sequence of datasets with entries related to each other, UMAP is able to maintain these relationships during the optimization process \footnote{See \citep{alignedumap} for implementation details.}. In our case these sequential datasets were the $T$ representations generated by the RNN model after observing $T$ games sessions for a group of individuals. Being able to take into account these temporal relationships allowed us to gather information not just on the characteristics of the representation produced by the RNN model but also on their evolution over time. To clarify, the encoder provided by the ANN was tasked to generate a multidimensional representation where distance represented similarity between individuals with respect to the intensity of their future interactions with a game (see the manifold hypothesis of attributed incentive salience in sections \ref{manifold_state} and \ref{manifold_learning}). The UMAP algorithm made this multidimensional representation interpretable to the human eye approximating it's manifold structure on a 2 dimensional plane 
and allowing us to evaluate the presence of those desirable properties that we mentioned at the beginning of section \ref{method}. Since we did not know the intrinsic dimensionality of the manifold we were trying to approximate, we conducted a Principal Component Analysis (PCA) of the representation generated by the RNN. Despite PCA and UMAP working under radically different assumptions and mechanisms, we thought this could provides us with a lower bound of how much variance we would be able to capture considering only two dimensions. The topological structure of the representation produced by the RNN was inferred by computing the cosine distance in a local neighborhood of 1000 points with a minimum distance of 0.8, while the dimensionality reduction was achieved by running the optimization part of the algorithm for 2000 iterations. The choice of a large neighborhood and minimum distance was made to better capture the global structure of the representation space \footnote{See \citep{umapwebs} for a visualization of the effects of these hyperparameters in UMAP.}.\\
\\
To understand the functions underlying the inferred representation, we conducted an exploratory investigation of the relationship between hidden units' activation in the recurrent layer and predictions produced by the model. To quantify the strength of the observed relationship we employed the Maximal Information Coefficient (MIC) \citep{reshef2011detecting}, a measure of mutual information that can quantify both linear and non-linear association between variables. The MIC can assume values between 0 to 1 with 1 corresponding to a perfect association. We adopted the implementation of UMAP provided McInnes \textit{et. al.} \citep{mcinnes2018umap-software} while the MIC was computed using the python library minepy \citep{albanese2013minerva}. Visualizations were produced using the python libraries matplotlib \citep{hunter2007matplotlib} and seaborn \citep{waskom2021seaborn}.

\paragraph{\textbf{Partition Analysis}}
\label{partition_analysis}
We conducted a partition analysis to individuate behavioural profiles associated with the representation generated by our model. As specified in section \ref{manifold_learning} the representation extracted by the encoder at time $t$ can be interpreted as a set of coordinates on the manifold generated by the RNN model after observing $t$ game sessions. Partitioning this representation allows us to identify areas of the manifold that hold information about the history of interactions between an individual and a video game object. These areas may represent variations in the levels of attributed incentive salience and therefore be associated with distinct patterns of behaviour. To partition the data, we used an unsupervised approach applying Mini-Batch K-Means \citep{sculley2010web}, a variation of K-Means, to the representation extracted by the encoder. Given a dataset, the algorithm attempts to divide it by iteratively moving $k$ centroids so as to reduce variance within each partition. The choice of Mini-Batch K-Means was dictated by the fact that it is one of the few distance-based algorithms that scales to very large datasets. To select the optimal $k$ value, we first fitted the algorithm with a varying number of centroids (i.e. 2 to 10) and computed the associated "inertia" (here, a measure of within cluster variance). Since inertia tends to zero as $k$ approaches the number of points in the dataset, we defined the optimal number of partitions as the value of $k$ at which the inertia reached its "elbow" or maximum curvature \citep{satopaa2011finding}. This allows to individuate at which number of partitions there are diminishing returns in terms of within cluster variance reduction. Every instance of Mini-Batch K-Means was initialized 3000 times at random and ran for a maximum of 3000 epochs. The input data were re-scaled to have zero mean and unit-variance and passed to the algorithm in random batches of size $(512 \times h)$. The associated behavioural profiles were found by applying this methodology separately to each game object and retrieving for each partition the mean of all the behavioural metrics over time. The Mini-Batch K-Means implementation used for this analysis was provided by the python library scikit-learn \citep{scikit-learn}. \\
\\
All the analyses were conducted using Python programming language version 3.6.2 \citep{10.5555/1593511}.

\section{Results}

\subsection{\textbf{Performance Analysis}}
\label{perf_results}
\begin{figure}[h]
\centering
\includegraphics[width=.8\columnwidth]{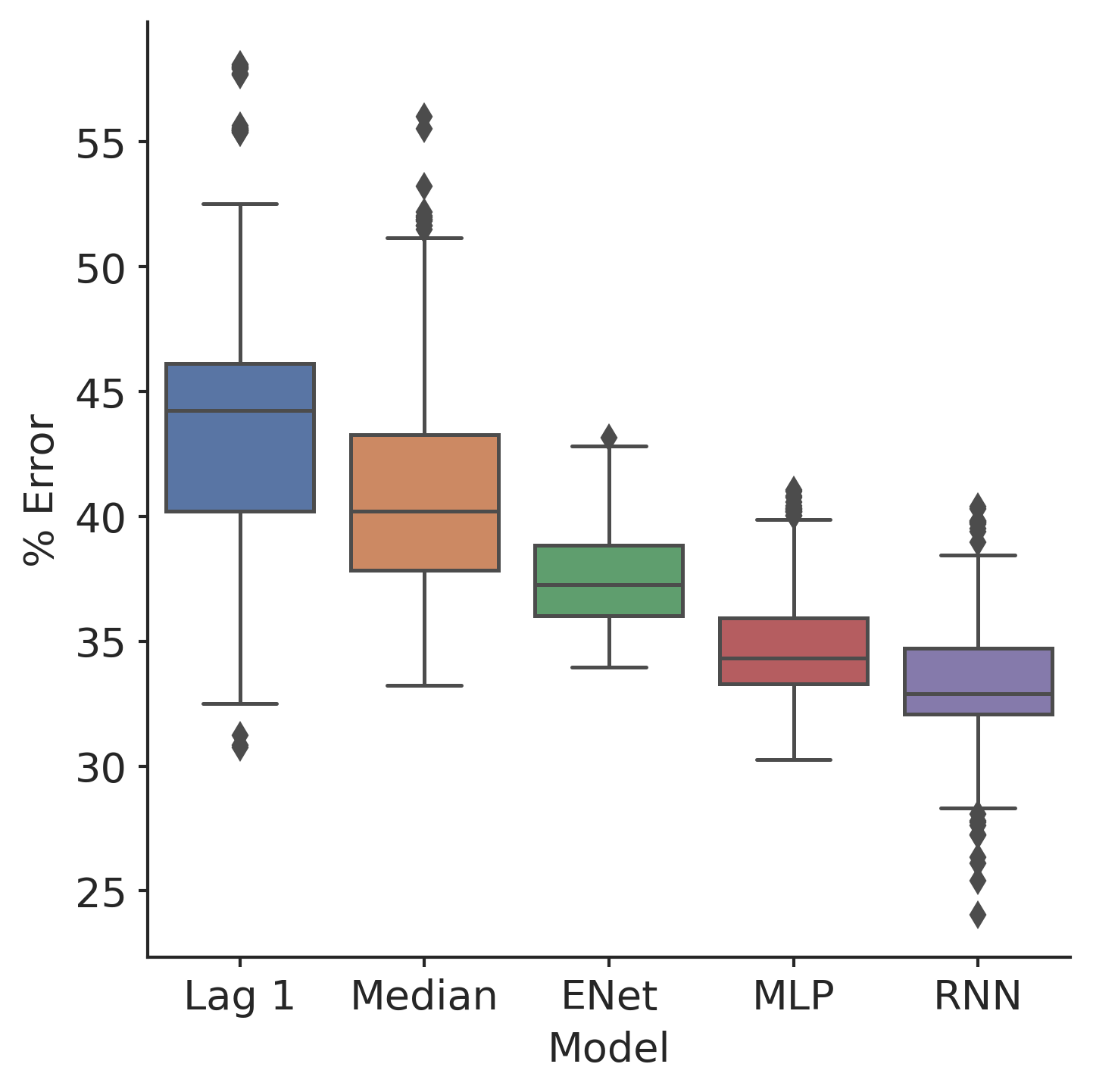}
\caption{\textbf{Aggregated comparison of model performance.} Overall, our approach (RNN) outperforms all the competing approaches. Box-plots show the 10-fold cross-validation performance expressed as the total percentage of error (i.e. SMAPE) of each model over the five targets.}
\label{model_comp_coll} 
\end{figure}

At the level of global performance the RNN model markedly outperformed all competing approaches as clearly shown in Figure \ref{model_comp_coll}. This can be also seen in the results of the regression (see Table \ref{collapsed_target_lmm}) and  \textit{post hoc} analysis (see Table \ref{collapsed_target_lmm_post_hoc}). From the \textit{post hoc} analysis we can also observe how all the pairwise hypotheses presented in paragraph \ref{perf_analysis} are confirmed. Here model performance is given by the sum of all the losses produced by the five targets and therefore provides a general indicator of model fit where lower values indicate a better performance overall.

\begin{table}[h]
\centering
\caption{Results of LMM on Collapsed Targets (Sum)}
\label{collapsed_target_lmm}
\begin{tabular}{ccccc}
\hline
\textbf{Model}           & \textbf{$\beta$} & \textbf{Z} & \textbf{p} & \textbf{95\% C.I.} \\ \hline
\multicolumn{5}{c}{\textbf{Collapsed Targets (Sum)}}                                                 \\ \hline
\textbf{Intercept (RNN)} & 33.129                & 41.799     & \textless .01   & 31.575 - 34.682      \\
\textbf{Lag 1}           & 10.010                 & 80.023     & \textless .01   & 9.765 - 10.255        \\
\textbf{Median}          & 7.482                 & 59.813     & \textless .01   & 7.237 - 7.727        \\
\textbf{ENet}            & 4.189                 & 33.491     & \textless .01   & 3.944 - 4.435        \\
\textbf{MLP}             & 1.411                 & 11.280     & \textless .01   & 1.166 - 1.656        \\ \hline
\end{tabular}
\end{table}
\begin{table}[h]
\centering
\caption{LMM Post-Hoc on Collapsed Targets (Sum)}
\label{collapsed_target_lmm_post_hoc}
\begin{tabular}{ccccc}
\hline
\textbf{Contrast}           & \textbf{$\beta_1$ - $\beta_2$} & \textbf{Z} & \textbf{p} & \textbf{95\% C.I.} \\ \hline
\multicolumn{5}{c}{\textbf{Collapsed Targets (Sum)}}                                                 \\ \hline
\textbf{Lag 1 - Median} & 2.528                & 20.210     & \textless .01   & 2.283 - 2.773      \\
\textbf{Median - ENet}          & 3.292                 & 26.322     & \textless .01   & 3.048 - 3.538        \\
\textbf{ENet - MLP}          & 2.778                 & 22.211     & \textless .01   & 2.533 - 3.024        \\ \hline
\end{tabular}
\end{table}

\begin{figure*}[h]
\centering
\includegraphics[width=.8\textwidth]{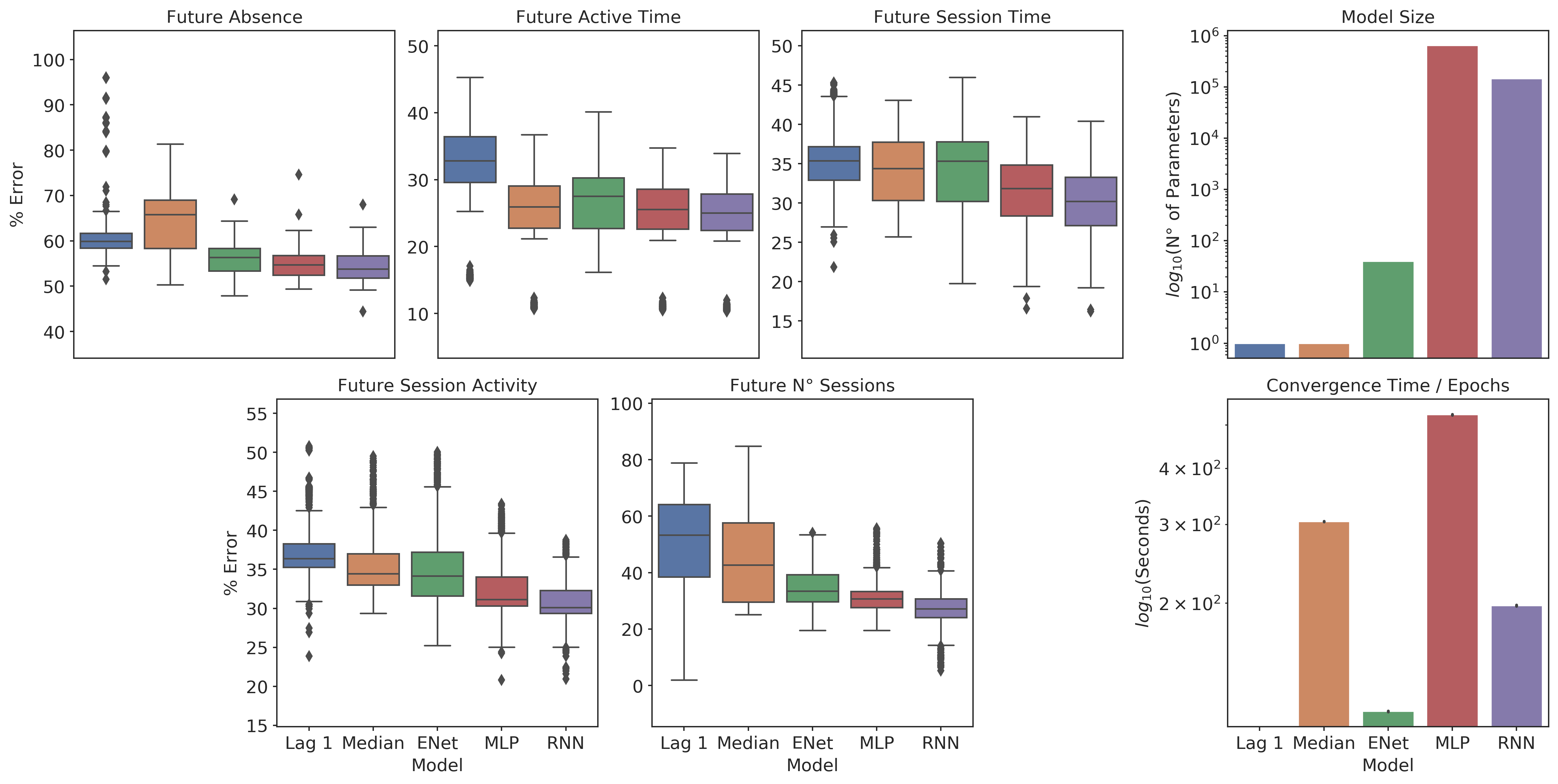}
\caption{\textbf{Dis-aggregated comparison of models' performance.}. Our approach (RNN) outperformed all competing ones on each target. It consistently used fewer parameters and had shorter computation time than the second best performing model. Box-plots show the 10-fold cross-validation performance expressed as percentage of error (i.e. SMAPE) of each model for the five targets. The bar-plot on the top row indicates the number of free parameters for each model while the bar plot on the bottom row shows the average time for each training epoch. Both bar-plots are $log_{10}$ scaled.}
\label{model_comp_non_coll} 
\end{figure*}

The superiority of the RNN model can still be observed when comparing the models on each target separately. However, the size of the effect varies depending on the target (see Table \ref{exploded_target_lmm}). The same trend is also present in the \textit{post hoc} analysis (see Table \ref{exploded_target_lmm_post_hoc}) where we observe only a partial confirmation of the pairwise hypotheses. The ENet model is outperformed by the Median baseline for three targets, namely Future Active Time, Session Time and Session Activity. All the coefficients in the regression analyses and the differences in the \textit{post hoc} analyses are non-standardized and can be interpreted as absolute changes in percentage error (i.e. SMAPE). In order to make these values more easily interpretable, we can use the information Table \ref{game_description}. For example, knowing that the median Session Time for the jc3 object is 162 minutes we can derive that when the RNN model achieves a SMAPE of 30\% in predicting Future Session Time, this equates on average to an absolute error of $1.62 \times 30 \sim 48$ minutes. All the p-values in the \textit{post hoc} analyses are Bonferroni corrected for multiple comparisons. The results of the statistical analyses suggest positive additive effects of non-linearity and recurrency on model performance both at the level of global and target-specific performance. This effect is more pronounced for certain targets (e.g. Future Session Time, Future N° Sessions) than for others (e.g. Future Absence, Future Active Time). Moreover, looking at Figure \ref{model_comp_non_coll} it appears that RNN improved on MLP (i.e. the second best model) using roughly half the parameters and per-epoch computation time. This could indicate that recurrency both improves model fit and allows for more efficient use of the available parameters.

\begin{table}[h]
\centering
\caption{Results of LMM on Non-Collapsed Targets}
\label{exploded_target_lmm}
\begin{tabular}{ccccc}
\hline
\textbf{Model}  & \textbf{$\beta$} & \textbf{Z} & \textbf{p} & \textbf{95\% C.I.}                  \\ \hline
\multicolumn{5}{c}{\textbf{Future Absence}}                                                                         \\ \hline
\textbf{Intercept (RNN)} & 54.46                & 144.316     & \textless .01   & 53.72 - 55.20                     \\
\textbf{Lag 1}           & 7.40                & 40.509     & \textless .01   & 7.04 - 7.76                     \\
\textbf{Median}          & 9.47                & 51.814     & \textless .01   & 9.11 - 9.83                     \\
\textbf{ENet}            & 1.71                & 9.353     & \textless .01   & 1.35 - 2.06                     \\
\textbf{MLP}             & .53                & 2.915     & \textless .01   & .175 - .891                       \\ \hline
\multicolumn{5}{c}{\textbf{Future Active Time}}                                                                     \\ \hline
\textbf{Intercept (RNN)} & 23.36                & 20.019      & \textless .01  & 21.78 - 24.93                     \\
\textbf{Lag 1}           & 7.32               & 119.028    & \textless .01  & 7.2 - 7.44                     \\
\textbf{Median}          & .77                & 12.515     & \textless .01  & .649 - .891                     \\
\textbf{ENet}            & 2.55                & 41.551     & \textless .01  & 2.43 - 2.67                     \\
\textbf{MLP}             & .41                & 6.739      & \textless .01  & .294 - .535                     \\ \hline
\multicolumn{5}{c}{\textbf{Future Session Time}}                                                                     \\ \hline
\textbf{Intercept (RNN)} & 30.02                & 63.663     & \textless .01  & 29.1 - 30.95                     \\
\textbf{Lag 1}            & 5.62                & 49.158     & \textless .01  & 5.39 - 5.83                     \\
\textbf{Median}          & 4.45                & 38.957     & \textless .01  & 4.23 - 4.67                     \\
\textbf{ENet}            & 4.81                & 42.098     & \textless .01  & 4.59 - 5.03                     \\
\textbf{MLP}             & 1.08                & 9.529      & \textless .01  & .86 - 1.31                     \\ \hline
\multicolumn{5}{c}{\textbf{Future Session Activity}}                                                                 \\ \hline
\textbf{Intercept (RNN)} & 31.04                & 70.241     & \textless .01  & \multicolumn{1}{l}{30.17 - 31.9} \\
\textbf{Lag 1}            & 6.52                & 61.025     & \textless .01  & \multicolumn{1}{l}{6.31 - 6.73} \\
\textbf{Median}          & 4.37                & 40.879     & \textless .01  & \multicolumn{1}{l}{4.16 - 4.58} \\
\textbf{ENet}            & 4.42                & 41.319     & \textless .01  & \multicolumn{1}{l}{4.21 - 4.63} \\
\textbf{MLP}             & 1.36                & 12.762     & \textless .01  & \multicolumn{1}{l}{1.15 - 1.57} \\ \hline
\multicolumn{5}{c}{\textbf{Future N° Sessions}}                                                                      \\ \hline
\textbf{Intercept (RNN)} & 26.77                & 7.445     & \textless .01  & 19.72 - 33.81                     \\
\textbf{Lag 1}            & 23.17                & 44.430     & \textless .01  & 22.15 - 24.19                     \\
\textbf{Median}          &  18.34              & 35.166     & \textless .01  & 17.31 - 19.36                     \\
\textbf{ENet}            & 7.44                & 14.277     & \textless .01  & 6.42 - 8.46                     \\
\textbf{MLP}             & 3.65                & 7.005      & \textless .01  & 2.63 - 4.67                       \\ \hline
\end{tabular}
\end{table}

\begin{table}[h]
\centering
\caption{LMM Post-Hoc on Non-Collapsed Targets}
\label{exploded_target_lmm_post_hoc}
\begin{tabular}{ccccc}
\hline
\textbf{Contrast}  & \textbf{$\beta_1$-$\beta_2$} & \textbf{Z} & \textbf{p} & \textbf{95\% C.I.}                  \\ \hline
\multicolumn{5}{c}{\textbf{Future Absence}}                                                                         \\ \hline
\textbf{Lag 1 - Median} & -2.06                & -11.305     & \textless .01   & -2.42 - -1.70                     \\
\textbf{Median - ENet}           & 7.76                & 42.461     & \textless .01   & 7.40 - 8.12                     \\
\textbf{ENet - MLP}          & 1.17                & 6.438     & \textless .01   & .81 - 1.535                    \\ \hline
\multicolumn{5}{c}{\textbf{Future Active Time}}                                                                     \\ \hline
\textbf{Lag 1 - Median} & 6.55                & 106.513      & \textless .01  & 6.433 - 6.67                     \\
\textbf{Median - ENet}           & -1.78                & -29.037    & \textless .01  & -1.9 - -1.66                     \\
\textbf{ENet - MLP}          & 2.14                & 34.812     & \textless .01  & 2.02 - 2.26                     \\\hline
\multicolumn{5}{c}{\textbf{Future Session Time}}                                                                     \\ \hline
\textbf{Lag 1 - Median} & 1.16                & 10.201     & \textless .01  & .94 - 1.39                     \\
\textbf{Median - ENet}            & -.35                & -3.141     & \textless .01  & -.58 - -.13                     \\
\textbf{ENet - MLP}          & 3.72                & 32.579     & \textless .01  & 3.5 - 3.95                     \\ \hline
\multicolumn{5}{c}{\textbf{Future Session Activity}}                                                                 \\ \hline
\textbf{Lag 1 - Median} & 2.15                & 20.146     & \textless .01  & \multicolumn{1}{l}{1.94 - 2.36} \\
\textbf{Median - ENet}            & -.04                & -.441     &  1.  & \multicolumn{1}{l}{-.257 - .163} \\
\textbf{ENet - MLP}          & 3.05                & 28.558     & \textless .01  & \multicolumn{1}{l}{2.84 - 3.26} \\ \hline
\multicolumn{5}{c}{\textbf{Future N° Sessions}}                                                                      \\ \hline
\textbf{Lag 1 - Median} & 4.83                & 9.264     & \textless .01  & 3.8 - 5.85                     \\
\textbf{Median - ENet}            & 10.89                & 20.889     & \textless .01  & 9.87 - 11.91                     \\
\textbf{ENet - MLP}          & 3.79                & 7.272     & \textless .01  & 2.77 - 4.81                     \\\hline
\end{tabular}
\end{table}

\subsection{\textbf{Representation Analysis}}
\label{repr_results}
\begin{figure}[h]
\centering
\includegraphics[width=\columnwidth]{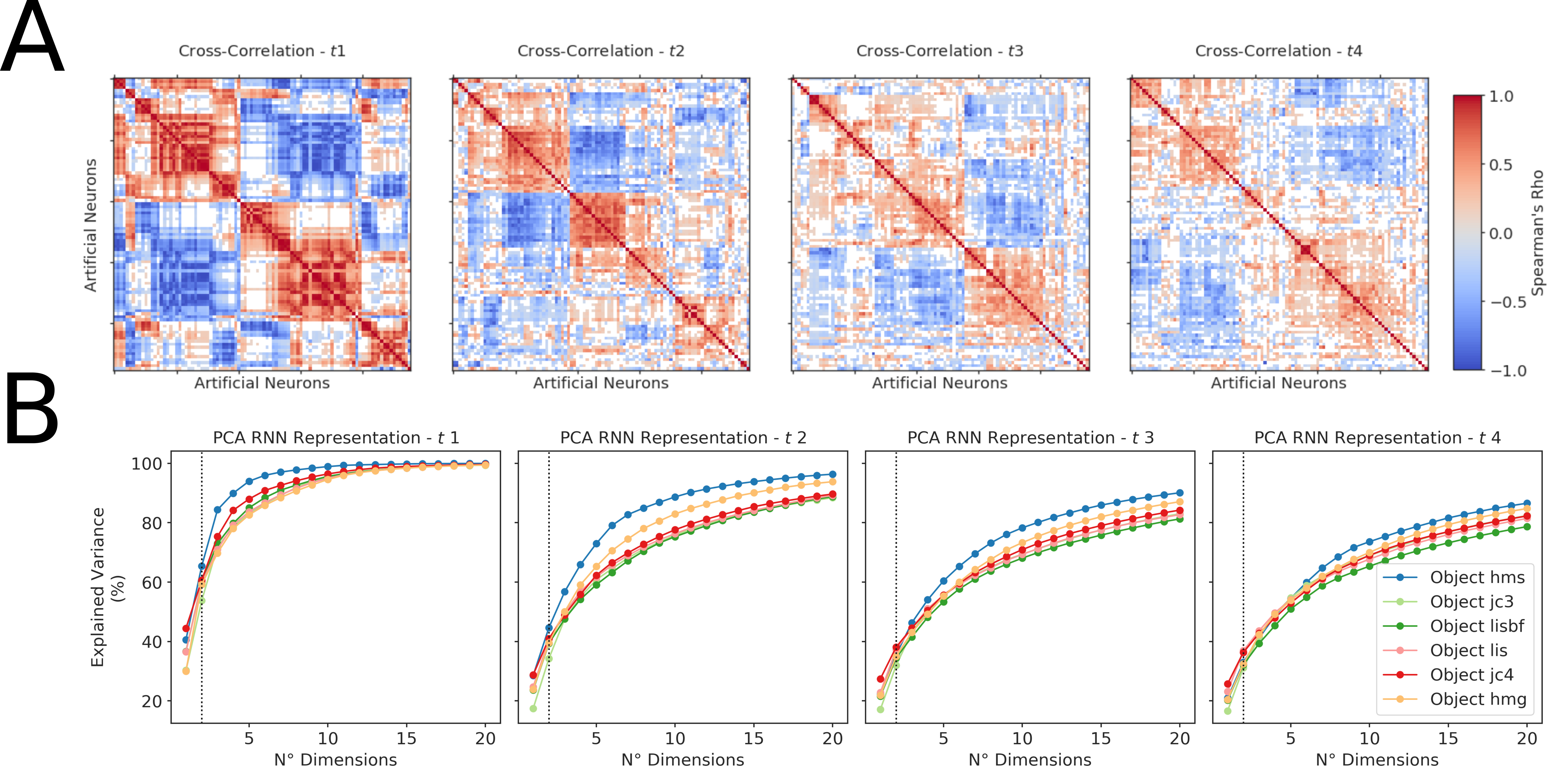}
\caption{\textbf{The activity of the RNN's artificial neurons is markedly redundant.} Panel A shows the cross correlation between the activity of the RNN's artificial neurons in the game object $hms$ going from $t1$ to $t4$. The y and x axes are symmetrical and identifies the RNN artificial neurons while the coloured cells report the Spearman's Rho correlation coefficient for the activation of each pair of neurons. White cells represent combinations for which the correlation coefficient resulted lower than 0.05. \textbf{Two principal components can explain a large portion of variance in the representation generated by the RNN.} Panel B shows the percentage of explained variance by considering 2 to 20 principal components for each game object going from $t1$ to $t4$. The y axis indicates the percentage of explained variance while the x axis the number of principal components considered.}
\label{pca_emb} 
\end{figure}

\begin{figure*}[ht]
\centering
\includegraphics[width=\textwidth]{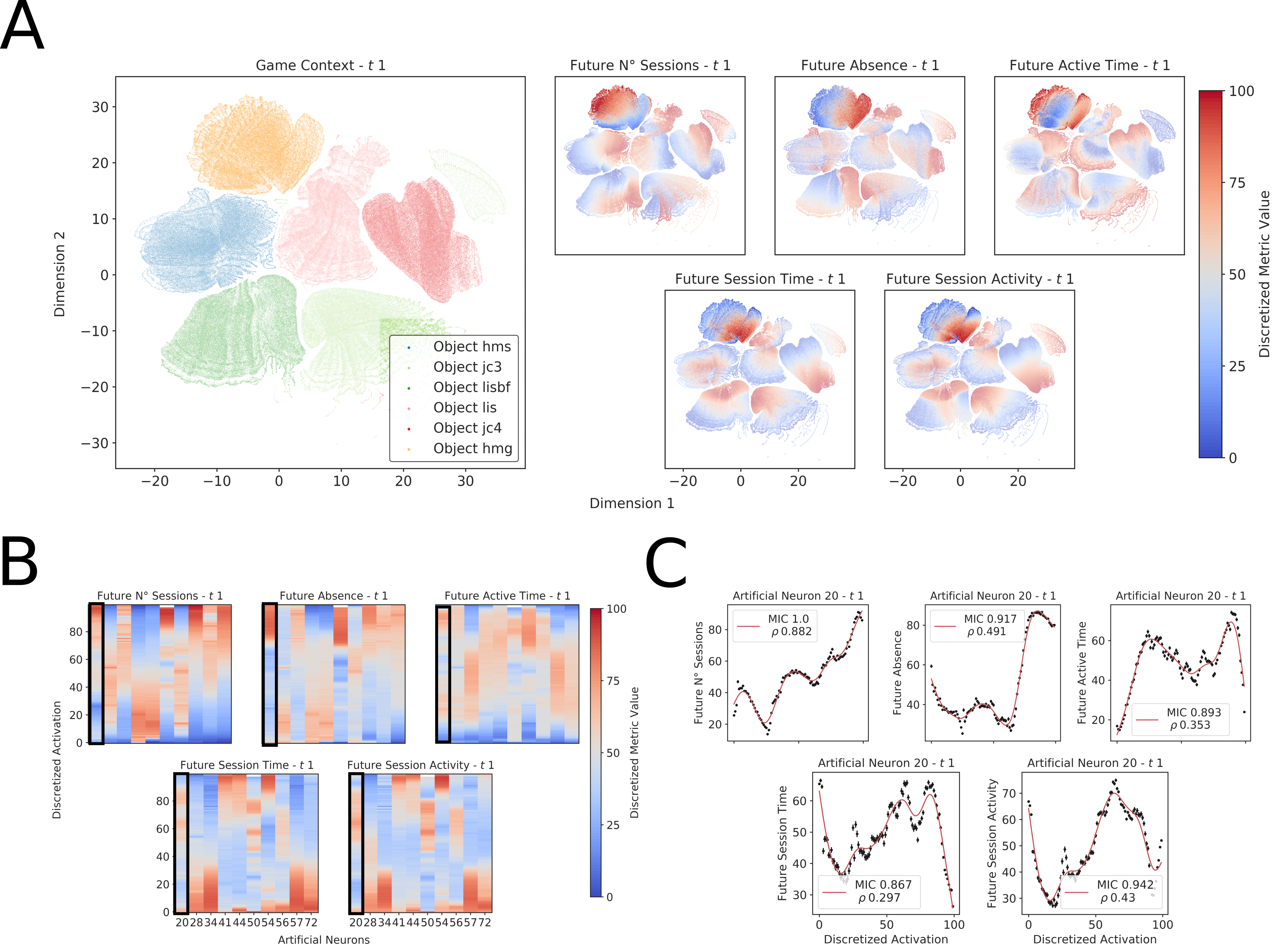}
\caption{\textbf{The representation generated by the RNN model distinguishes between different game objects while maintaining an overarching organization able to capture variations in the expected intensity of future interactions that individuals will have with a specific game object}. Panel A shows the two-dimensional projection, produced by UMAP, of the multi-dimensional representation inferred by the RNN at $\mathbf{t1}$ as produced by UMAP. We can read the values of the x and y axes as a coordinate system where proximity represents similarity between points in the original high-dimensional space. Each point indicates the representation inferred by the RNN model after observing one game session from a single user. The colours in the Game Context panel indicate the game object from which the representation is coming. Colours in the small panels represent the discounted sum of all future predictions for a particular target (for example, estimated Future Session Time) $\widehat{B}_{t2:T}$ which is given by $\sum_{i=0}^{t2:T} \gamma^i\widehat{B_i}$ with $\gamma=0.1$ as illustrated in equation \ref{td_v}. \textbf{Each unit  encodes the intensity of future interactions through multiple non-monotonic functions}. Panels B and C show the relationship between the activation of randomly-selected hidden units in the LSTM layer of the RNN and the model's predictions at $\mathbf{t1}$. Panel B shows the relationship between the discretized activation of 10 randomly selected units (artificial neurons) plotted along the y axis and the predictions made by the model at $t1$ (colour coded from blue to red as in the small panels in A) for the game object $hmg$. Panel C shows in more detail the relationship between discretized activation and RNN predictions for a single unit highlighted by a black box in Panel B. Here the x axis indicates the discretized activation while the y axis the mean discretized discounted sum of all future predictions produced by the model. Vertical lines are standard errors of the mean. The red curve is the line of best fit provided by a generalized additive model \citep{serven2018} while the box reports the MIC and the correlation coefficient (Spearman's $\rho$) between the artificial neuron activation and the model's predictions.}
\label{full_panel_static}
\end{figure*}

\begin{figure*}[ht]
\centering
\includegraphics[width=\textwidth]{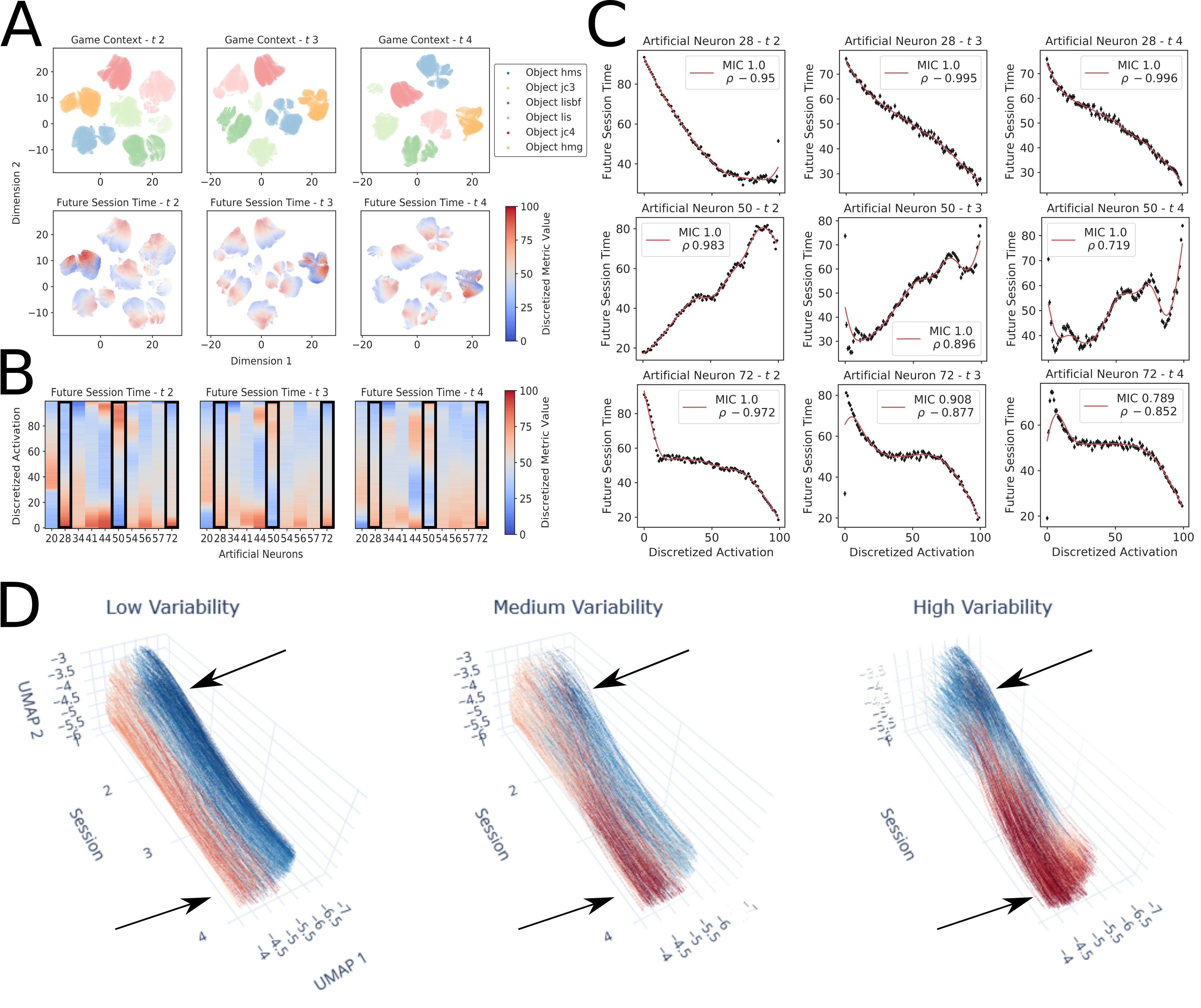}
\caption{\textbf{The representation generated by the RNN model maintains its discriminant properties over time}. Panel A shows a two-dimensional projection of the multi-dimensional representation inferred by the RNN at $t2$, $t3$ and $t4$. The inferred representation maintains its gradient-like organization over time with an increased ability to differentiate between game objects. As in Figure \ref{full_panel_static}, x and y axes are dimensions individuated by the UMAP algorithm and can be interpreted as a coordinate system where proximity represents similarity between points. Colours in the first row indicate which game object the representation is coming from while those in the second row indicate the discounted sum of future predictions for a single target (i.e. "Future Session Time"). \textbf{The units constituting the generated representation encode for functions that are consistent over time.} Panels B and C show the relationship between units' activation and the model's predictions over time for the game object $hmg$. Different units appear to encode the same target with different non non-monotonic functions which are relatively consistent over time. Panel B illustrates the relationship between the same 10 randomly selected units specified in figure \ref{full_panel_static} and the predictions made by the model for Future Session Time at $t2$, $t3$ and $t4$. Panel C shows in more detail the relationship of the three artificial neurons, highlighted by black boxes in B, across time. Each row is a different unit while each column corresponds to a different $t$. The x axis indicates the discretized activation while the y axis the mean discretized discounted sum of all future predictions. Vertical lines are standard errors of the mean. The red curve is the line of best fit provided by a generalized additive model \citep{serven2018} while the box report the MIC and the correlation coefficient (Spearman's $\rho$) between the artificial neuron activation and the model's predictions. \textbf{The generated representation produces areas of low and high expected intensity among which individuals move over time.} Panel D shows trajectories through time produced by a version of UMAP that incorporates temporal information. Data are drawn from random subsets of individuals having low, medium and high variability in their expected amount of future behaviour. The representation inferred by the RNN model produces "hot" (i.e. the left side) and "cold" (i.e. the right side) regions, representing high and low expected Future Session Time, that are spatially consistent over time. Individuals appear to either stay in the same region or to move between regions over time. Here each line represents variations in the representation generated by the RNN model for a single user over four temporal steps. Continuity is generated by means of cubic spline interpolation for the lines and by linear interpolation for the colours. The x and y axes are the dimensions individuated by the UMAP algorithm while the z axis indicates the associated point in time. Colours indicate the discounted sum of future predictions produced by the model at a specific point in time.}
\label{full_panel_temporal}
\end{figure*}

\begin{figure*}[ht]
\centering
\includegraphics[width=\textwidth]{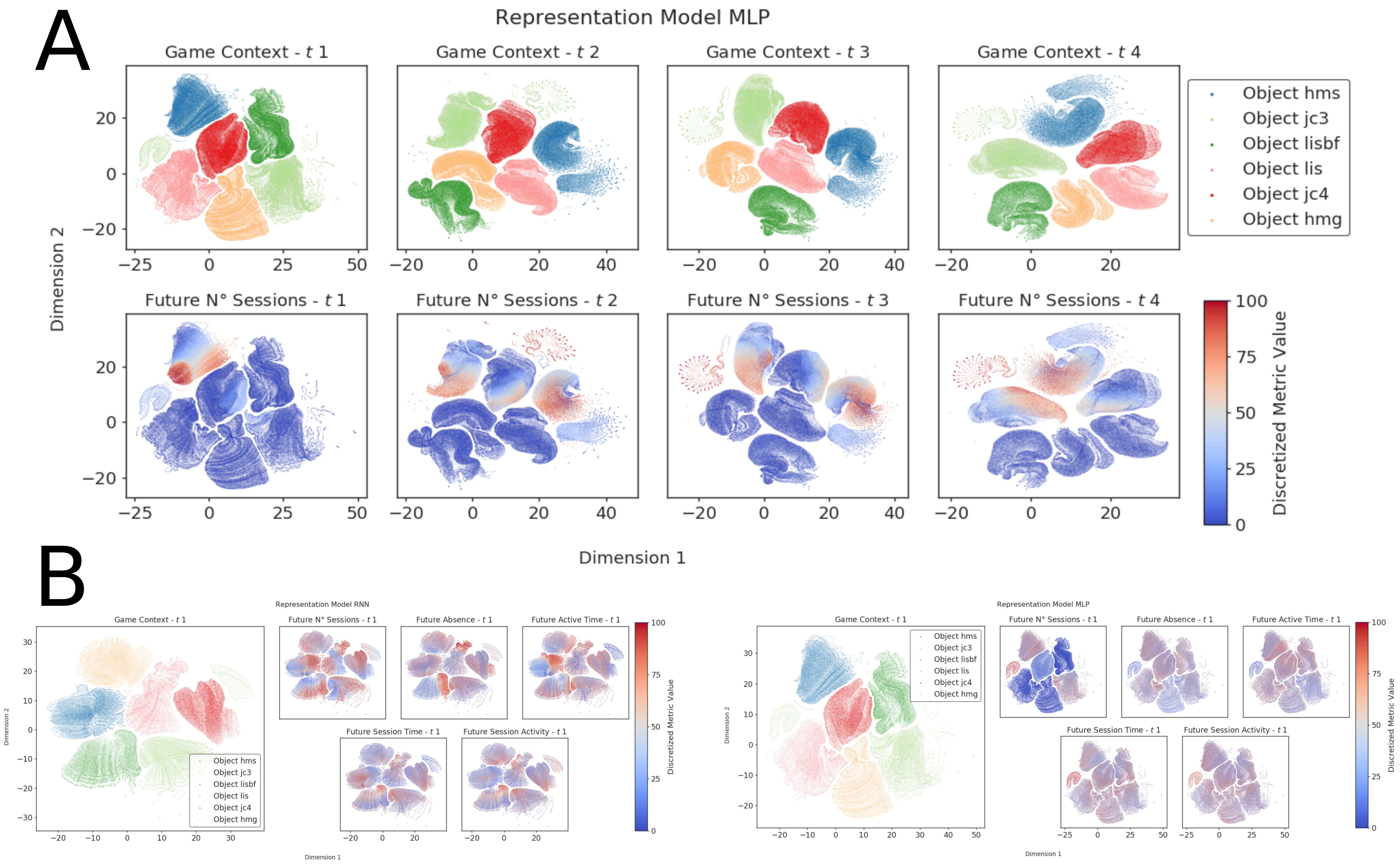}
\caption{\textbf{The representation generated by the MLP model is less effective at distinguishing between different game objects and different levels of expected future behaviour intensity.}. Panel A shows a two-dimensional projection of the multi-dimensional representation inferred by the MLP at $t1$, $t2$, $t3$ and $t4$. Differently from the RNN, the representation shows a disruption in the gradient-like organization and a reduced ability to differentiate between game objects which remain constant over time. The x and y axes are dimensions individuated by the UMAP algorithm and can be interpreted as a coordinate system where proximity represents similarity between points. Colours in the first row indicate which game object the representation is coming from while those in the second row indicate the discounted sum of future predictions for a single target (i.e. "Future N° of Sessions") \textbf{The representation generated by the MLP model is less effective at at distinguishing different levels of expected behaviour intensity for states that are further away in the future.} Panel B shows a two-dimensional projection of the multi-dimensional representation inferred by the RNN (left) and MLP(right) at $t1$ but colour coded with the discounted sum of future predictions from $t4$ onward. The representation generated by the RNN is able to maintain a gradient-like organization even from states that are further away in the future while this capacity is almost entirely lost for the MLP. The colours in the Game Context panel indicate the game object from which the representation is coming. Colours in the small panels represent the discounted sum of all future predictions for a particular target computed from $t4$ onward instead that from $t1$. The x and y axes are the dimensions individuated by the UMAP algorithm.}
\label{predictive_panel}
\end{figure*}

From figure \ref{pca_emb}A we can observe consistent patterns of cross-correlation for the activity of the artificial neurons constituting the RNN representation. This is supported by the fact that, considering only two dimensions, PCA was able to explain from 30 to 60\% of the variance in the representation generated by the RNN, with maximum explanatory power around 6 and 8 principal components.
Inspecting the representation generated by the RNN model at $t1$ (see Figure \ref{full_panel_static}A) we observe that the model was able to effectively distinguish between different game objects while simultaneously encoding for variations in the expected intensity of future interactions. This is illustrated by the fact that each game object occupies different and distinct regions in the representation space while showing a within-object gradient-like organization that places individuals (i.e. single dots) on a continuum based on the estimated magnitude of their future behaviour. This organization is preserved for each of the six targets showing how the representation inferred by the model is a suitable meta-descriptor for different behavioural indicators. As expected, some targets show a very similar but not identical organization (e.g. Future Session Time and Future Session Activity) while others appear to be independent (e.g. Future Session Time and Future Absence). We note that the absolute location of each game aggregate (i.e. all the points belonging to a specific game object) on the 2D plane is arbitrary. As we can see in figures \ref{full_panel_static} and \ref{full_panel_temporal}, this will change at every run of the algorithm due to the stochastic nature of UMAP. Panels \ref{full_panel_static}B and \ref{full_panel_static}C provide more insight into the activation profiles of individual hidden units constituting the generated representation. Panel \ref{full_panel_static}B shows the relationship between the activity of 10 randomly-chosen units and the predictions generated for the five targets. These are essentially transducer functions illustrating how the estimate for a particular target varies (on average) as the output of a units increases or decreases. Each unit seems to encode for multiple non-monotonic functions, one for each of the considered targets. Differences in the shape of these functions reflect similarities between their associated targets. For example, the functions associated to two highly related targets like Future Session Time and Future Session Activity (see panel \ref{full_panel_static}A) appear to be very similar in shape (see panels \ref{full_panel_static}B and \ref{full_panel_static}C). Interestingly, although most units appear to encode for unique functions some of them (e.g. 41 and 44) show an almost identical behaviour. This suggests the presence of redundancy in the functions underlying the representation generated by the RNN model. These observations are clarified in panel \ref{full_panel_static}C, where the functions associated with a single unit (20, indicated by a dark box in \ref{full_panel_static}B) are presented. Here we observe a strong, non-linear relationship between the unit's activity and the estimated targets (see the high MIC values and the line of best fit). In addition, the between-targets variation in MIC values suggest how the chosen unit is not equally informative for all targets.
\\
\\
The analyses in Figure \ref{full_panel_static} were performed at a single time point $t1$. When performed at subsequent time points the results appear to be qualitatively similar. For example, focusing on Future Session Time (see Appendix \ref{appendix_representation} for results connected to other targets), we see in Figure \ref{full_panel_temporal}A that the model's ability to segregate different game objects while providing an  overarching representation of the intensity of future interactions is preserved over time. This supports the hypothesis that the representation inferred by our model is dynamic in nature which is further corroborated by panel \ref{full_panel_temporal}D. There we can see how the RNN model was able to individuate a "space" with temporally consistent ”hot” and ”cold” regions between which individuals moved over time depending on the expected intensity of their future interactions. This means that given the history of interaction of a particular individual with a specific game object, our model would determine their "position" (i.e. their "internal state") in the "attributed incentive salience space". This aligns with the manifold hypothesis mentioned in sections \ref{manifold_state} and \ref{manifold_learning}: changes in the propensity to interact with a specific game object (i.e. variations in the amount of attributed incentive salience) can be expressed moving on a manifold embedded within an $h$ dimensional space, with $h$ being the dimensionality of the representation generated by our RNN model. It appears that the hidden units constituting this representation tend to be consistent over time in the type of functions they encode (see Figure \ref{full_panel_temporal}B and C). As expected, we can again observe a strong non linear association between units' activation and targets' predictions, see MIC values and lines of best fit. The decrease in MIC value observed in Figure \ref{full_panel_temporal}C for the artificial neuron 72 might indicate how certain units lose their informative power over time.
As we mentioned in section \ref{comp_framework}, both ANNs try to predict the intensity of future behaviour given the history of interactions. They do so relying on the same type of metrics, leveraging similar computational mechanisms (i.e. multitask learning and non-linearity) and producing representation according to the same underlying principle (i.e. the manifold hypothesis). Nevertheless, the fact that MLP provides poorer fit to data already suggests that whatever representation it has inferred it is likely a sub-optimal approximation of the manifold structure of incentive salience. Looking at figure \ref{predictive_panel}A, and knowing that UMAP represents differences and similarities between points through distance, we can see how the representation generated by the MLP less clearly differentiate between game objects. On the same figure, we can notice how the gradient representation for the metric Future N° Sessions Time is largely disrupted. This effect is however consistently less pronounced for other metrics (see our \href{https://htmlpreview.github.io/?https://github.com/vb690/approx_incentive_salience/blob/main/notebooks_html/embedding_analysis.html}{GitHub} for additional visualizations), in accordance to the differences we observed in predictive performance (see Figure \ref{model_comp_non_coll}). Recalling what mentioned in section \ref{comp_framework}, the latent state produced by the level of attributed incentive salience should retain at any point in time some predictive power over the intensity of all the future interactions (i.e. not just the one that follows). Figure \ref{predictive_panel}B shows the representation generated by RNN and MLP at $t1$ but color coded with the discounted sum of the predictions made from $t4$ onward. We can see that, even if degraded, RNN still preserves some of the desired gradient-like organization which is instead much more disrupted for MLP. This is in accordance to what is shown by Figure \ref{full_panel_temporal}D: the RNN appears to define regions of high and low expected behavioural intensity which are consistent over time rather than constrained to the region around $t+1$.

\subsection{\textbf{Partition Analysis}}
\label{part_results}
\begin{figure*}[ht]
\includegraphics[width=\textwidth]{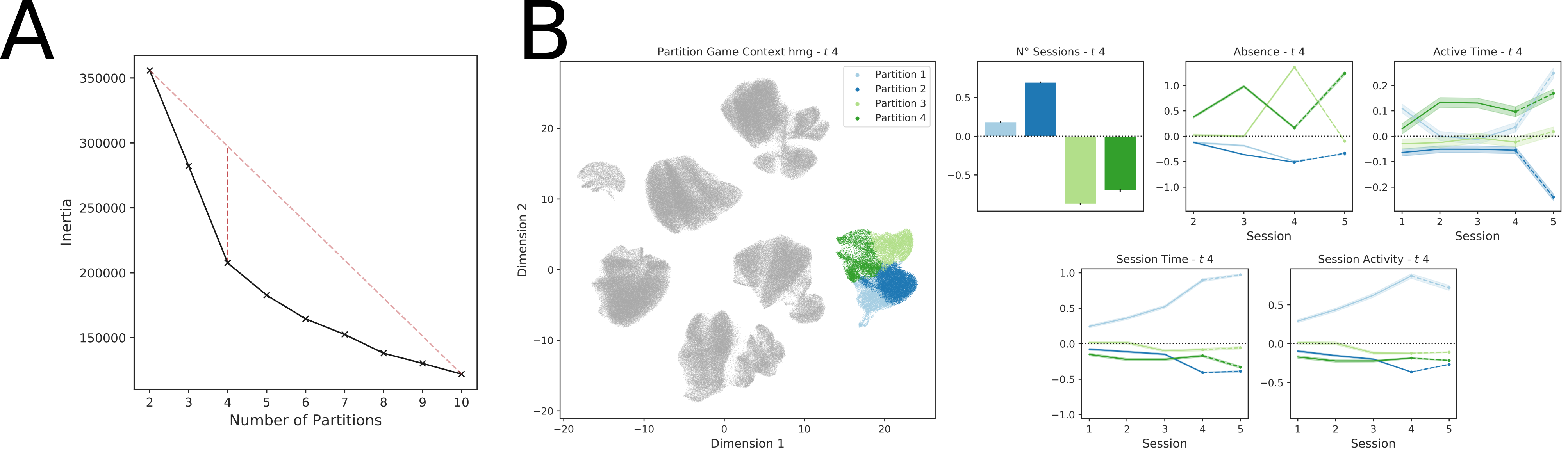}
\caption{\textbf{Partitioning the representation generated by the RNN model produces behavioural profiles with distinct characteristics.} Panel A shows how the optimal number of partitions was individuated using the "elbow method". Here the x axis indicates the number of partitions tested while the y axis shows the associated inertia. The point of maximum curvature (i.e. the optimal number of partitions) was found by identifying the number of partitions maximizing the distance (i.e. the vertical line) from the overall gradient of the inertia (i.e. the oblique line). Panel B shows the individuated partitions and associated behavioural profiles for the game object hmg at $t4$. The big panel reports the same UMAP reduction presented in the last column of Figure \ref{full_panel_temporal}. Each dot is the representation associated with a particular individual and is colour coded based on the partition to which it belongs. Small panels represent the temporal evolution of four of the considered behavioural metrics for each individuated partition. The panel relative to N°Sessions only reports the prediction produced by the model as the number of preceding session is constant for all the partitions. The x axis reports the game sessions while the y axis the value assumed by the considered metric at a specific point in time. The y axis is expressed in terms of number of standard deviations from the game population mean (i.e. z-scores). Each line indicates the mean z-score while the shaded area around the line its 95\% confidence interval. The solid part of each line indicates the portion of the temporal series observed by the model (i.e. the input) while the dotted part the predictions produced at that point in time.}
\label{partitioning} 
\end{figure*}

We will focus on the representation generated at $t4$ for the game object $hmg$, as highlighted in the rightmost panel of Figure \ref{full_panel_temporal}A (results related to other game objects can be found in Appendix \ref{appendix_partition}). As we can see from Figure \ref{partitioning}A, following the methodology outlined in section \ref{procedure}, among all the Mini-Batch K-Means runs, the one with $k=4$ was identified as optimal. All the partitions were associated with a distinct behavioural profile, each one with a distinct offset and temporal evolution.
At a global level, the four partitions seem to belong to two general groups: a group with a high propensity to produce future interactions (i.e. partitions 1 and 2) and a group with low propensity (partitions 3 and 4). Noticeably, when looking in detail at each specific partition they appear as variations on the macro group they belong to. Interestingly the percentage of Session Time spent actively interacting with the game object (i.e. Active Time) seems to be a relevant component in this more granular characterization.

\paragraph{\textbf{Partition 1}} represents individuals producing very high intensity interactions (see Session Time and Activity) at a high frequency (see Absence). The high amount of Active Time highlights how the individuals were actively interacting with the game object. The individuals in this partition are projected to produce a number of future interactions that is slightly above average while maintaining a high intensity profile. It can be speculated that the history of high intensity interactions reflected a positive propensity towards the game. This might have prompted individuals in this partition to consume most of the available contents in the game leading to a reduced amount of expected future interactions.

\paragraph{\textbf{Partition 2}} describes individuals that have a history of very frequent (see Absence) but brief interactions with little activity and active time. These individuals are expected to maintain this trend in the future although producing a number of interactions that is largely above the average. An hypothetical explanation might see individuals in this partitions constituting a variant of those in Partition 1. The high frequency of interactions could suggest an eagerness to interact with the game object. This, combined with the low amount of consumed content (see Session Time, Active Time and Session Activity) could explain the projected high amount of future interactions.

\paragraph{\textbf{Partition 3}} includes individuals whose interactions have been average both in terms of length, amount of activity and frequency until session 3. From there, a reduction in both length and activity can be observed concomitant with a large temporal hiatus before the following interaction. This reduction is expected to continue and these individuals are estimated to produce a number of future interactions markedly below average while also maintaining a low intensity of interactions. These individuals might have started with a normal propensity towards the game which suddenly degraded around session 4, leading to a marked reduction in the expected number of future interactions.

\paragraph{\textbf{Partition 4}} contains individuals producing the least intense and frequent interactions. However, these individuals have the highest amount of active time. Similarly to partition 3 a long temporal hiatus seems to follow a slight dip in session time and activity. Also similarly to partition 3 these individuals are estimated to produce a number of future interactions below average while maintaining the original low intensity profile. Differently from partition 3, these individuals started and maintained a low intensity profile, suggesting a negative propensity toward the game. However, the short idle time (see Active Time) characterizing their game sessions might have worked as an attenuating factor leading to a higher amount of expected future interactions.\\
\\
Looking at the relationship between behavioural metrics we observe that Session Time and Session Activity are usually highly correlated. Low absence seems to be a good indicator of the propensity to produce more interactions in the future. Similarly, high absence seems to be associated with a general history of low intensity interactions. It is also worth noting that variations in this metric seem to follow and be proportional to increases and decreases in interactions' intensity (e.g. see partitions 1, 3 and 4).

\section{Discussion}
In this paper we presented a modelling approach based on ANNs for approximating the manifold structure of motivation-related latent states in scenarios where only large amounts of behavioural data are available. Our approach produces predictions of the intensity of future interactions between individuals and a diverse range of video games while also generating representations that well approximated some of the properties of attributed incentive salience. We also show how integrating theoretical and computational insights related to the concept of attributed incentive salience in the design of our modelling approach improved its efficiency and effectiveness.

\paragraph{\textbf{Theoretical Implications}}
The advantage provided by the combination of non-linearity and recurrency in the estimation task is in line with the dynamical nature of motivation and incentive salience attribution \citep{toates1994comparing,robinson1993neural,zhang2009neural,tindell2009dynamic,berridge2012prediction}. This is also consistent with a body of research showing that the attribution of value to potentially rewarding objects or actions is often carried out by non-linear recurrent operations \citep{song2017reward,wang2018prefrontal} and that Artificial Neural Networks with recurrent connections are well suited for approximating these operations \citep{kietzmann2018deep}. These findings are corroborated not just by the superior performance of the RNN model in the prediction task (see section \ref{perf_results}) but also by its capacity to produce more stable representations (see Figure \ref{predictive_panel}). As mentioned in section  \ref{incentive_salience}, incentive salience attribution produces latent representations of objects which, when imbued with value, make future interactions with those objects more likely and intense \citep{berridge1998role,berridge2004motivation}. The representation generated by our model showed similar functional properties in their global-local organization. At the global level, different game-objects were organized in distinct and coherent regions (see Figure \ref{full_panel_static}A) showing how the model attempted to operate on a meta-level by partitioning a global representation in several object-specific ones. This finding aligns with what highlighted in various work on neural manifold where the responses related to qualitatively different stimuli tends to show a cluster-like organization when reduced to a lower dimensional space \citep{stopfer2003intensity, gallego2017neural, ganmor2015thesaurus}. At the local level, each object-specific representation showed an internal gradient-like organization distinguishing individuals based on the estimated intensity of their future interactions with that specific object. This was true for each of the considered behavioural targets (see Figure \ref{full_panel_static}A) showing how the model attempted to provide an holistic description of the intensity of future interactions. The presence of this type of gradient-like organization emerged in a work by Nieh et al. \citep{nieh2021geometry} when analyzing neural responses during an evidence accumulation task in virtual reality. When reducing the neural activity to a 3 dimensional space, the resulting manifold presented a clear gradient able to code simultaneously for position and levels of accumulated evidence \citep{nieh2021geometry}. A similar finding was present in the work by Stopfner et al. \citep{stopfer2003intensity} where the manifold structure extracted from the activity of olfactory neurons was able to represent qualitative and quantitative differences between odours through a global-local organization similar to that showed in section \ref{repr_results}. The dynamic nature of the representation generated by our approach also nicely fits with that of attributed incentive salience \citep{toates1994comparing,robinson1993neural,zhang2009neural,tindell2009dynamic,berridge2012prediction}. In particular, the fact that the aforementioned global-local organization is maintained over time (see Figure \ref{full_panel_temporal}A) corroborate the hypothesis that our model approximated state changes originated from a dynamic process. In support of this, we also observed that the representation generated by our model was spatially coherent over time: it produced distinct regions of low and high expected intensity between which individuals moved over time (see Figure\ref{full_panel_temporal}D). These results appear to match the definition of motivation and incentive salience attribution specified in section \ref{motivation}: a single overarching process able to dynamically predict the likelyhood and intensity by which individuals will interact with a varied set of objects \citep{simpson2016behavioral,toates1994comparing,berridge2004motivation,zhang2009neural}. Many other cognitive and affective functions might rely on a latent representation that is functionally similar to the one described in our work (e.g. credit assignment and optimal control \cite{wang2018prefrontal, barto2004reinforcement}, cognitive control, learning \cite{skinner1965science} or various forms of reward processing \cite{schultz1997neural, schultz2000reward}). Similarly to attributed incentive salience, these functions are all involved in generating motivated behaviour and heavily rely on reward signals, however none of them is concerned with attributing and describing the motivational saliency that an object possess. This is made evident in the works by McClure et al. \cite{mcclure2003computational} and Zhang et al. \cite{zhang2009neural} where the system involved in salience attribution is functionally separate from the one assigning credit and executing actions: the former provide a representation that informs and biases the decisions taken by the latter serving an almost exclusively qualifying role (see the role of attributed incentive salience in addiction-like conditions \cite{robinson1993neural}). Similarly, the representation generated by our model doesn't provide any insight on the decision making process underlying the observed playing behaviour but simply provide an approximate description of the "motivational pull" that a particular game object has on a particular individual at a certain point in time. The functions encoded by the hidden units constituting the representation appeared to have a series of distinctive properties, namely: redundancy, non linearity, multiplicity (single units code for multiple functions) and consistency over time. These may have played a role in providing the representation generated by our model with its distinctive characteristics. For example, as we mentioned in section \ref{manifold_state} redundancy and inter-correlation are characteristics of the signals from which the manifold representation of internal states arises \citep{seung2000manifold,gallego2017neural}. Multiplicity on the other hand, might be the factor underlying the ability of our model to produce a single unitary representation which holds predictive power over different behavioural targets. Finally, consistency over time could be the mechanisms supporting the type of temporal coherence observed in panel \ref{full_panel_temporal}D. We want to stress that these findings are to be considered exploratory in nature since they do not rely on a-priori hypotheses. A comparison between these computational properties and those underlying the attribution of incentive salience is required and would constitute a potential venue for future investigations. This supports the idea that our approach, by giving full access to its constituent parts, provides a certain degree of interpretability and offers the possibility of generating testable hypotheses. The partition analysis revealed a set of diverse profiles that largely reflect expected behavioural correlates of different levels of attributed incentive salience (i.e. high vs low intensity profiles) \citep{berridge2004motivation}. The various offsets that each partition showed might suggest different levels of predisposition towards the individual game-objects. The dynamic nature of these profiles provided a more granular characterization allowing to observe variations in the entire history of interactions and not just in the expected intensity of future ones. For example, it was possible to see how a higher likelihood of future interactions was supported both by a history of low intensity but high frequency interactions as well as by a series of high frequency and high intensity interactions (see partitions 1 and 2 in Figure\ref{partitioning}B). In this sense, these behavioural profiles can be seen as useful devices for investigating the existence of inter-individual differences in schedules of interactions with potentially rewarding objects.

\paragraph{\textbf{Applicative Implications}}
The present work outlined a method for embedding theory-driven knowledge in data-driven approaches, allowing to more easily interpret and test hypotheses on the representation they produce. In comparison to other works focusing on the identification of latent states (or their manifold representation) from behavioural data \citep{calhoun2019unsupervised, luxem2020identifying, pereira2020quantifying, shi2021learning, mccullough2021unsupervised}, the present methodology offers a series of advantages. It does not require the Markov assumption, it generates continuous rather than discrete state space (hence the number of hidden states doesn't need to be specified) and it relies on a more easily scalable class of algorithms. Moreover, in contrast with a general tendency of utilising completely unsupervised techniques for capturing the manifold structure underlying behavioural data \citep{calhoun2019unsupervised, luxem2020identifying, pereira2020quantifying, shi2021learning, mccullough2021unsupervised}, our methodology attempts to extract representations which obey to specific functional constrains (see section \ref{manifold_learning}) and can therefore be more easily interpreted within a specific theoretical framework. Our approach offers a convenient framework for dealing with a diverse series of tasks. It allows to produce predictions of the amount and intensity of future interactions that an individual will have with a specific object. It generates a representation that can be analyzed (similarly to what has been done in section \ref{representation_analysis}) or provided as input to other algorithms. Indeed, the encoder mentioned in section \ref{representation_analysis} can be thought of as an automatic feature extractor. This can be used to reduce complex time series data of varying length to fixed-size vectors able to describe the propensity of an individual to interact with an object. For example, the analysis presented in section \ref{partition_analysis} showed how this process could be applied for time-series partitioning of large dataset. The present work leveraged data coming from video games but the adopted approach could easily be applied to other contexts. They only key requirement is the access to behavioural quantifiers describing the amount and intensity of interactions that an individual has with a particular object, service or task. This means that natural areas of application for our approach are those relying on the remote acquisition of behavioural data (e.g. web services or online experiments) but also situations in which large volumes of experimental data are available (e.g. large multi-center studies).  

\section{Limitations and Future Directions}
The work we just presented is not exempt from limitations. First, since our approach is attempting to solve an inverse problem, the issue of uniqueness arises. Many different latent states might have produced the behavioural patterns that our model observed and there is no guarantee of a strict one-to-one mapping between the representation generated by our model and attributed incentive salience. Our approach is formally different from that of TD Learning \footnote{See \citep{barto2004reinforcement} for a  review of the differences between supervised and reinforcement learning.} and does not model the process of incentive salience attribution but rather attempt to approximate the product of this process (i.e. changes in attributed incentive salience). For this reason a direct comparison with the work of McClure \textit{et. al.} \citep{mcclure2003computational} and Zhang \textit{et. al.} \citep{zhang2009neural} is difficult. Moreover, unlike TD learning \citep{schultz1997neural} our model is not guaranteed to converge on a quantification of $V$ that is directly comparable to its biological counterpart or that has arisen from the same type of computations. This is also reinforced by the differences in mechanistic functioning between biological and artificial neural networks \citep{lillicrap2019backpropagation,lillicrap2020backpropagation}. These issues are partially attenuated by the constraints provided by our theoretical framework but in line with similar reports in the literature \citep{calhoun2019unsupervised,wang2018prefrontal} a verification based on controlled experiments is desirable. This could be achieved applying our approach to behavioural data acquired in laboratory settings or investigating differences and similarities between the computations achieved by our approach and those produced in simulation experiments. Differently from the works of Calhoun \textit{et. al.} \citep{calhoun2019unsupervised},  McClure \textit{et. al.} \citep{mcclure2003computational} and Zhang \textit{et. al.} \citep{zhang2009neural}, our methodology relies on a  supervised learning approach to perform both prediction of future behaviour and latent state estimation, making this two tasks infeasible before any data is observed. This limitation could be attenuated by initializing our model using a representation  generated in an unsupervised manner. As we mentioned in section \ref{videogame_telemetries} the reward dynamics generated by the interaction between the individual and the game incentive mechanics play an important role in determining the intensity of future playing behaviour \citep{agarwal2017quitting, avserivskis2017computational, wang2018beyond}. In addition to this, we know that these dynamics are modulated by the internal state of the individual \citep{zhang2009neural} and by the context in which in which they are generated \citep{palminteri2015contextual}. These factors, were only partially captured by our approach as they require a higher temporal resolution (i.e. within rather than between sessions) as well as more granular indices (i.e. in-game and environmental information) than those we employed. As a consequence we can see how our approach, despite outperforming competing ones, still achieves a relatively high error rate in predicting some behavioural targets (e.g. future Absence). Moreover, it is not possible to determine if the differences in the behavioural profiles observed in Figure \ref{partitioning} should be ascribed to internal factors of the individuals, to changes in their environment or inside the game. A possible solution to this would be to incorporate information about the context in which the observed behaviour occurred (e.g. time, location or in-game events) and adapt the architecture of our model accordingly. As well as improving the performance of the model, this should also increase the quality of the generated representation and consequently also that of the derived behavioural profiles. The behavioural profiles individuated by the partition analysis generally reflect those predicted by theories of reward-driven motivation \citep{thorndike1927law,skinner1965science,berridge2004motivation} but they also show some unexpected and potentially contradictory results (see the differences between partitions 1 and 2 and between partitions 3 and 4 in Figure \ref{partitioning}B). Given the observational setting and the unsupervised learning analysis we adopted, the explanations provided in section \ref{partition_analysis} should be taken with caution and be seen mostly as a starting point for future investigations. Clarifying the the nature of these discrepancies may require experimental work in more controlled settings. Lastly, despite the fact that our approach appeared to deal gracefully  with objects having different structural characteristics, these were limited to the domain of video games. In order to verify the generalizability of our approach, future work should include data generated from a variety of contexts (e.g. web services, online and laboratory-based experiments).

%%%%%%%%%%%%%%%%%%%%%%%%%%%%%%%%%%%%%%%%%%%%%%%%%%%%%%%%%%%%%%%%%%%%%%

\section*{Declarations}

\paragraph{\textbf{Ethical Approval}}
All data were obtained and processed in compliance with the European Union's General Data Protection Regulation \citep{EUdataregulations2018}.

\paragraph{\textbf{Consent to Participatel}}
Not Applicable.

\paragraph{\textbf{Consent to Publish}}
Not Applicable.

\paragraph{\textbf{Funding}}
This work was supported by the EPSRC Centre for Doctoral Training in Intelligent Games \& Games Intelligence (IGGI) [EP/L015846/1].

\paragraph{\textbf{Availability of Data and Material}}
Due to commercial and privacy concerns we cannot share the raw data used for fitting the models nor we can provide any derived transformations (e.g. models' weights or learned representations). The data used for conducting the statistical analysis on the models' performance are available at the following GitHub repository \url{https://github.com/vb690/approx\_incentive\_salience}.

\paragraph{\textbf{Code Availability}}
The code used for the present paper, except for the SQL queries employed for retrieving the data, is available at the following GitHub repository \url{https://github.com/vb690/approx\_incentive\_salience}.

\paragraph{\textbf{Conflict of interest}}
\textbf{VB} at the time of writing was developing his doctoral thesis in partnership with the company providing the data while \textbf{MR} was employed by the  same company. No further conflicts of interest could be identified by the authors. 

\paragraph{\textbf{Authors Contribution}}
% this is from https://www.elsevier.com/authors/policies-and-guidelines/credit-author-statement
\textbf{VB}: Conceptualization, Methodology, Software, Formal analysis,
Investigation, Data Curation, Writing Original Draft, Visualization.  
\textbf{MR}: Resources, Data Curation,  Supervision.
\textbf{AD}: Resources, Supervision, Project Administration, Funding Acquisition.
\textbf{AW}: Supervision, Writing Original Draft, Methodology, Project Administration.
All authors approved the final manuscript.

\paragraph{\textbf{Acknowledgements}}
We would like to thank Charles Ringer and Ivan Bravi for the invaluable insights provided during the realization of this work and \textit{Square Enix Ltd.} Analytics and Insight Team for the continuous support.

% BibTeX users please use one of
%\bibliographystyle{spbasic}   % basic style, author-year citations
%\bibliographystyle{spmpsci}   % mathematics and physical sciences
%\bibliographystyle{unsrt}
%\bibliographystyle{spphys}    % APS-like style for physics
%\clearpage
\bibliography{bibliography}

\begin{thebibliography}{117}
\providecommand{\natexlab}[1]{#1}
\providecommand{\url}[1]{\texttt{#1}}
\expandafter\ifx\csname urlstyle\endcsname\relax
  \providecommand{\doi}[1]{doi: #1}\else
  \providecommand{\doi}{doi: \begingroup \urlstyle{rm}\Url}\fi

\bibitem[EUd()]{EUdataregulations2018}
2018 reform of eu data protection rules.
\newblock URL
  \url{https://ec.europa.eu/commission/sites/beta-political/files/data-protection-factsheet-changes_en.pdf}.

\bibitem[ali()]{alignedumap}
Alignedumap.
\newblock
  \url{https://umap-learn.readthedocs.io/en/latest/aligned_umap_basic_usage.html}.
\newblock Accessed: 2021-04-30.

\bibitem[Abadi et~al.(2015)Abadi, Agarwal, Barham, Brevdo, Chen, Citro,
  Corrado, Davis, Dean, Devin, Ghemawat, Goodfellow, Harp, Irving, Isard, Jia,
  Jozefowicz, Kaiser, Kudlur, Levenberg, Man\'{e}, Monga, Moore, Murray, Olah,
  Schuster, Shlens, Steiner, Sutskever, Talwar, Tucker, Vanhoucke, Vasudevan,
  Vi\'{e}gas, Vinyals, Warden, Wattenberg, Wicke, Yu, and
  Zheng]{tensorflow2015-whitepaper}
M.~Abadi, A.~Agarwal, P.~Barham, E.~Brevdo, Z.~Chen, C.~Citro, G.~S. Corrado,
  A.~Davis, J.~Dean, M.~Devin, S.~Ghemawat, I.~Goodfellow, A.~Harp, G.~Irving,
  M.~Isard, Y.~Jia, R.~Jozefowicz, L.~Kaiser, M.~Kudlur, J.~Levenberg,
  D.~Man\'{e}, R.~Monga, S.~Moore, D.~Murray, C.~Olah, M.~Schuster, J.~Shlens,
  B.~Steiner, I.~Sutskever, K.~Talwar, P.~Tucker, V.~Vanhoucke, V.~Vasudevan,
  F.~Vi\'{e}gas, O.~Vinyals, P.~Warden, M.~Wattenberg, M.~Wicke, Y.~Yu, and
  X.~Zheng.
\newblock {TensorFlow}: Large-scale machine learning on heterogeneous systems,
  2015.
\newblock URL \url{http://tensorflow.org/}.
\newblock Software available from tensorflow.org.

\bibitem[Agarwal et~al.(2017)Agarwal, Burghardt, and
  Lerman]{agarwal2017quitting}
T.~Agarwal, K.~Burghardt, and K.~Lerman.
\newblock On quitting: performance and practice in online game play.
\newblock In \emph{Proceedings of the International AAAI Conference on Web and
  Social Media}, volume~11, pages 452--455, 2017.

\bibitem[Albanese et~al.(2013)Albanese, Filosi, Visintainer, Riccadonna,
  Jurman, and Furlanello]{albanese2013minerva}
D.~Albanese, M.~Filosi, R.~Visintainer, S.~Riccadonna, G.~Jurman, and
  C.~Furlanello.
\newblock Minerva and minepy: a c engine for the mine suite and its r, python
  and matlab wrappers.
\newblock \emph{Bioinformatics}, 29\penalty0 (3):\penalty0 407--408, 2013.

\bibitem[Andy~Coenen()]{umapwebs}
A.~P. Andy~Coenen.
\newblock Understanding umap.
\newblock URL \url{https://pair-code.github.io/understanding-umap/}.

\bibitem[Armony and Vuilleumier(2013)]{armony2013cambridge}
J.~Armony and P.~Vuilleumier.
\newblock \emph{The Cambridge handbook of human affective neuroscience}.
\newblock Cambridge university press, 2013.

\bibitem[A{\v{s}}eri{\v{s}}kis and
  Dama{\v{s}}evi{\v{c}}ius(2017)]{avserivskis2017computational}
D.~A{\v{s}}eri{\v{s}}kis and R.~Dama{\v{s}}evi{\v{c}}ius.
\newblock Computational evaluation of effects of motivation reinforcement on
  player retention.
\newblock \emph{Journal of Universal Computer Science}, 23\penalty0
  (5):\penalty0 432--453, 2017.

\bibitem[Barak(2017)]{barak2017recurrent}
O.~Barak.
\newblock Recurrent neural networks as versatile tools of neuroscience
  research.
\newblock \emph{Current opinion in neurobiology}, 46:\penalty0 1--6, 2017.

\bibitem[Barto and Dietterich(2004)]{barto2004reinforcement}
A.~G. Barto and T.~G. Dietterich.
\newblock Reinforcement learning and its relationship to supervised learning.
\newblock \emph{Handbook of learning and approximate dynamic programming},
  10:\penalty0 9780470544785, 2004.

\bibitem[Bauckhage et~al.(2012)Bauckhage, Kersting, Sifa, Thurau, Drachen, and
  Canossa]{bauckhage2012players}
C.~Bauckhage, K.~Kersting, R.~Sifa, C.~Thurau, A.~Drachen, and A.~Canossa.
\newblock How players lose interest in playing a game: An empirical study based
  on distributions of total playing times.
\newblock In \emph{2012 IEEE Conference on Computational Intelligence and Games
  (CIG)}, pages 139--146. IEEE, 2012.

\bibitem[Bengio et~al.(2017)Bengio, Goodfellow, and Courville]{bengio2017deep}
Y.~Bengio, I.~Goodfellow, and A.~Courville.
\newblock \emph{Deep learning}, volume~1.
\newblock MIT press Massachusetts, USA:, 2017.

\bibitem[Berridge(2004)]{berridge2004motivation}
K.~C. Berridge.
\newblock Motivation concepts in behavioral neuroscience.
\newblock \emph{Physiology \& behavior}, 81\penalty0 (2):\penalty0 179--209,
  2004.

\bibitem[Berridge(2012)]{berridge2012prediction}
K.~C. Berridge.
\newblock From prediction error to incentive salience: mesolimbic computation
  of reward motivation.
\newblock \emph{European Journal of Neuroscience}, 35\penalty0 (7):\penalty0
  1124--1143, 2012.

\bibitem[Berridge and Kringelbach(2008)]{berridge2008affective}
K.~C. Berridge and M.~L. Kringelbach.
\newblock Affective neuroscience of pleasure: reward in humans and animals.
\newblock \emph{Psychopharmacology}, 199\penalty0 (3):\penalty0 457--480, 2008.

\bibitem[Berridge and Robinson(1998)]{berridge1998role}
K.~C. Berridge and T.~E. Robinson.
\newblock What is the role of dopamine in reward: hedonic impact, reward
  learning, or incentive salience?
\newblock \emph{Brain research reviews}, 28\penalty0 (3):\penalty0 309--369,
  1998.

\bibitem[Berridge et~al.(2009)Berridge, Robinson, and
  Aldridge]{berridge2009dissecting}
K.~C. Berridge, T.~E. Robinson, and J.~W. Aldridge.
\newblock Dissecting components of reward:‘liking’,‘wanting’, and
  learning.
\newblock \emph{Current opinion in pharmacology}, 9\penalty0 (1):\penalty0
  65--73, 2009.

\bibitem[Bindra(1978)]{bindra1978adaptive}
D.~Bindra.
\newblock How adaptive behavior is produced: A perceptual-motivational
  alternative to response-reinforcement.
\newblock \emph{Behavioral and brain sciences}, 1\penalty0 (1):\penalty0
  41--91, 1978.

\bibitem[Bishop(2006)]{bishop2006pattern}
C.~M. Bishop.
\newblock \emph{Pattern recognition and machine learning}.
\newblock springer, 2006.

\bibitem[Bolles(1972)]{bolles1972reinforcement}
R.~C. Bolles.
\newblock Reinforcement, expectancy, and learning.
\newblock \emph{Psychological review}, 79\penalty0 (5):\penalty0 394, 1972.

\bibitem[Boyle et~al.(2012)Boyle, Connolly, Hainey, and
  Boyle]{boyle2012engagement}
E.~A. Boyle, T.~M. Connolly, T.~Hainey, and J.~M. Boyle.
\newblock Engagement in digital entertainment games: A systematic review.
\newblock \emph{Computers in human behavior}, 28\penalty0 (3):\penalty0
  771--780, 2012.

\bibitem[Bromberg-Martin et~al.(2010)Bromberg-Martin, Hikosaka, and
  Nakamura]{bromberg2010coding}
E.~S. Bromberg-Martin, O.~Hikosaka, and K.~Nakamura.
\newblock Coding of task reward value in the dorsal raphe nucleus.
\newblock \emph{Journal of Neuroscience}, 30\penalty0 (18):\penalty0
  6262--6272, 2010.

\bibitem[Calhoun et~al.(2019)Calhoun, Pillow, and
  Murthy]{calhoun2019unsupervised}
A.~J. Calhoun, J.~W. Pillow, and M.~Murthy.
\newblock Unsupervised identification of the internal states that shape natural
  behavior.
\newblock \emph{Nature neuroscience}, 22\penalty0 (12):\penalty0 2040--2049,
  2019.

\bibitem[Chollet et~al.(2015)]{chollet2015keras}
F.~Chollet et~al.
\newblock Keras.
\newblock \url{https://keras.io}, 2015.

\bibitem[Chumbley and Griffiths(2006)]{chumbley2006affect}
J.~Chumbley and M.~Griffiths.
\newblock Affect and the computer game player: the effect of gender,
  personality, and game reinforcement structure on affective responses to
  computer game-play.
\newblock \emph{CyberPsychology \& Behavior}, 9\penalty0 (3):\penalty0
  308--316, 2006.

\bibitem[Cole et~al.(2012)Cole, Yoo, and Knutson]{cole2012interactivity}
S.~W. Cole, D.~J. Yoo, and B.~Knutson.
\newblock Interactivity and reward-related neural activation during a serious
  videogame.
\newblock \emph{PLoS one}, 7\penalty0 (3):\penalty0 e33909, 2012.

\bibitem[Corbit and Balleine(2015)]{corbit2015learning}
L.~H. Corbit and B.~W. Balleine.
\newblock Learning and motivational processes contributing to
  pavlovian--instrumental transfer and their neural bases: dopamine and beyond.
\newblock In \emph{Behavioral neuroscience of motivation}, pages 259--289.
  Springer, 2015.

\bibitem[Derdikman and Moser(2011)]{derdikman2011manifold}
D.~Derdikman and E.~I. Moser.
\newblock A manifold of spatial maps in the brain.
\newblock \emph{Space, Time and Number in the Brain}, pages 41--57, 2011.

\bibitem[Drachen(2015)]{drachen2015behavioral}
A.~Drachen.
\newblock Behavioral telemetry in games user research.
\newblock In \emph{Game User Experience Evaluation}, pages 135--165. Springer,
  2015.

\bibitem[Drummond and Sauer(2018)]{drummond2018video}
A.~Drummond and J.~D. Sauer.
\newblock Video game loot boxes are psychologically akin to gambling.
\newblock \emph{Nature Human Behaviour}, 2\penalty0 (8):\penalty0 530--532,
  2018.

\bibitem[El-Nasr et~al.(2016)El-Nasr, Drachen, and Canossa]{el2016game}
M.~S. El-Nasr, A.~Drachen, and A.~Canossa.
\newblock \emph{Game analytics}.
\newblock Springer, 2016.

\bibitem[Eyjolfsdottir et~al.(2016)Eyjolfsdottir, Branson, Yue, and
  Perona]{eyjolfsdottir2016learning}
E.~Eyjolfsdottir, K.~Branson, Y.~Yue, and P.~Perona.
\newblock Learning recurrent representations for hierarchical behavior
  modeling.
\newblock \emph{arXiv preprint arXiv:1611.00094}, 2016.

\bibitem[Flagel et~al.(2011)Flagel, Clark, Robinson, Mayo, Czuj, Willuhn,
  Akers, Clinton, Phillips, and Akil]{flagel2011selective}
S.~B. Flagel, J.~J. Clark, T.~E. Robinson, L.~Mayo, A.~Czuj, I.~Willuhn, C.~A.
  Akers, S.~M. Clinton, P.~E. Phillips, and H.~Akil.
\newblock A selective role for dopamine in stimulus--reward learning.
\newblock \emph{Nature}, 469\penalty0 (7328):\penalty0 53--57, 2011.

\bibitem[Gallego et~al.(2017)Gallego, Perich, Miller, and
  Solla]{gallego2017neural}
J.~A. Gallego, M.~G. Perich, L.~E. Miller, and S.~A. Solla.
\newblock Neural manifolds for the control of movement.
\newblock \emph{Neuron}, 94\penalty0 (5):\penalty0 978--984, 2017.

\bibitem[Ganmor et~al.(2015)Ganmor, Segev, and Schneidman]{ganmor2015thesaurus}
E.~Ganmor, R.~Segev, and E.~Schneidman.
\newblock A thesaurus for a neural population code.
\newblock \emph{Elife}, 4:\penalty0 e06134, 2015.

\bibitem[Gao et~al.(2021)Gao, Mishne, and Scheinost]{gao2021nonlinear}
S.~Gao, G.~Mishne, and D.~Scheinost.
\newblock Nonlinear manifold learning in functional magnetic resonance imaging
  uncovers a low-dimensional space of brain dynamics.
\newblock \emph{Human brain mapping}, 42\penalty0 (14):\penalty0 4510--4524,
  2021.

\bibitem[Gleich et~al.(2017)Gleich, Lorenz, Gallinat, and
  K{\"u}hn]{gleich2017functional}
T.~Gleich, R.~C. Lorenz, J.~Gallinat, and S.~K{\"u}hn.
\newblock Functional changes in the reward circuit in response to
  gaming-related cues after training with a commercial video game.
\newblock \emph{Neuroimage}, 152:\penalty0 467--475, 2017.

\bibitem[Harris et~al.(2020)Harris, Millman, van~der Walt, Gommers, Virtanen,
  Cournapeau, Wieser, Taylor, Berg, Smith, Kern, Picus, Hoyer, van Kerkwijk,
  Brett, Haldane, del R{'{\i}}o, Wiebe, Peterson, G{'{e}}rard-Marchant,
  Sheppard, Reddy, Weckesser, Abbasi, Gohlke, and Oliphant]{harris2020array}
C.~R. Harris, K.~J. Millman, S.~J. van~der Walt, R.~Gommers, P.~Virtanen,
  D.~Cournapeau, E.~Wieser, J.~Taylor, S.~Berg, N.~J. Smith, R.~Kern, M.~Picus,
  S.~Hoyer, M.~H. van Kerkwijk, M.~Brett, A.~Haldane, J.~F. del R{'{\i}}o,
  M.~Wiebe, P.~Peterson, P.~G{'{e}}rard-Marchant, K.~Sheppard, T.~Reddy,
  W.~Weckesser, H.~Abbasi, C.~Gohlke, and T.~E. Oliphant.
\newblock Array programming with {NumPy}.
\newblock \emph{Nature}, 585\penalty0 (7825):\penalty0 357--362, Sept. 2020.
\newblock \doi{10.1038/s41586-020-2649-2}.
\newblock URL \url{https://doi.org/10.1038/s41586-020-2649-2}.

\bibitem[Hashem et~al.(2015)Hashem, Yaqoob, Anuar, Mokhtar, Gani, and
  Khan]{hashem2015rise}
I.~A.~T. Hashem, I.~Yaqoob, N.~B. Anuar, S.~Mokhtar, A.~Gani, and S.~U. Khan.
\newblock The rise of “big data” on cloud computing: Review and open
  research issues.
\newblock \emph{Information systems}, 47:\penalty0 98--115, 2015.

\bibitem[Hochreiter and Schmidhuber(1997)]{hochreiter1997long}
S.~Hochreiter and J.~Schmidhuber.
\newblock Long short-term memory.
\newblock \emph{Neural computation}, 9\penalty0 (8):\penalty0 1735--1780, 1997.

\bibitem[Hoeft et~al.(2008)Hoeft, Watson, Kesler, Bettinger, and
  Reiss]{hoeft2008gender}
F.~Hoeft, C.~L. Watson, S.~R. Kesler, K.~E. Bettinger, and A.~L. Reiss.
\newblock Gender differences in the mesocorticolimbic system during computer
  game-play.
\newblock \emph{Journal of psychiatric research}, 42\penalty0 (4):\penalty0
  253--258, 2008.

\bibitem[Hornik et~al.(1989)Hornik, Stinchcombe, White,
  et~al.]{hornik1989multilayer}
K.~Hornik, M.~Stinchcombe, H.~White, et~al.
\newblock Multilayer feedforward networks are universal approximators.
\newblock \emph{Neural networks}, 2\penalty0 (5):\penalty0 359--366, 1989.

\bibitem[Hunter(2007)]{hunter2007matplotlib}
J.~D. Hunter.
\newblock Matplotlib: A 2d graphics environment.
\newblock \emph{IEEE Annals of the History of Computing}, 9\penalty0
  (03):\penalty0 90--95, 2007.

\bibitem[Hyndman and Athanasopoulos(2018)]{hyndman2018forecasting}
R.~J. Hyndman and G.~Athanasopoulos.
\newblock \emph{Forecasting: principles and practice}.
\newblock OTexts, 2018.

\bibitem[Ikemoto and Panksepp(1996)]{ikemoto1996dissociations}
S.~Ikemoto and J.~Panksepp.
\newblock Dissociations between appetitive and consummatory responses by
  pharmacological manipulations of reward-relevant brain regions.
\newblock \emph{Behavioral neuroscience}, 110\penalty0 (2):\penalty0 331, 1996.

\bibitem[Ikemoto and Panksepp(1999)]{ikemoto1999role}
S.~Ikemoto and J.~Panksepp.
\newblock The role of nucleus accumbens dopamine in motivated behavior: a
  unifying interpretation with special reference to reward-seeking.
\newblock \emph{Brain Research Reviews}, 31\penalty0 (1):\penalty0 6--41, 1999.

\bibitem[Jamieson()]{hyperwebs}
K.~Jamieson.
\newblock Hyperband: A novel bandit-based approach to hyperparameter
  optimization.
\newblock URL \url{https://homes.cs.washington.edu/~jamieson/hyperband.html}.

\bibitem[Kietzmann et~al.(2018)Kietzmann, McClure, and
  Kriegeskorte]{kietzmann2018deep}
T.~C. Kietzmann, P.~McClure, and N.~Kriegeskorte.
\newblock Deep neural networks in computational neuroscience.
\newblock \emph{BioRxiv}, page 133504, 2018.

\bibitem[King et~al.(2010{\natexlab{a}})King, Delfabbro, and
  Griffiths]{king2010role}
D.~King, P.~Delfabbro, and M.~Griffiths.
\newblock The role of structural characteristics in problem video game playing:
  A review.
\newblock \emph{Cyberpsychology: Journal of Psychosocial Research on
  Cyberspace}, 4\penalty0 (1), 2010{\natexlab{a}}.

\bibitem[King et~al.(2010{\natexlab{b}})King, Delfabbro, and
  Griffiths]{king2010video}
D.~King, P.~Delfabbro, and M.~Griffiths.
\newblock Video game structural characteristics: A new psychological taxonomy.
\newblock \emph{International journal of mental health and addiction},
  8\penalty0 (1):\penalty0 90--106, 2010{\natexlab{b}}.

\bibitem[Kingma and Ba(2014)]{kingma2014adam}
D.~P. Kingma and J.~Ba.
\newblock Adam: A method for stochastic optimization.
\newblock \emph{arXiv preprint arXiv:1412.6980}, 2014.

\bibitem[Klasen et~al.(2012)Klasen, Weber, Kircher, Mathiak, and
  Mathiak]{klasen2012neural}
M.~Klasen, R.~Weber, T.~T. Kircher, K.~A. Mathiak, and K.~Mathiak.
\newblock Neural contributions to flow experience during video game playing.
\newblock \emph{Social cognitive and affective neuroscience}, 7\penalty0
  (4):\penalty0 485--495, 2012.

\bibitem[Koepp et~al.(1998)Koepp, Gunn, Lawrence, Cunningham, Dagher, Jones,
  Brooks, Bench, and Grasby]{koepp1998evidence}
M.~J. Koepp, R.~N. Gunn, A.~D. Lawrence, V.~J. Cunningham, A.~Dagher, T.~Jones,
  D.~J. Brooks, C.~J. Bench, and P.~Grasby.
\newblock Evidence for striatal dopamine release during a video game.
\newblock \emph{Nature}, 393\penalty0 (6682):\penalty0 266--268, 1998.

\bibitem[Li et~al.(2017)Li, Jamieson, DeSalvo, Rostamizadeh, and
  Talwalkar]{li2017hyperband}
L.~Li, K.~Jamieson, G.~DeSalvo, A.~Rostamizadeh, and A.~Talwalkar.
\newblock Hyperband: A novel bandit-based approach to hyperparameter
  optimization.
\newblock \emph{The Journal of Machine Learning Research}, 18\penalty0
  (1):\penalty0 6765--6816, 2017.

\bibitem[Lillicrap and Santoro(2019)]{lillicrap2019backpropagation}
T.~P. Lillicrap and A.~Santoro.
\newblock Backpropagation through time and the brain.
\newblock \emph{Current opinion in neurobiology}, 55:\penalty0 82--89, 2019.

\bibitem[Lillicrap et~al.(2020)Lillicrap, Santoro, Marris, Akerman, and
  Hinton]{lillicrap2020backpropagation}
T.~P. Lillicrap, A.~Santoro, L.~Marris, C.~J. Akerman, and G.~Hinton.
\newblock Backpropagation and the brain.
\newblock \emph{Nature Reviews Neuroscience}, pages 1--12, 2020.

\bibitem[Lorenz et~al.(2015)Lorenz, Gleich, Gallinat, and
  K{\"u}hn]{lorenz2015video}
R.~C. Lorenz, T.~Gleich, J.~Gallinat, and S.~K{\"u}hn.
\newblock Video game training and the reward system.
\newblock \emph{Frontiers in human neuroscience}, 9:\penalty0 40, 2015.

\bibitem[Luxem et~al.(2020)Luxem, Fuhrmann, K{\"u}rsch, Remy, and
  Bauer]{luxem2020identifying}
K.~Luxem, F.~Fuhrmann, J.~K{\"u}rsch, S.~Remy, and P.~Bauer.
\newblock Identifying behavioral structure from deep variational embeddings of
  animal motion.
\newblock \emph{BioRxiv}, 2020.

\bibitem[Mathiak et~al.(2011)Mathiak, Klasen, Weber, Ackermann, Shergill, and
  Mathiak]{mathiak2011reward}
K.~A. Mathiak, M.~Klasen, R.~Weber, H.~Ackermann, S.~S. Shergill, and
  K.~Mathiak.
\newblock Reward system and temporal pole contributions to affective evaluation
  during a first person shooter video game.
\newblock \emph{BMC neuroscience}, 12\penalty0 (1):\penalty0 1--11, 2011.

\bibitem[McClure et~al.(2003)McClure, Daw, and
  Montague]{mcclure2003computational}
S.~M. McClure, N.~D. Daw, and P.~R. Montague.
\newblock A computational substrate for incentive salience.
\newblock \emph{Trends in neurosciences}, 26\penalty0 (8):\penalty0 423--428,
  2003.

\bibitem[McCullough and Goodhill(2021)]{mccullough2021unsupervised}
M.~H. McCullough and G.~J. Goodhill.
\newblock Unsupervised quantification of naturalistic animal behaviors for
  gaining insight into the brain.
\newblock \emph{Current opinion in neurobiology}, 70:\penalty0 89--100, 2021.

\bibitem[{McInnes} et~al.(2018){McInnes}, {Healy}, and
  {Melville}]{2018arXivUMAP}
L.~{McInnes}, J.~{Healy}, and J.~{Melville}.
\newblock {UMAP: Uniform Manifold Approximation and Projection for Dimension
  Reduction}.
\newblock \emph{ArXiv e-prints}, Feb. 2018.

\bibitem[McInnes et~al.(2018)McInnes, Healy, Saul, and
  Grossberger]{mcinnes2018umap-software}
L.~McInnes, J.~Healy, N.~Saul, and L.~Grossberger.
\newblock Umap: Uniform manifold approximation and projection.
\newblock \emph{The Journal of Open Source Software}, 3\penalty0 (29):\penalty0
  861, 2018.

\bibitem[Merel et~al.(2019)Merel, Aldarondo, Marshall, Tassa, Wayne, and
  {\"O}lveczky]{merel2019deep}
J.~Merel, D.~Aldarondo, J.~Marshall, Y.~Tassa, G.~Wayne, and B.~{\"O}lveczky.
\newblock Deep neuroethology of a virtual rodent.
\newblock \emph{arXiv preprint arXiv:1911.09451}, 2019.

\bibitem[Meyer et~al.(2015)Meyer, King, and Ferrario]{meyer2015motivational}
P.~J. Meyer, C.~P. King, and C.~R. Ferrario.
\newblock Motivational processes underlying substance abuse disorder.
\newblock In \emph{Behavioral Neuroscience of Motivation}, pages 473--506.
  Springer, 2015.

\bibitem[Nevin and Grace(2000)]{nevin2000behavioral}
J.~A. Nevin and R.~C. Grace.
\newblock Behavioral momentum and the law of effect.
\newblock \emph{Behavioral and Brain Sciences}, 23\penalty0 (1):\penalty0
  73--90, 2000.

\bibitem[Nieh et~al.(2021)Nieh, Schottdorf, Freeman, Low, Lewallen, Koay,
  Pinto, Gauthier, Brody, and Tank]{nieh2021geometry}
E.~H. Nieh, M.~Schottdorf, N.~W. Freeman, R.~J. Low, S.~Lewallen, S.~A. Koay,
  L.~Pinto, J.~L. Gauthier, C.~D. Brody, and D.~W. Tank.
\newblock Geometry of abstract learned knowledge in the hippocampus.
\newblock \emph{Nature}, 595\penalty0 (7865):\penalty0 80--84, 2021.

\bibitem[O'Doherty et~al.(2003)O'Doherty, Dayan, Friston, Critchley, and
  Dolan]{o2003temporal}
J.~P. O'Doherty, P.~Dayan, K.~Friston, H.~Critchley, and R.~J. Dolan.
\newblock Temporal difference models and reward-related learning in the human
  brain.
\newblock \emph{Neuron}, 38\penalty0 (2):\penalty0 329--337, 2003.

\bibitem[Oh and Jung(2004)]{oh2004gpu}
K.-S. Oh and K.~Jung.
\newblock Gpu implementation of neural networks.
\newblock \emph{Pattern Recognition}, 37\penalty0 (6):\penalty0 1311--1314,
  2004.

\bibitem[O'Malley et~al.(2019)O'Malley, Bursztein, Long, Chollet, Jin,
  Invernizzi, et~al.]{omalley2019kerastuner}
T.~O'Malley, E.~Bursztein, J.~Long, F.~Chollet, H.~Jin, L.~Invernizzi, et~al.
\newblock Keras {Tuner}.
\newblock \url{https://github.com/keras-team/keras-tuner}, 2019.

\bibitem[Palminteri et~al.(2015)Palminteri, Khamassi, Joffily, and
  Coricelli]{palminteri2015contextual}
S.~Palminteri, M.~Khamassi, M.~Joffily, and G.~Coricelli.
\newblock Contextual modulation of value signals in reward and punishment
  learning.
\newblock \emph{Nature communications}, 6\penalty0 (1):\penalty0 1--14, 2015.

\bibitem[pandas~development team(2020)]{reback2020pandas}
T.~pandas~development team.
\newblock pandas-dev/pandas: Pandas, Feb. 2020.
\newblock URL \url{https://doi.org/10.5281/zenodo.3509134}.

\bibitem[Pang et~al.(2016)Pang, Lansdell, and Fairhall]{pang2016dimensionality}
R.~Pang, B.~J. Lansdell, and A.~L. Fairhall.
\newblock Dimensionality reduction in neuroscience.
\newblock \emph{Current Biology}, 26\penalty0 (14):\penalty0 R656--R660, 2016.

\bibitem[Pedregosa et~al.(2011)Pedregosa, Varoquaux, Gramfort, Michel, Thirion,
  Grisel, Blondel, Prettenhofer, Weiss, Dubourg, Vanderplas, Passos,
  Cournapeau, Brucher, Perrot, and Duchesnay]{scikit-learn}
F.~Pedregosa, G.~Varoquaux, A.~Gramfort, V.~Michel, B.~Thirion, O.~Grisel,
  M.~Blondel, P.~Prettenhofer, R.~Weiss, V.~Dubourg, J.~Vanderplas, A.~Passos,
  D.~Cournapeau, M.~Brucher, M.~Perrot, and E.~Duchesnay.
\newblock Scikit-learn: Machine learning in {P}ython.
\newblock \emph{Journal of Machine Learning Research}, 12:\penalty0 2825--2830,
  2011.

\bibitem[Pereira et~al.(2020)Pereira, Shaevitz, and
  Murthy]{pereira2020quantifying}
T.~D. Pereira, J.~W. Shaevitz, and M.~Murthy.
\newblock Quantifying behavior to understand the brain.
\newblock \emph{Nature neuroscience}, 23\penalty0 (12):\penalty0 1537--1549,
  2020.

\bibitem[Phillips et~al.(2013)Phillips, Johnson, and
  Wyeth]{phillips2013videogame}
C.~Phillips, D.~Johnson, and P.~Wyeth.
\newblock Videogame reward types.
\newblock In \emph{Proceedings of the First International Conference on Gameful
  Design, Research, and Applications}, pages 103--106. ACM, 2013.

\bibitem[Reshef et~al.(2011)Reshef, Reshef, Finucane, Grossman, McVean,
  Turnbaugh, Lander, Mitzenmacher, and Sabeti]{reshef2011detecting}
D.~N. Reshef, Y.~A. Reshef, H.~K. Finucane, S.~R. Grossman, G.~McVean, P.~J.
  Turnbaugh, E.~S. Lander, M.~Mitzenmacher, and P.~C. Sabeti.
\newblock Detecting novel associations in large data sets.
\newblock \emph{science}, 334\penalty0 (6062):\penalty0 1518--1524, 2011.

\bibitem[Robinson et~al.(2015)Robinson, Fischer, Ahuja, Lesser, and
  Maniates]{robinson2015roles}
M.~J.~F. Robinson, A.~Fischer, A.~Ahuja, E.~Lesser, and H.~Maniates.
\newblock Roles of “wanting” and “liking” in motivating behavior:
  gambling, food, and drug addictions.
\newblock In \emph{Behavioral neuroscience of motivation}, pages 105--136.
  Springer, 2015.

\bibitem[Robinson and Berridge(1993)]{robinson1993neural}
T.~E. Robinson and K.~C. Berridge.
\newblock The neural basis of drug craving: an incentive-sensitization theory
  of addiction.
\newblock \emph{Brain research reviews}, 18\penalty0 (3):\penalty0 247--291,
  1993.

\bibitem[Ru{\'e}-Queralt et~al.(2021)Ru{\'e}-Queralt, Stevner, Tagliazucchi,
  Laufs, Kringelbach, Deco, and Atasoy]{rue2021decoding}
J.~Ru{\'e}-Queralt, A.~Stevner, E.~Tagliazucchi, H.~Laufs, M.~L. Kringelbach,
  G.~Deco, and S.~Atasoy.
\newblock Decoding brain states on the intrinsic manifold of human brain
  dynamics across wakefulness and sleep.
\newblock \emph{Communications Biology}, 4\penalty0 (1):\penalty0 1--11, 2021.

\bibitem[Rumelhart et~al.(1986)Rumelhart, Hinton, and
  Williams]{rumelhart1986learning}
D.~E. Rumelhart, G.~E. Hinton, and R.~J. Williams.
\newblock Learning representations by back-propagating errors.
\newblock \emph{nature}, 323\penalty0 (6088):\penalty0 533--536, 1986.

\bibitem[Salamone and Correa(2002)]{salamone2002motivational}
J.~D. Salamone and M.~Correa.
\newblock Motivational views of reinforcement: implications for understanding
  the behavioral functions of nucleus accumbens dopamine.
\newblock \emph{Behavioural brain research}, 137\penalty0 (1-2):\penalty0
  3--25, 2002.

\bibitem[Satopaa et~al.(2011)Satopaa, Albrecht, Irwin, and
  Raghavan]{satopaa2011finding}
V.~Satopaa, J.~Albrecht, D.~Irwin, and B.~Raghavan.
\newblock Finding a" dle" in a haystack: Detecting knee points in system
  behavior.
\newblock In \emph{2011 31st international conference on distributed computing
  systems workshops}, pages 166--171. IEEE, 2011.

\bibitem[Schultz(2000)]{schultz2000multiple}
W.~Schultz.
\newblock Multiple reward signals in the brain.
\newblock \emph{Nature reviews neuroscience}, 1\penalty0 (3):\penalty0
  199--207, 2000.

\bibitem[Schultz(2017)]{schultz2017reward}
W.~Schultz.
\newblock Reward prediction error.
\newblock \emph{Current Biology}, 27\penalty0 (10):\penalty0 R369--R371, 2017.

\bibitem[Schultz et~al.(1997)Schultz, Dayan, and Montague]{schultz1997neural}
W.~Schultz, P.~Dayan, and P.~R. Montague.
\newblock A neural substrate of prediction and reward.
\newblock \emph{Science}, 275\penalty0 (5306):\penalty0 1593--1599, 1997.

\bibitem[Schultz et~al.(2000)Schultz, Tremblay, and
  Hollerman]{schultz2000reward}
W.~Schultz, L.~Tremblay, and J.~R. Hollerman.
\newblock Reward processing in primate orbitofrontal cortex and basal ganglia.
\newblock \emph{Cerebral cortex}, 10\penalty0 (3):\penalty0 272--283, 2000.

\bibitem[Schuster-B{\"o}ckler and Bateman(2007)]{schuster2007introduction}
B.~Schuster-B{\"o}ckler and A.~Bateman.
\newblock An introduction to hidden markov models.
\newblock \emph{Current protocols in bioinformatics}, 18\penalty0 (1):\penalty0
  A--3A, 2007.

\bibitem[Sculley(2010)]{sculley2010web}
D.~Sculley.
\newblock Web-scale k-means clustering.
\newblock In \emph{Proceedings of the 19th international conference on World
  wide web}, pages 1177--1178, 2010.

\bibitem[Seabold and Perktold(2010)]{seabold2010statsmodels}
S.~Seabold and J.~Perktold.
\newblock statsmodels: Econometric and statistical modeling with python.
\newblock In \emph{9th Python in Science Conference}, 2010.

\bibitem[Servén and Brummitt(2018)]{serven2018}
D.~Servén and C.~Brummitt.
\newblock pygam: Generalized additive models in python, Mar. 2018.
\newblock URL \url{https://doi.org/10.5281/zenodo.1208723}.

\bibitem[Sescousse et~al.(2013)Sescousse, Cald{\'u}, Segura, and
  Dreher]{sescousse2013processing}
G.~Sescousse, X.~Cald{\'u}, B.~Segura, and J.-C. Dreher.
\newblock Processing of primary and secondary rewards: a quantitative
  meta-analysis and review of human functional neuroimaging studies.
\newblock \emph{Neuroscience \& Biobehavioral Reviews}, 37\penalty0
  (4):\penalty0 681--696, 2013.

\bibitem[Seung and Lee(2000)]{seung2000manifold}
H.~S. Seung and D.~D. Lee.
\newblock The manifold ways of perception.
\newblock \emph{science}, 290\penalty0 (5500):\penalty0 2268--2269, 2000.

\bibitem[Shi et~al.(2021)Shi, Schwartz, Levy, Achvat, Abboud, Ghanayim,
  Schiller, and Mishne]{shi2021learning}
C.~Shi, S.~Schwartz, S.~Levy, S.~Achvat, M.~Abboud, A.~Ghanayim, J.~Schiller,
  and G.~Mishne.
\newblock Learning disentangled behavior embeddings.
\newblock \emph{Advances in Neural Information Processing Systems}, 34, 2021.

\bibitem[Simpson and Balsam(2016)]{simpson2016behavioral}
E.~H. Simpson and P.~D. Balsam.
\newblock \emph{Behavioral neuroscience of motivation}.
\newblock Springer, 2016.

\bibitem[Skinner(1965)]{skinner1965science}
B.~F. Skinner.
\newblock \emph{Science and human behavior}.
\newblock Number 92904. Simon and Schuster, 1965.

\bibitem[Smith et~al.(2011)Smith, Berridge, and
  Aldridge]{smith2011disentangling}
K.~S. Smith, K.~C. Berridge, and J.~W. Aldridge.
\newblock Disentangling pleasure from incentive salience and learning signals
  in brain reward circuitry.
\newblock \emph{Proceedings of the National Academy of Sciences}, 108\penalty0
  (27):\penalty0 E255--E264, 2011.

\bibitem[Song et~al.(2017)Song, Yang, and Wang]{song2017reward}
H.~F. Song, G.~R. Yang, and X.-J. Wang.
\newblock Reward-based training of recurrent neural networks for cognitive and
  value-based tasks.
\newblock \emph{Elife}, 6:\penalty0 e21492, 2017.

\bibitem[Spearman(1961)]{spearman1961general}
C.~Spearman.
\newblock " general intelligence" objectively determined and measured.
\newblock 1961.

\bibitem[Steyvers and Benjamin(2019)]{steyvers2019joint}
M.~Steyvers and A.~S. Benjamin.
\newblock The joint contribution of participation and performance to learning
  functions: Exploring the effects of age in large-scale data sets.
\newblock \emph{Behavior research methods}, 51\penalty0 (4):\penalty0
  1531--1543, 2019.

\bibitem[Stopfer et~al.(2003)Stopfer, Jayaraman, and
  Laurent]{stopfer2003intensity}
M.~Stopfer, V.~Jayaraman, and G.~Laurent.
\newblock Intensity versus identity coding in an olfactory system.
\newblock \emph{Neuron}, 39\penalty0 (6):\penalty0 991--1004, 2003.

\bibitem[Sutton(1988)]{sutton1988learning}
R.~S. Sutton.
\newblock Learning to predict by the methods of temporal differences.
\newblock \emph{Machine learning}, 3\penalty0 (1):\penalty0 9--44, 1988.

\bibitem[Sutton and Barto(2018)]{sutton2018reinforcement}
R.~S. Sutton and A.~G. Barto.
\newblock \emph{Reinforcement learning: An introduction}.
\newblock MIT press, 2018.

\bibitem[Thorndike(1927)]{thorndike1927law}
E.~L. Thorndike.
\newblock The law of effect.
\newblock \emph{The American journal of psychology}, 39\penalty0
  (1/4):\penalty0 212--222, 1927.

\bibitem[Tindell et~al.(2009)Tindell, Smith, Berridge, and
  Aldridge]{tindell2009dynamic}
A.~J. Tindell, K.~S. Smith, K.~C. Berridge, and J.~W. Aldridge.
\newblock Dynamic computation of incentive salience:“wanting” what was
  never “liked”.
\newblock \emph{Journal of Neuroscience}, 29\penalty0 (39):\penalty0
  12220--12228, 2009.

\bibitem[Toates(1994)]{toates1994comparing}
F.~Toates.
\newblock Comparing motivational systems—an incentive motivation perspective.
\newblock 1994.

\bibitem[Touloupou et~al.(2020)Touloupou, Finkenst{\"a}dt, and
  Spencer]{touloupou2020scalable}
P.~Touloupou, B.~Finkenst{\"a}dt, and S.~E. Spencer.
\newblock Scalable bayesian inference for coupled hidden markov and semi-markov
  models.
\newblock \emph{Journal of Computational and Graphical Statistics}, 29\penalty0
  (2):\penalty0 238--249, 2020.

\bibitem[Van~Rossum and Drake(2009)]{10.5555/1593511}
G.~Van~Rossum and F.~L. Drake.
\newblock \emph{Python 3 Reference Manual}.
\newblock CreateSpace, Scotts Valley, CA, 2009.
\newblock ISBN 1441412697.

\bibitem[Wang et~al.(2018{\natexlab{a}})Wang, Sun, and Zheng]{wang2018beyond}
B.~Wang, T.~Sun, and X.~S. Zheng.
\newblock Beyond winning and losing: modeling human motivations and behaviors
  using inverse reinforcement learning.
\newblock \emph{arXiv preprint arXiv:1807.00366}, 2018{\natexlab{a}}.

\bibitem[Wang and Sun(2011)]{wang2011game}
H.~Wang and C.-T. Sun.
\newblock Game reward systems: Gaming experiences and social meanings.
\newblock In \emph{DiGRA Conference}, pages 1--15, 2011.

\bibitem[Wang et~al.(2018{\natexlab{b}})Wang, Kurth-Nelson, Kumaran, Tirumala,
  Soyer, Leibo, Hassabis, and Botvinick]{wang2018prefrontal}
J.~X. Wang, Z.~Kurth-Nelson, D.~Kumaran, D.~Tirumala, H.~Soyer, J.~Z. Leibo,
  D.~Hassabis, and M.~Botvinick.
\newblock Prefrontal cortex as a meta-reinforcement learning system.
\newblock \emph{Nature neuroscience}, 21\penalty0 (6):\penalty0 860--868,
  2018{\natexlab{b}}.

\bibitem[Waskom(2021)]{waskom2021seaborn}
M.~L. Waskom.
\newblock Seaborn: statistical data visualization.
\newblock \emph{Journal of Open Source Software}, 6\penalty0 (60):\penalty0
  3021, 2021.

\bibitem[Widrow and Hoff(1960)]{widrow1960adaptive}
B.~Widrow and M.~E. Hoff.
\newblock Adaptive switching circuits.
\newblock Technical report, Stanford Univ Ca Stanford Electronics Labs, 1960.

\bibitem[Zendle and Cairns(2018)]{zendle2018video}
D.~Zendle and P.~Cairns.
\newblock Video game loot boxes are linked to problem gambling: Results of a
  large-scale survey.
\newblock \emph{PloS one}, 13\penalty0 (11):\penalty0 e0206767, 2018.

\bibitem[Zhang et~al.(2009)Zhang, Berridge, Tindell, Smith, and
  Aldridge]{zhang2009neural}
J.~Zhang, K.~C. Berridge, A.~J. Tindell, K.~S. Smith, and J.~W. Aldridge.
\newblock A neural computational model of incentive salience.
\newblock \emph{PLoS Comput Biol}, 5\penalty0 (7):\penalty0 e1000437, 2009.

\bibitem[Zhu and Laptev(2017)]{zhu2017deep}
L.~Zhu and N.~Laptev.
\newblock Deep and confident prediction for time series at uber.
\newblock In \emph{2017 IEEE International Conference on Data Mining Workshops
  (ICDMW)}, pages 103--110. IEEE, 2017.

\bibitem[Zou and Hastie(2005)]{zou2005regularization}
H.~Zou and T.~Hastie.
\newblock Regularization and variable selection via the elastic net.
\newblock \emph{Journal of the royal statistical society: series B (statistical
  methodology)}, 67\penalty0 (2):\penalty0 301--320, 2005.

\end{thebibliography}

\clearpage

\begin{appendices}

\section{Representation Analysis}
\label{appendix_representation}

\begin{figure}[h]
\centering
\begin{subfigure}[b]{\columnwidth}
    \includegraphics[width=\columnwidth]{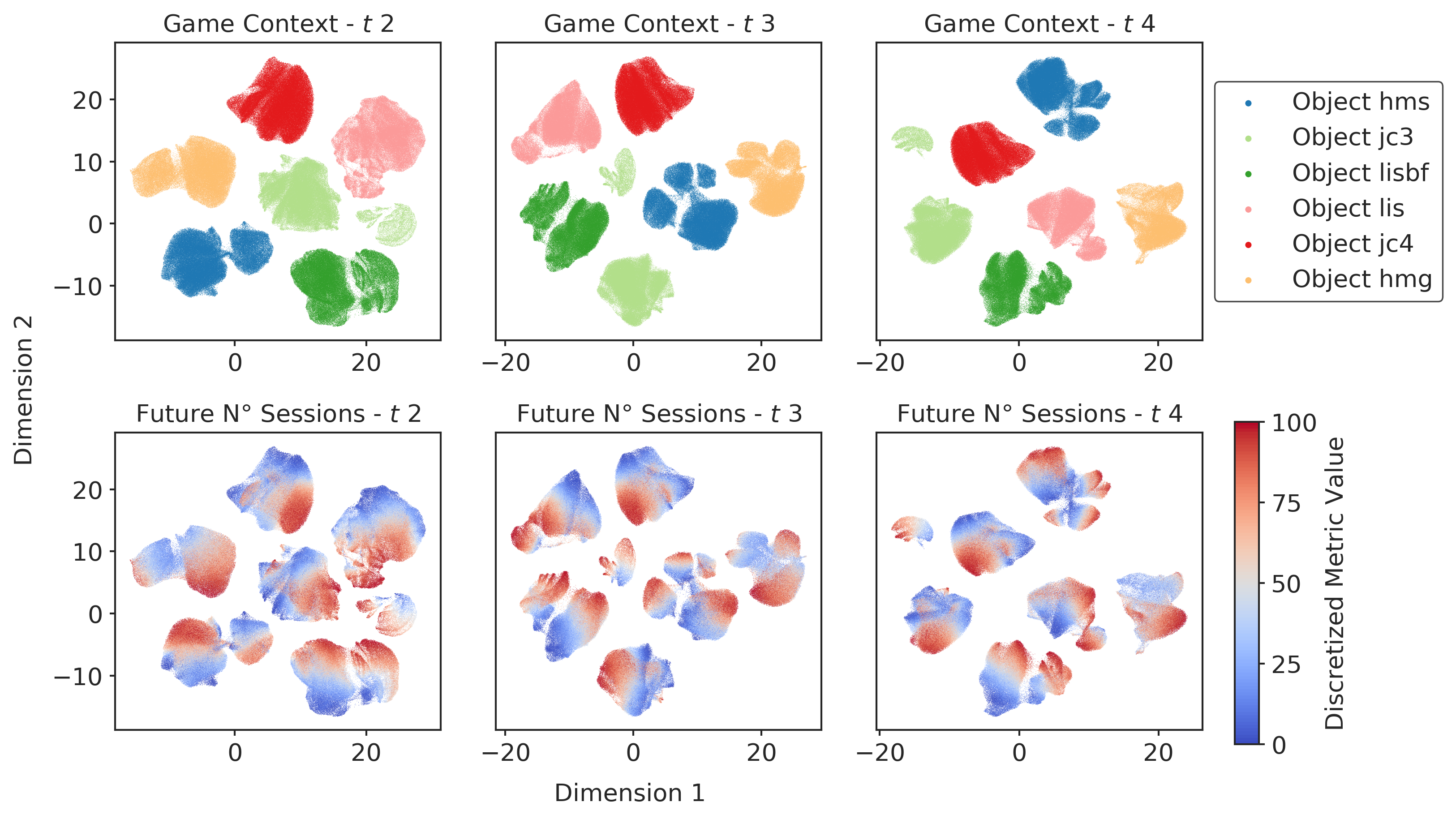}
    \label{umap_n_sess}
\end{subfigure}
\begin{subfigure}[b]{\columnwidth}
    \includegraphics[width=\columnwidth]{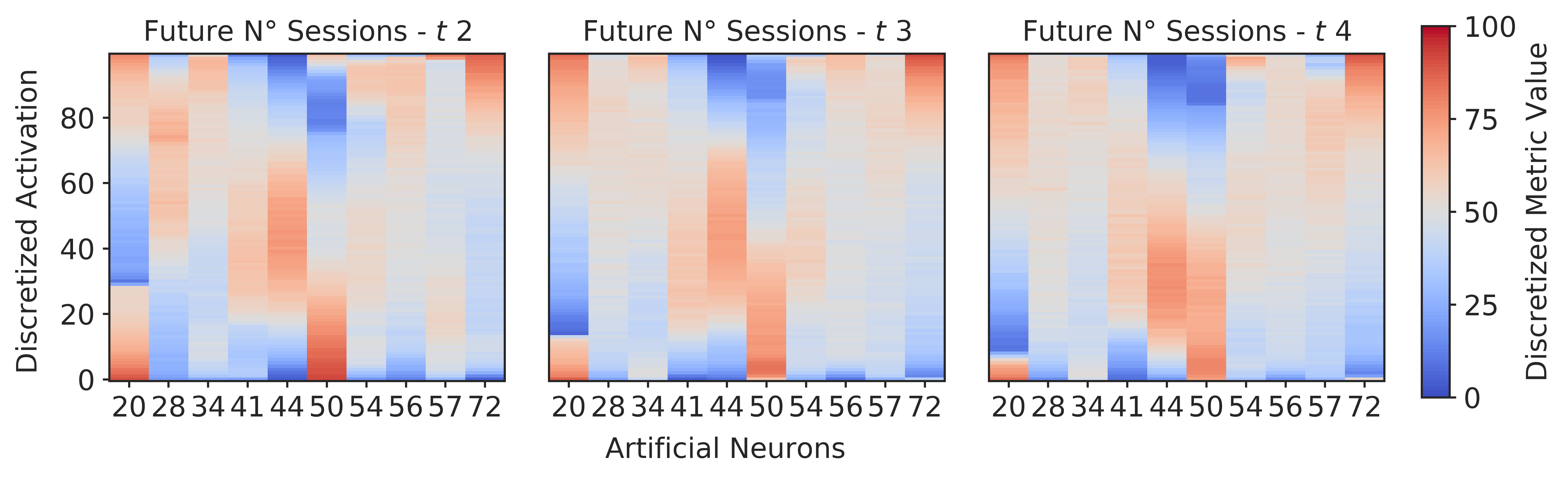}
    \label{profile_n_sess}
\end{subfigure}
\caption{\textbf{UMAP reduction and artificial neurons activations profile of the RNN representation at $t2$, $t3$ and $t4$ for the target Future N° Sessions.}}
\end{figure}

\begin{figure}[h]
\centering
\begin{subfigure}[b]{\columnwidth}
    \includegraphics[width=\columnwidth]{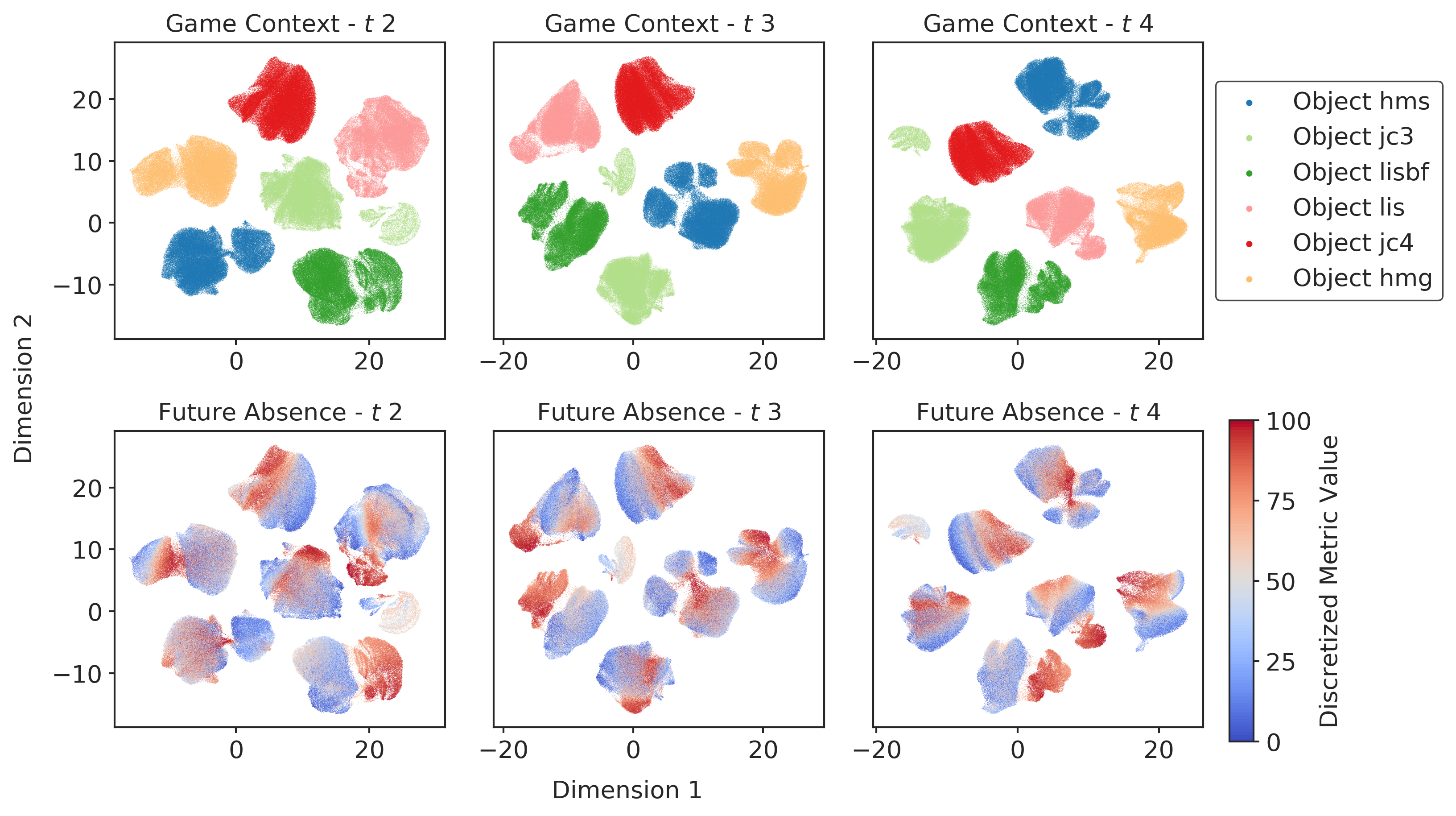}
    \label{umap_absence}
\end{subfigure}
\begin{subfigure}[b]{\columnwidth}
    \includegraphics[width=\columnwidth]{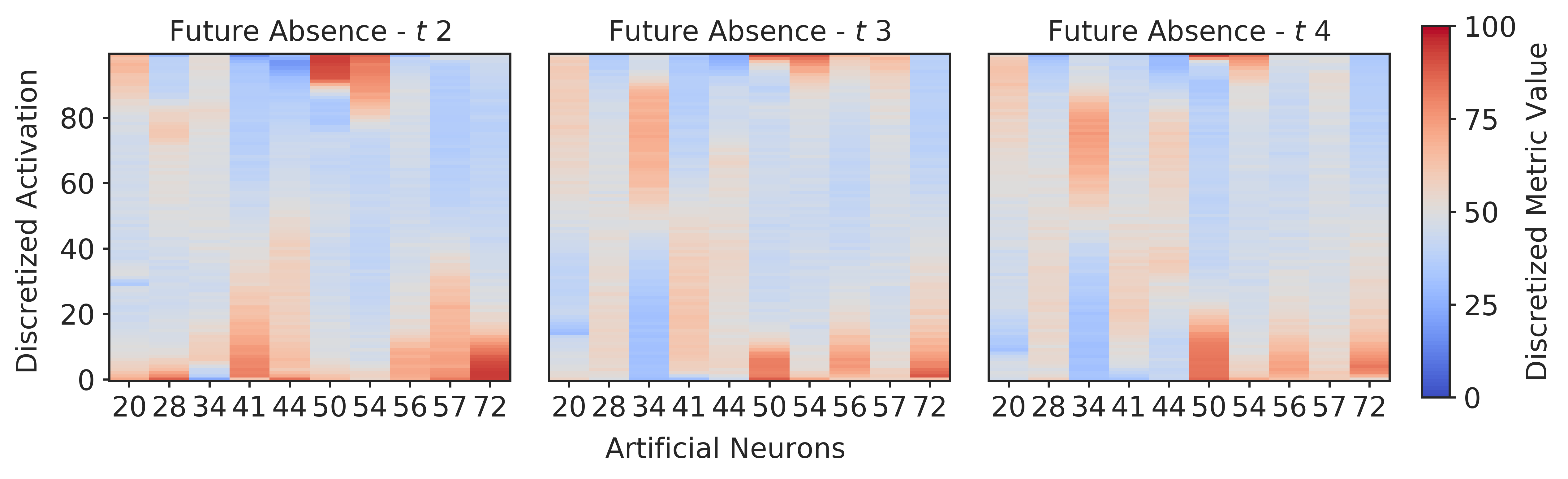}
    \label{profile_abs}
\end{subfigure}
\caption{\textbf{UMAP reduction and artificial neurons activations profile of the RNN representation at $t2$, $t3$ and $t4$ for the target Future Absence.}}
\end{figure}

\begin{figure}[h]
\centering
\begin{subfigure}[b]{\columnwidth}
    \includegraphics[width=\columnwidth]{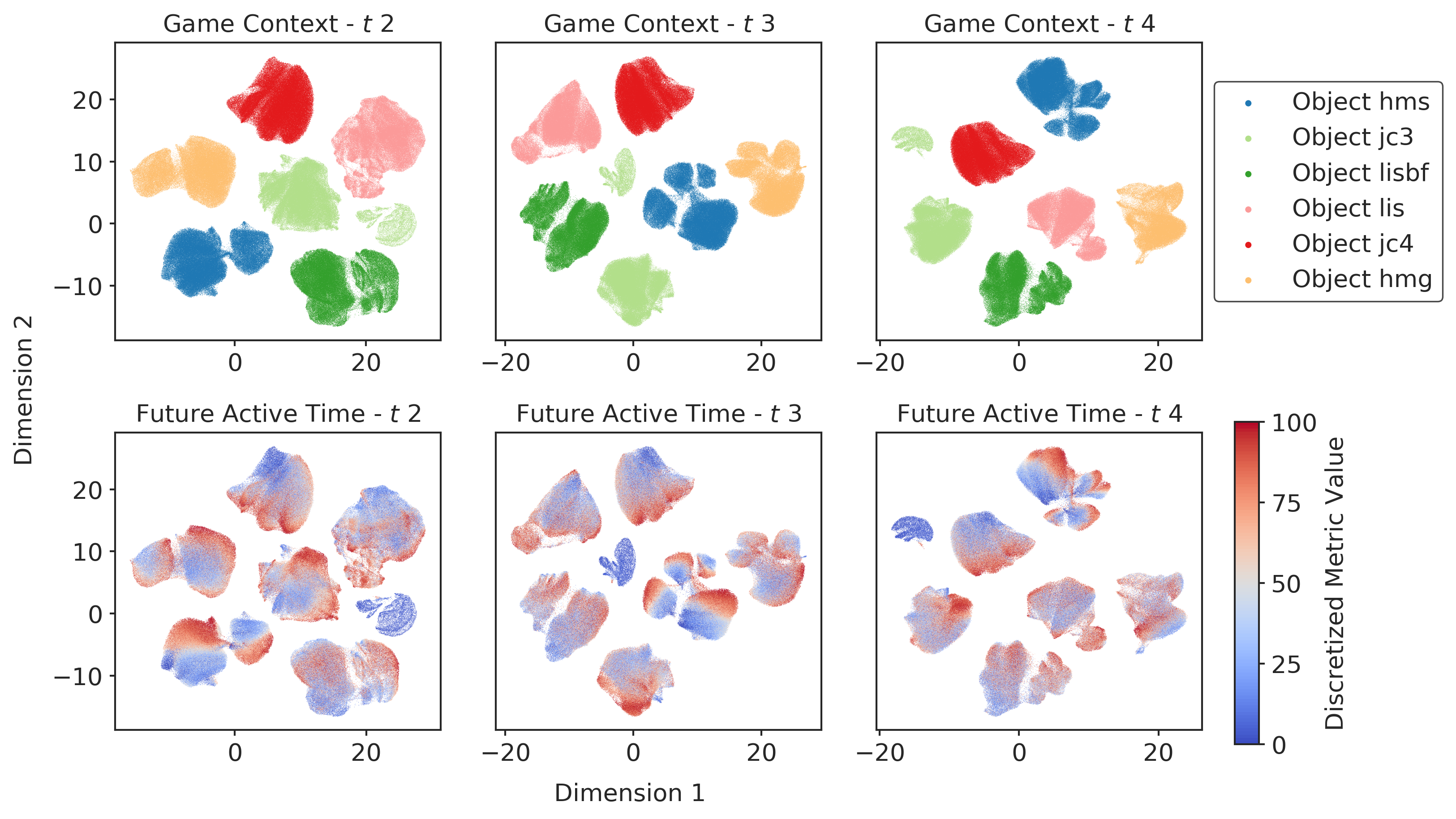}
    \label{umap_active_time}
\end{subfigure}
\begin{subfigure}[b]{\columnwidth}
    \includegraphics[width=\columnwidth]{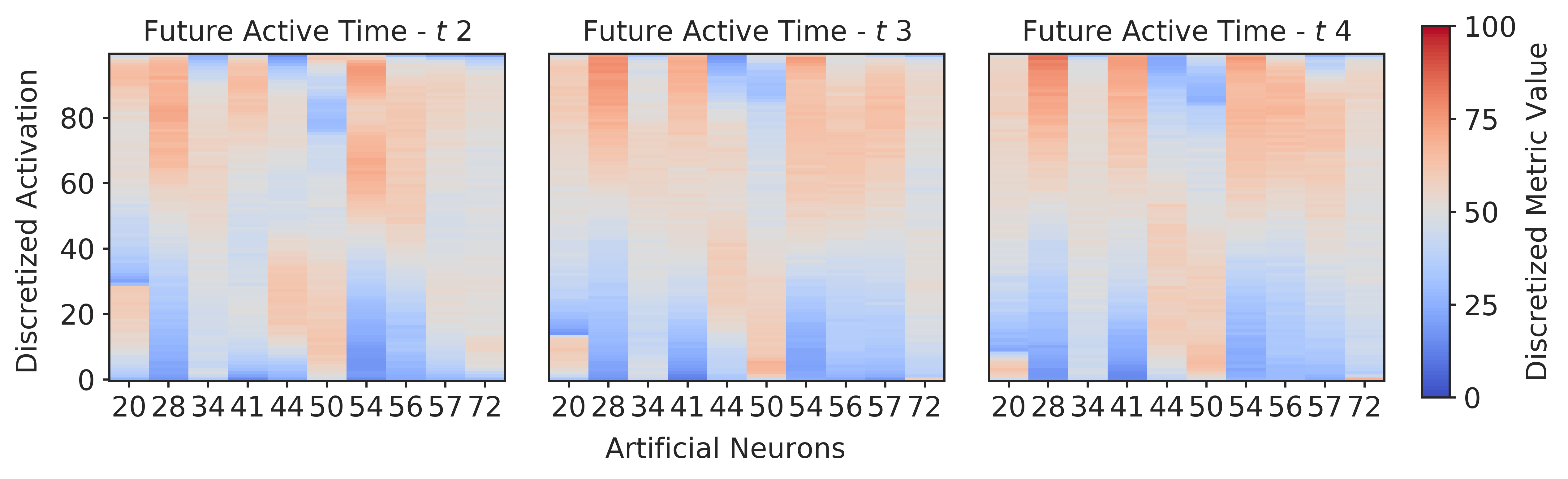}
    \label{profile_act_time}
\end{subfigure}
\caption{\textbf{UMAP reduction and artificial neurons activations profile of the RNN representation at $t2$, $t3$ and $t4$ for the target Future Active Time.}}
\end{figure}

\begin{figure}[h]
\centering
\begin{subfigure}[b]{\columnwidth}
    \includegraphics[width=\columnwidth]{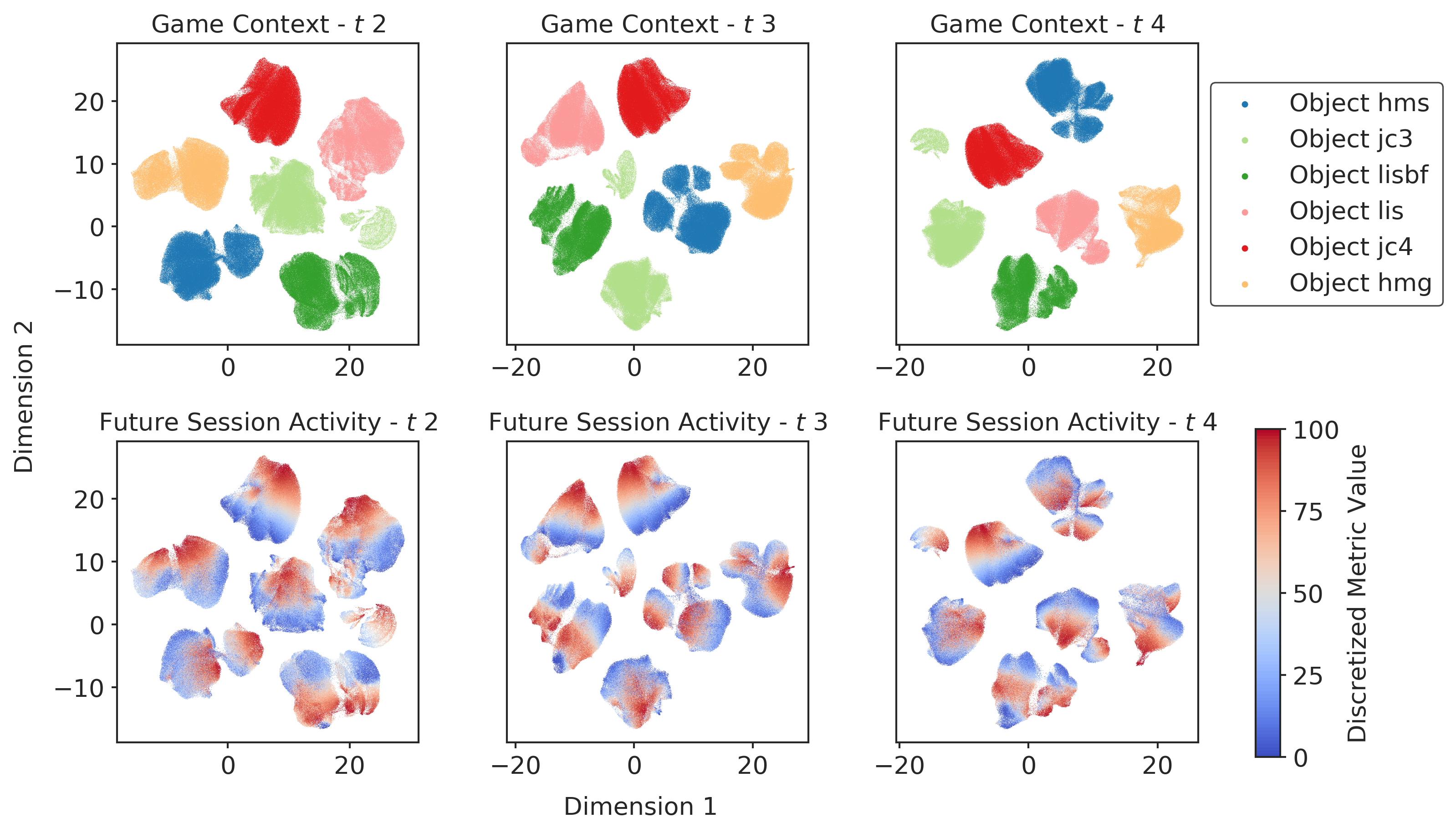}
    \label{umap_act}
\end{subfigure}
\begin{subfigure}[b]{\columnwidth}
    \includegraphics[width=\columnwidth]{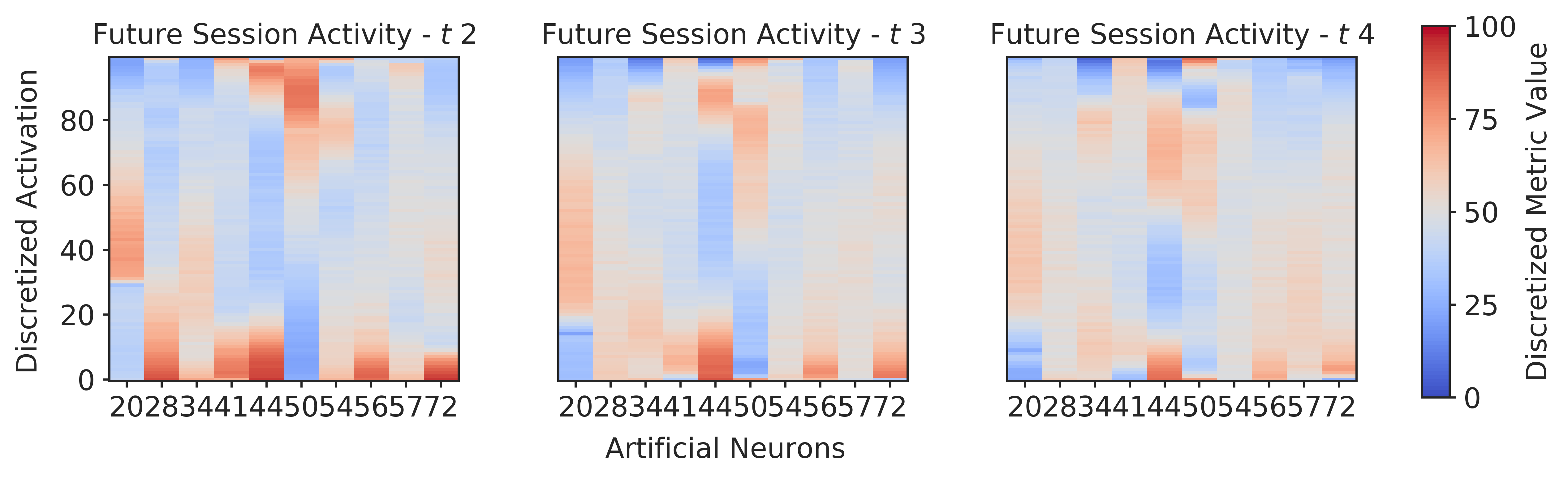}
    \label{profile_act}
\end{subfigure}
\caption{\textbf{UMAP reduction and artificial neurons activations profile of the RNN representation at $t2$, $t3$ and $t4$ for the target Future Session Activity.}}
\end{figure}

\section{Partition Analysis}
\label{appendix_partition}

\begin{figure}[h]
\centering
\includegraphics[width=\columnwidth]{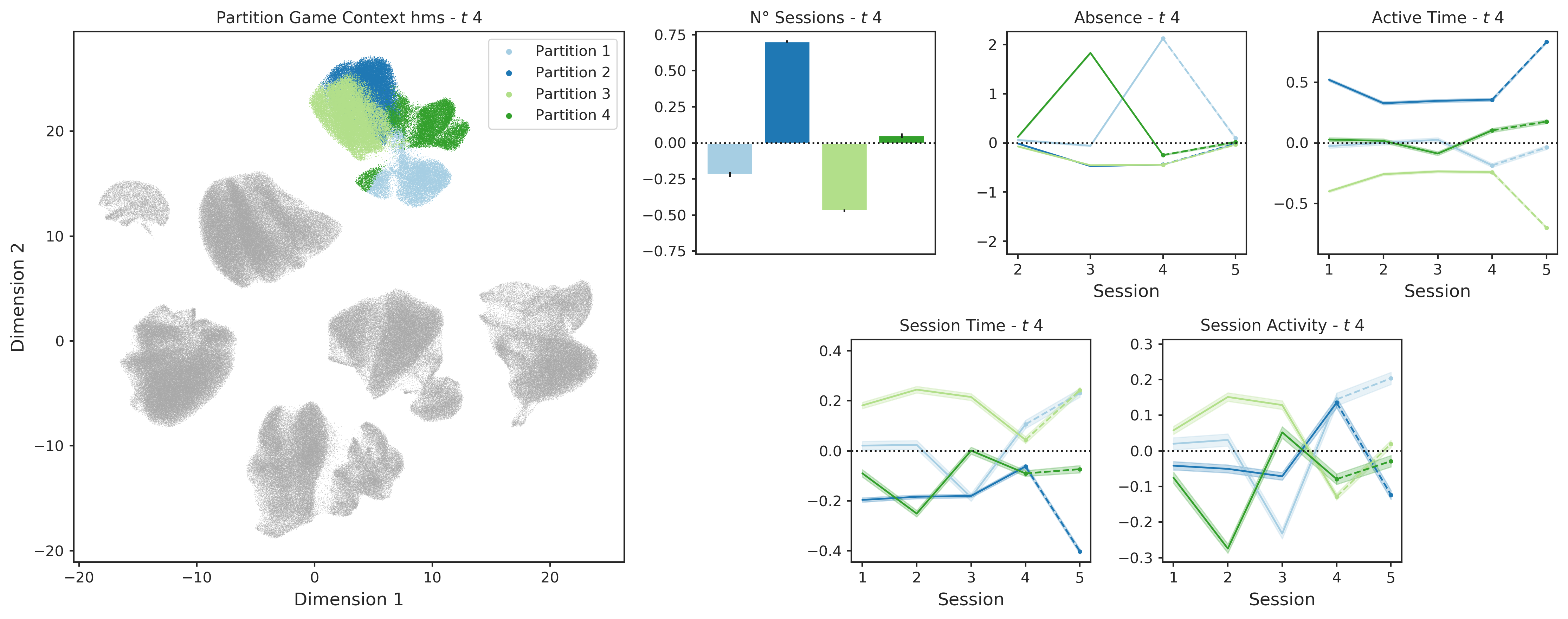}
\caption{\textbf{Partitions and associated behavioural profiles for the game object hms.}}
\label{hms_part} 
\end{figure}

\begin{figure}[h]
\centering
\includegraphics[width=\columnwidth]{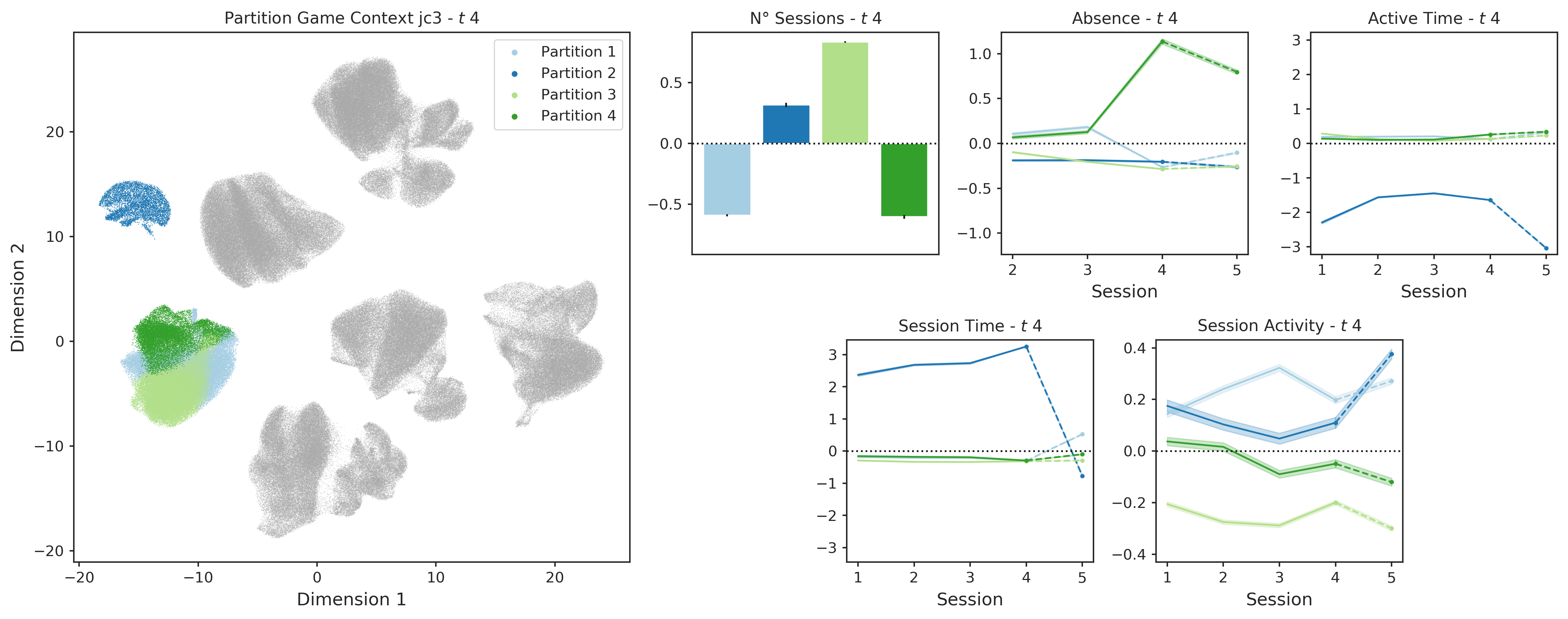}
\caption{\textbf{Partitions and associated behavioural profiles for the game object jc3.}}
\label{jc3_part} 
\end{figure}

\begin{figure}[h]
\centering
\includegraphics[width=\columnwidth]{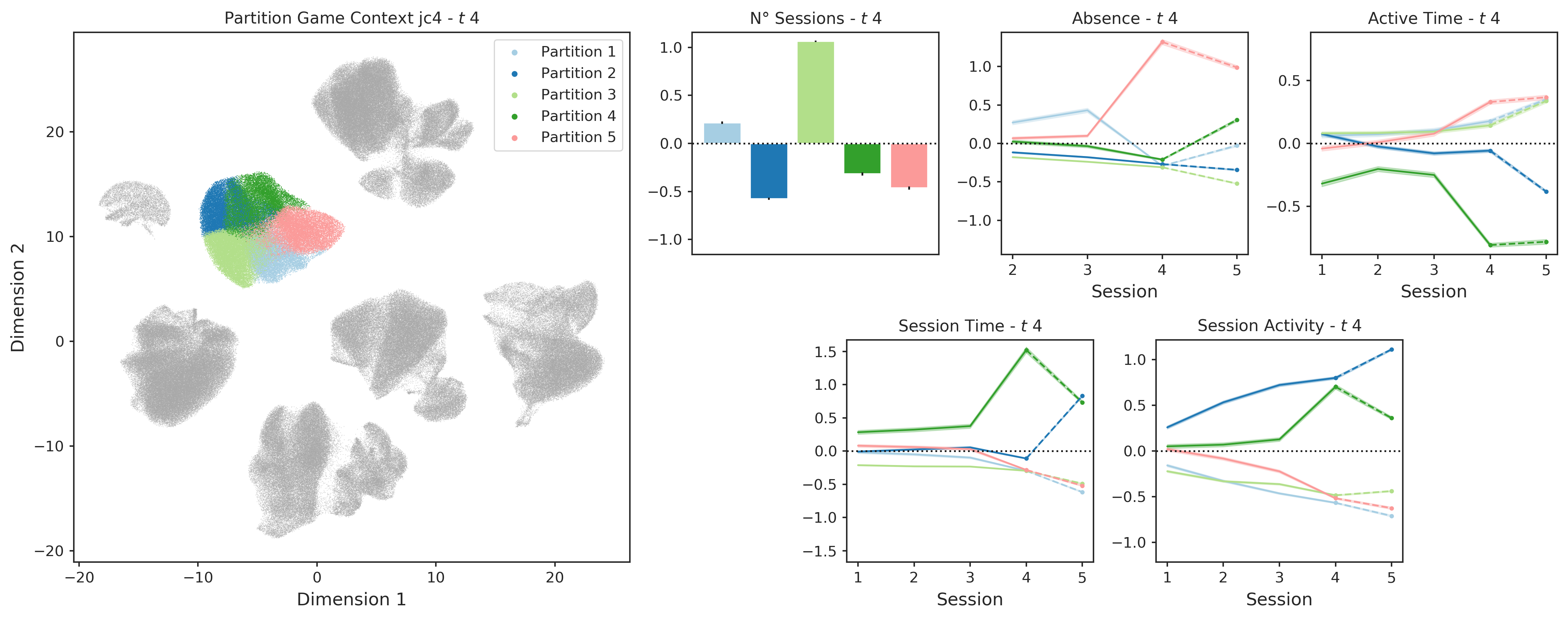}
\caption{\textbf{Partitions and associated behavioural profiles for the game object jc4.}}
\label{jc4_part} 
\end{figure}

\begin{figure}[h]
\centering
\includegraphics[width=\columnwidth]{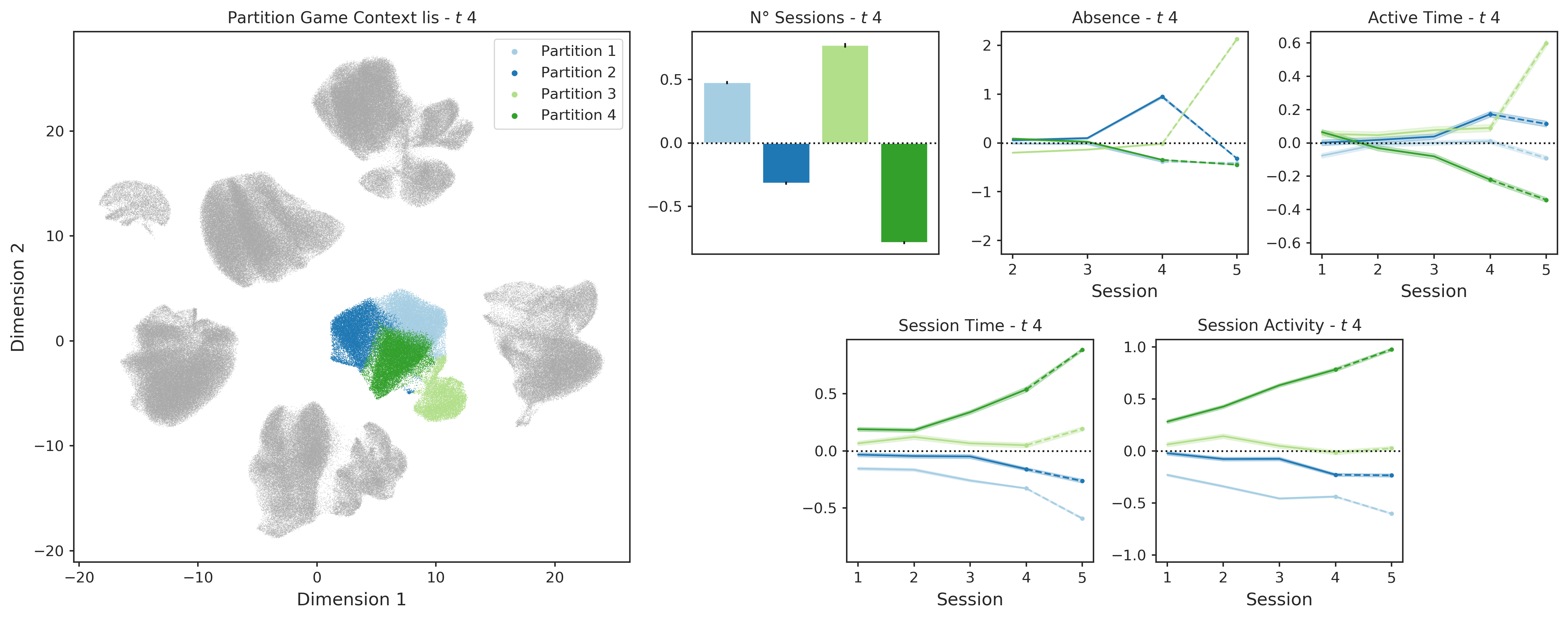}
\caption{\textbf{Partitions and associated behavioural profiles for the game object lis.}}
\label{lis_part} 
\end{figure}

\begin{figure}[h]
\centering
\includegraphics[width=\columnwidth]{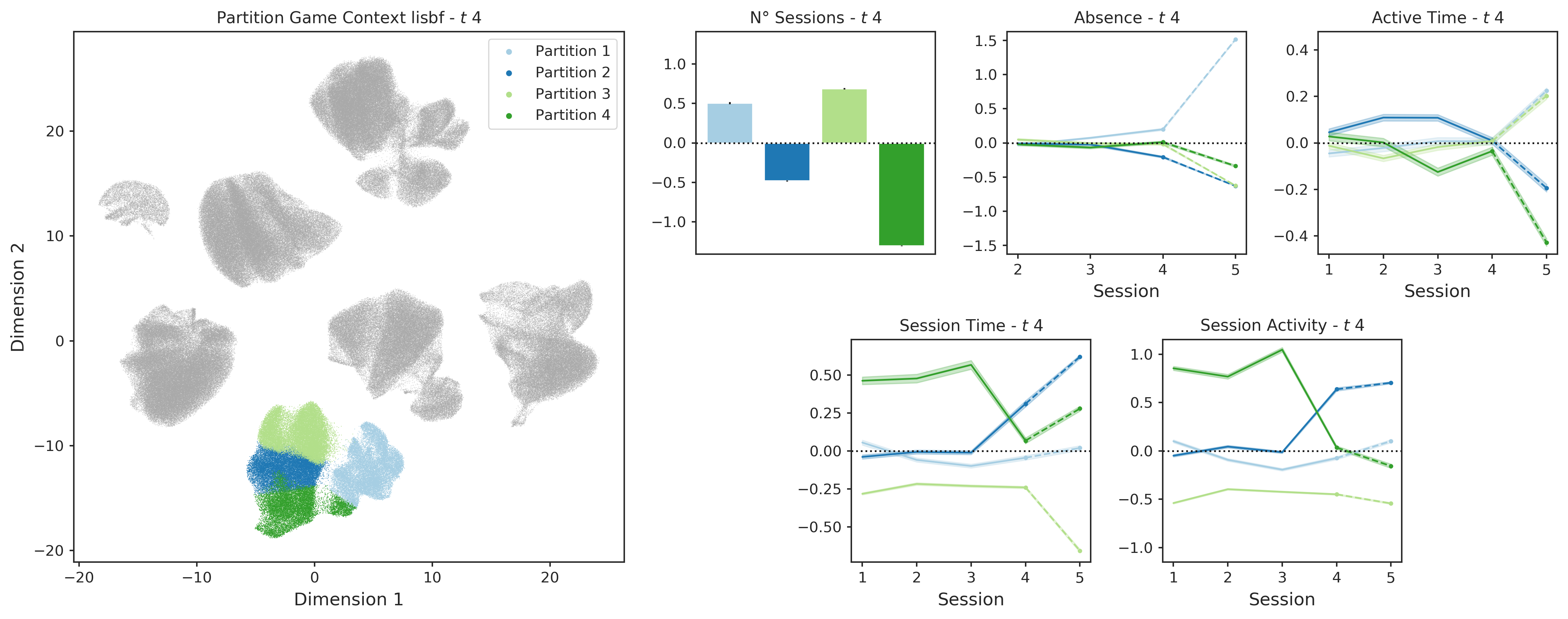}
\caption{\textbf{Partitions and associated behavioural profiles for the game object lisbf.}}
\label{lisbf_part} 
\end{figure}

\end{appendices}

\end{document}